\documentclass[a4paper,11pt]{article}
\usepackage{graphicx,times,amsmath}
\usepackage[ruled,vlined,linesnumbered]{algorithm2e}
\usepackage{algorithmic}
\usepackage{amsfonts}
\usepackage{subfigure}
\begin{document}
\date{}
\title{Meme as Building Block for Evolutionary Optimization of Problem Instances}
\author{Liang Feng, Yew-Soon Ong, Ah-Hwee Tan and Ivor Wai-Hung Tsang\thanks{Liang Feng, Yew-Soon Ong, Ah-Hwee Tan and Ivor Wai-Hung Tsang are with the Center for Computational Intelligence, School of Computer Engineering, Nanyang Technological University, Singapore. E-mail: \{feng0039, asysong, asahtan, ivortsang\}@ntu.edu.sg}}

\maketitle

\begin{abstract}
A significantly under-explored area of evolutionary optimization in the literature is the study of optimization methodologies that can evolve along with the problems solved. Particularly, present evolutionary optimization approaches generally start their search from scratch or the ground-zero state of knowledge, independent of how similar the given new problem of interest is to those optimized previously. There has thus been the apparent lack of automated knowledge transfers and reuse across problems. Taking the cue, this paper introduces a novel \emph{Memetic Computational Paradigm for search}, one that models after how human solves problems, and embarks on a study towards intelligent evolutionary optimization of problems through the transfers of structured knowledge in the form of memes learned from previous problem-solving experiences, to enhance future evolutionary searches. In particular, the proposed memetic search paradigm is composed of four culture-inspired operators, namely, \emph{Meme Learning}, \emph{Meme Selection}, \emph{Meme Variation} and \emph{Meme Imitation}. The learning operator mines for memes in the form of latent structures derived from past experiences of problem-solving. The selection operator identifies the fit memes that replicate and transmit across problems, while the variation operator introduces innovations into the memes. The imitation operator, on the other hand, defines how fit memes assimilate into the search process of newly encountered problems, thus gearing towards efficient and effective evolutionary optimization. Finally, comprehensive studies on two widely studied challenging well established NP-hard routing problem domains, particularly, the capacitated vehicle routing (CVR) and capacitated arc routing (CAR), confirm the high efficacy of the proposed memetic computational search paradigm for intelligent evolutionary optimization of problems.
\end{abstract}

\section{Introduction}
Like gene in genetics, a meme is synonymous to memetic as being the building block of cultural know-how that is transmissible and replicable \cite{XYMK11}. In the last decades, meme has inspired the new science of memetics which today represents the mind-universe analog to genetics in cultural evolution, stretching across the field of biology, cognition, psychology, and sociology \cite{YMX10}.

Looking back on the history of \emph{meme}, the term can be traced back to Dawkins \cite{RD76} in his book ``The selfish Gene'', where he defined it as ``a unit of information residing in the brain and is the replicator in human cultural evolution''. Like genes that serve as ``instructions for building proteins'', memes are then ``instructions for carrying out behavior, stored in brains''. As discussed by Blackmore in her famous book ``The Meme Machine'', where she reaffirmed meme as information copied from one person to another and discussed on the theory of ``memetic selection'' as the survival of the fittest among competitive ideas down through generations \cite{SB99}. Other definitions of meme that took flights from there have since emerged to include ``memory item, or portion of an organism's neurally-stored information'' \cite{AL91}, ``unit of information in a mind whose existence influences events such that more copies of itself get created in other minds'' \cite{RB96}, and ``contagious information pattern that replicates by parasitically infecting human minds'' \cite{GG90}.

In the context of computational intelligence, particularly evolutionary computation, a meme has been perceived as a form of individual learning procedure, adaptive improvement procedure or local search operator to enhance the capability of population based search algorithm \cite{YA04, QYM09}. From the last decades, this integration has been established as an extension of the canonical evolutionary algorithm, by the names of hybrid, adaptive hybrid or Memetic Algorithm (MA) in the literature \cite{XYMK11}, where many success stories for solving complex real world search problems ranging from continuous optimization \cite{QYM09}, combinatorial optimization \cite{LYQA10, KY09}, constrained optimization \cite{ARDC09}, multi-objective optimization \cite{JDDW00} to optimization in dynamic environment \cite{WWY09}, etc., have been reported. In spite of the attention and successes that memetic algorithm has enjoyed, meme does seem to play much of a complimentary role in computational intelligence, i.e., in the form of a local search or refinement operator within the evolutionary search cycle.

Falling back on the basic definition of a meme by Dawkins and Blackmore, as the fundamental building blocks of culture evolution, research on memetic computation can perhaps be more meme-centric focus by treating memes as the building blocks of a given problem domain, much like gene serving as the building blocks of life. Taking the cue, this paper embarks on a study towards a novel \emph{Memetic Computational Paradigm for search}, one that models after how human solves problems.

Today, it is well recognized that the processes of learning and the transfer of what has been learned are central to humans in problem-solving \cite{JAR00}. Learning has been established to be fundamental to human in functioning and adapting to the fast evolving society. Besides learning from the successes and mistakes of the past and learning to avoid making the same mistakes again, the ability of human in selecting, generalizing and drawing upon what have been experienced and learned in one context, and extending them to new problems is deem to be most remarkable \cite{JP96}. Since the inception of artificial intelligence more than five decades ago, a significant number of success stories to emulate the learning, reasoning and problem-solving intelligence of humankind have been made. The aspiration has been to develop machines that do things like humans are capable of performing intelligibly. Within the context of computational intelligence, several core learning technologies in neural and cognitive systems, fuzzy systems, probabilistic and possibilistic reasoning have been notable for their ability in emulating some of human's cultural and generalization capabilities, with many now used to enhance our daily life. \emph{In spite of the accomplishments made in computational intelligence, the attempts to emulate the cultural intelligence of human in search, evolutionary optimization in particular, has to date received far less attention}. The learning, generalization and evolution of useful traits across related tasks or problems and the study of optimization methodology that evolves along with the problems solved has been significantly under-explored in the context of evolutionary computation. In the present study, we aspire to fill in this gap.

In particular, we propose a memetic search paradigm that takes advantage of the possible relations among problem instances, i.e., topological properties, data distribution or otherwise, to allow the effective assessments of future unseen problem instances, without the need to perform an exhaustive search each time or start the evolutionary search from a ground-zero knowledge state. The proposed memetic search paradigm is composed of four culture-inspired operators, namely, \emph{Meme Learning}, \emph{Meme Selection}, \emph{Meme Variation} and \emph{Meme Imitation}. The \emph{learning operator} extracts memes in the form of structured knowledge from past experiences of problem-solving. In the present context, meme manifests as the instructions to intelligently bias the search (i.e., thus narrowing down the search space) on the given problems of interest. The \emph{selection operator}, on the other hand, selects the appropriate or fit memes that shall replicate and undergo innovations via the \emph{variation operator}, before drawing upon them for enhancing future evolutionary optimizations. Last but not least, the \emph{imitation operator} defines the process of memes assimilation into the search process of new problems, via the generations of positive biased solutions that would lead to more efficient and effective evolutionary searches.

In this paper, we showcase the proposed memetic evolutionary search paradigm on combinatorial optimization problems. Particularly, we consider the general programming problem where there exists a number of agents and a number of tasks. Any agent can be assigned to perform any task, incurring some cost and profit that may vary, depending on the agent-task assignment. Moreover, each agent has a budget constraint and the sum of the costs of tasks assigned to it cannot exceed the constraint imposed. The objective is then to find an assignment with maximum total profit, while not exceeding the budget imposed on each agent. In the literature, several well established combinatorial optimization problems belonging to the described general programming problem include vehicle routing problem \cite{LR96}, generalized assignment problem \cite{Chu199717}, arc routing problem \cite{LYQA10}, etc. To demonstrate the memetic computational search paradigm for ``intelligent'' evolutionary optimization, the two widely studied challenging routing problem domains, namely, capacitated vehicle routing (CVR) and capacitated arc routing (CAR) are showcased in the current paper. To summarize, the core contributions of the current work is multi-facets, which are outlined as follows:
\begin{enumerate}
\item To date, almost all search methods start the optimization process from scratch, with the assumption of zero usable information, i.e., independent of how similar the current problem instance of interest is to those encountered in the pasts. The current work presents an attempt to fill this gap by embarking a study on evolutionary optimization methodology with intellectual capabilities that can evolve along with the problems solved.
\item To our knowledge, this is the first work that focuses on cultural evolutionary optimization of problem instances. It presents a novel effort on learning of useful traits from past problem solving experiences in the form of structured knowledge or latent patterns as memes, and subsequently through the mechanisms of cultural learning, selection, variation and imitation, appropriately acquired knowledge are transferred to enhance the evolutionary search on future unseen problems.
\item Beyond the formalism of simple and adaptive hybrids as memetic algorithm, this paper introduces and showcases the novel representation of acquired knowledge from past optimization experiences in the form of memes. In contrast to the manifestation of memes as refinement procedures in hybrids, here memes manifest as natural building blocks of meaningful information, and in the present context, serving as the instructions for generating solutions that would lead towards optimized solutions both effectively and efficiently.
\item This work proposes a storage of building blocks or memes to problems that share some common characteristics, such as instances from related problem domains, and supports the reuse of the memes across problems. The capacity to imitate and draw on memes from past instances of problem-solving sessions thus allows future evolutionary search to be more intelligent, leading to solutions that can be attained more efficiently on unseen problems of increasing complexity or dynamic in nature. These atomized units of memes then form societies of the mind for more effective problem solving.
\item In the experimental study, the proposed memetic evolutionary search paradigm is shown to be more efficient and effective for problem solving than conventional evolutionary optimization. Specifically, we show that the memetic paradigm enhances the search performances of existing state-of-the-art evolutionary optimization methodologies in the domains of capacitated vehicle routing and capacitated arc routing, which contain NP-hard combinatorial optimization problems of diverse properties.
\end{enumerate}

The rest of this paper is organized as follows: a brief discussion on traditional evolutionary optimization and the related works is given in Section \ref{RW}. Section \ref{ME} introduces the proposed memetic computational search paradigm for ``intelligent'' evolutionary optimization of problems, via the transfer of structured knowledge in the form of memes from past problem-solving experiences, to enhance future evolutionary searches. Section \ref{CSCC} presents the brief mathematical formulations of the capacitated vehicle routing problem (CVRP) and capacitated arc routing problem (CARP), and discusses the existing state-of-the-art optimization methodologies for solving CVRP and CARP. Detailed designs of the \emph{meme learning}, \emph{meme selection}, \emph{meme variation} and \emph{meme imitation} operators for routing problems are then described in Section \ref{CPIEO}. Particularly, the culture-inspired operators are realized based on the \emph{Hilbert-Schmidt Independence} (HSIC) \cite{Gretton05measuringstatistical}, \emph{Maximum Mean Discrepancy} criteria \cite{KMAM06} and \emph{K-means Clustering}. Last but not least, section \ref{ES} presents and analyzes the experimental results on the well established CVRP and CARP benchmarks, which serve as the representatives routing problems considered in the current work. Lastly, the conclusive remarks of this paper are drawn in Section \ref{Con}.

\section{Related Works}\label{RW}
Evolutionary algorithms (EAs) are approaches that take their inspirations from the principles of natural selection and survival of the fittest in the biological kingdom. In the last decades, it is becoming well established that many effective optimizations using EA have been achieved by the use of inductive biases that fit to the structure of the problem of interest well \cite{PE08}. In spite of the great deal of attention that EA has received, where many specialized EAs have been developed with the manual incorporations of human expert domain knowledge (and hence inductive biases) into the search scheme, it is worth noting here that, existing EA approaches have yet to exploit the useful traits that may exist in similar tasks or problems. In particular, the study of optimization methodology that evolves along with the problems solved has been significantly under-explored in the context of evolutionary computation. Instead, the common practice of evolutionary computation is to start the search on a given new problem of interest from ground zero state, independent of how similar the new problem instance of interest is to those encountered in the past. Thus a major impediment of existing evolutionary search methodology in the literature is the \emph{apparent lack of automated useful knowledge transfers and reuse from past problem solving experiences as appropriate biases that can enhance evolutionary search of new problems}.

In practice, problems seldom exist in isolation, and previous related problems encountered often yield useful information that when properly harnessed, can lead to more effective future evolutionary search. To date, few works have attempted to reuse useful traits from across problems. Louis \emph{et al.} \cite{SJJM04}, for instance, presented a study to acquire problem specific knowledge and subsequently using them to aid in the genetic algorithm (GA) search via case-based reasoning. Rather than starting anew on each problem, appropriate intermediate solutions drawn from similar problems that have been previously solved are periodically injected into the GA population. In a separate study, Cunningham and Smyth \cite{PB97} also explored the reuse of established high quality schedules from past problems to bias the search on new traveling salesman problems (TSPs). Despite these early attempts, the idea of reusing useful traits from across problems in evolutionary search did not garner significant attention. We believe this is a result of the weak generality trait of the approaches introduced in these initial studies. Particularly, both \cite{SJJM04} and \cite{PB97} considered the exact storage of past solutions or partial-solutions from previous problems solved, and subsequently regurgitate them directly into the solution population of a new evolutionary search. It is worth noting that such a simple scheme of the memory at work would proof to be futile when the nature or characteristics of the new optimization problem of interest differs, such as the problem size or dimensionality, structures, representations, etc. Thus what have been previously memorized from past problems solved cannot be directly injected into new searches for successful reuse.

In contrast to existing memory based approaches, the proposed memetic computational search paradigm addresses the non-trivial task of learning the generic building blocks or memes of useful traits from past problems solving experiences and subsequently drawing upon them through the cultural evolutionary mechanisms of meme learning, selection, variation and imitation (as opposed to a simple direct copying of past solutions in previous works) to enhance the search on new problems. In such a manner, the transfer of memes as generic building blocks of useful knowledge, can lead to enhanced search on problems of differing dimensions, topological structures, and representations, etc.

\section{Memetic Evolution} \label{ME}
In this section we present the proposed memetic computational paradigm for ``intelligent'' evolutionary optimization, one that is modelled after how human solves problems, capable of evolving along with the problems solved.
In the proposed memetic computational paradigm, the instructions for carrying out the behavior to act on a given problem are modeled as memes stored in brains. These memes then serve as the building blocks of past problems solving experiences that may be efficiently passed on to support the search on future unseen problems, by means of \emph{cultural evolution}. This capacity to draw on memes from previous instances of problem-solving sessions thus allows the search to be more intelligent, leading to a more effective and efficient future search.

\subsection{Cultural Operators}
\begin{figure*}[ht]
    \centering
    \includegraphics[width=0.8\textwidth]{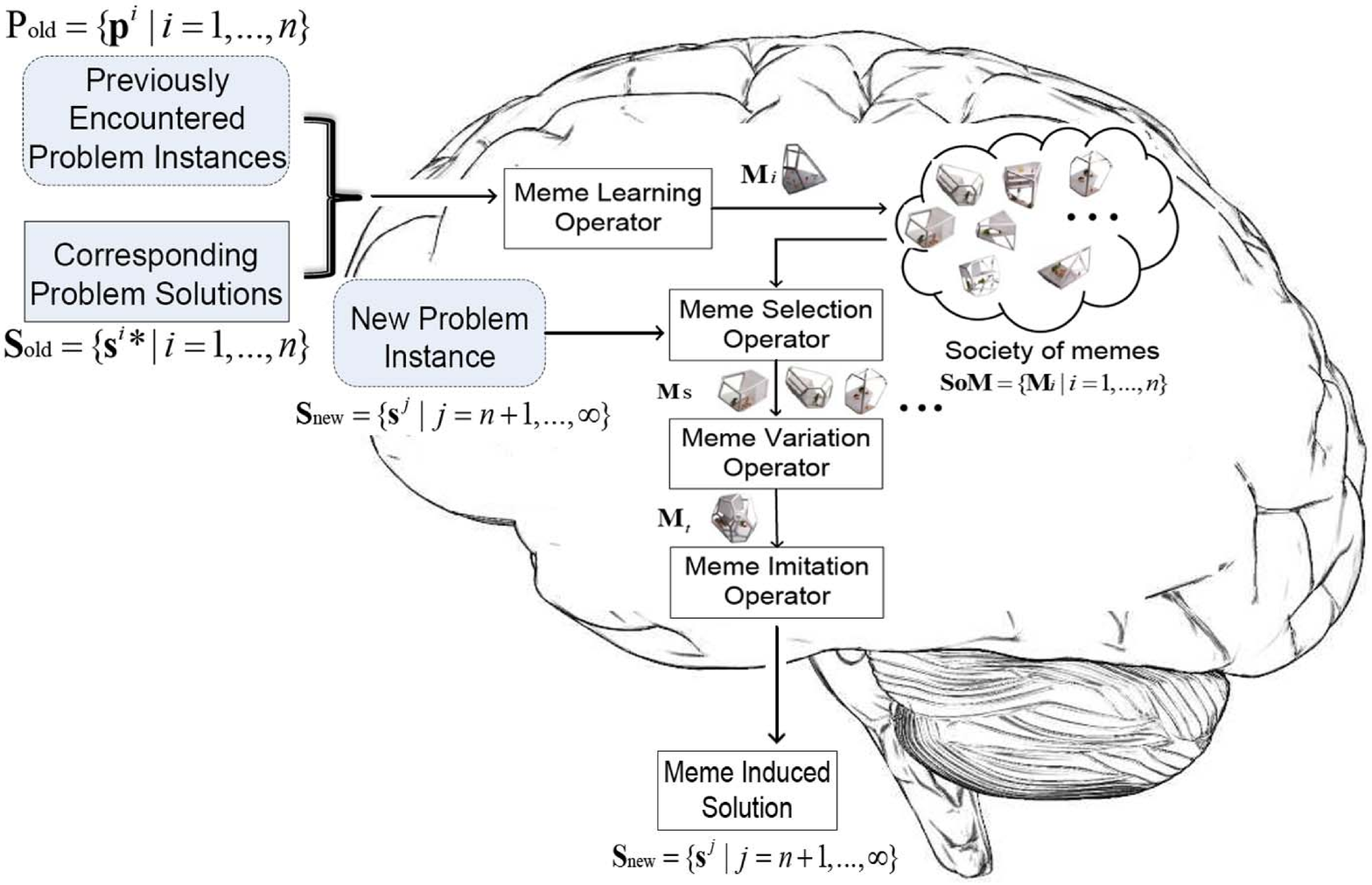}
    \caption{A depiction of how human solves problems in Memetic Computation.}
    \label{IOPEF}
\end{figure*}

The proposed memetic computational paradigm is composed of four culturally-inspired operators, namely \emph{Meme Learning}, \emph{Meme Selection}, \emph{Meme Variation} and \emph{Meme Imitation} as depicted in Fig. \ref{IOPEF}, whose functions are described in what follows:
\newline
\begin{enumerate}
\item[-] \emph{Meme Learning} Operator: Given that $\mathbf{p}$ corresponds to a problem instance and $\mathbf{s^*}$ denotes the optimized solution of $\mathbf{p}$, as attained by an evolutionary solver (labeled here as $OS$). The learning operator takes the role of modelling and generalizing the mapping from $\mathbf{p}$ to $\mathbf{s^*}$, to derive the meme. The meme learning process evolves in an incremental manner, and builds up the wealth of knowledge in the form of memes, along with the number of problem instances solved. Note the contrast to a simple storage or exact memory of specific problem instance $\mathbf{p}$ with associated solution $\mathbf{s^*}$ as considered in the previous studies based on case-based reasoning \cite{SJJM04}.
    \newline
\item[-] \emph{Meme Selection} Operator: Different prior knowledge introduces unique forms of bias into the search. Hence a certain biases would make the search more efficient on some classes of problem instances but not for others. Inappropriately harnessed knowledge, on the other hand, may lead to the possible impairments of the search. The meme selection operator thus serves to identify or select the fit memes, from the society of memes, that replicate successfully.
    \newline
\item[-] \emph{Meme Variation} Operator: The meme variation forms the intrinsic innovation tendency of the cultural evolution. Without variations, maladaptive form of bias may be introduced in the evolutionary searches involving new problem instances. For instance, a piece of knowledge as represented by the memes, which has been established as constructive based on its particular demonstration of success on a given problem instance would quickly spiral out of control via replication. This will suppress the diversity and search of the evolutionary optimization across problems. Therefore, variation is clearly essential for retaining diversity in the society of memes towards efficient and effective evolutionary search.
    \newline
\item[-] \emph{Meme Imitation} Operator: From Dawkins's book entitled ``The selfish Gene'' \cite{RD76}, ideas or memes are copied from one person to another via imitation. In the present context, memes that are learned from past problem solving experiences replicates by means of imitation and used to enhance future evolutionary search on newly encountered problems.
\end{enumerate}

\subsection{'\emph{Intelligent}' Evolutionary Optimization via Memetic Evolution}
In the spirit of how human solves problems, the schemata representation of meme in computing as the structured knowledge or latent patterns that is encoded in the mystical mind of nature is first identified. The problem solving experiences on the encountered problems are then captured via \emph{meme learning} and crystallized as a part of the meme pool that form the building blocks in the society of mind \cite{Minsky1986}. In this manner, whenever a new problem comes about, the \emph{meme selection} operator kicks in to first identify the appropriate memes from the wealth of previously accumulated memes. These memes then undergo \emph{variations} to effect the emergence of innovative memes. Enhancements to subsequent problem-solving effectiveness and efficiency on given new problems is then achieved by means of \emph{meme imitation}.

Referring to Fig. \ref{IOPEF}, at time step $i=1$, the optimization solver $OS$ is faced with the first problem instance $\mathbf{p^1}$ to search on. Since $\mathbf{p^1}$ denotes the first problem of its kind to be optimized, no memes is available for enhancing the evolutionary solver, $OS$, search. \footnote{Unless, of course, a database of memes that are learned from past problem solving experiences can be loaded and leveraged upon.} This is equivalent to the case where a novice encounters a new problem of its kind to work on, in the absence of \emph{a prior} knowledge that he/she can leverage upon. This condition is considered as ``no relevant meme available'' and the search by solver $OS$ shall proceed normally, i.e., the \emph{meme selection} operator remains dormant. The corresponding optimized solution attained by solver $OS$ on $\mathbf{p^1}$ is denoted by $\mathbf{s^{1*}}$. Through the \emph{meme learning} operator, meme $\mathbf{M_1}$ is learned and derived from the problem instance and corresponding optimized solution denoted by $({\mathbf{p^1},\mathbf{s^{1*}}})$. On subsequent unseen problem instances $j=2,\ldots,\infty$, \emph{meme selection} kicks in to identify the appropriate memes $\mathbf{M}$s from the society of memes denoted by $\mathbf{SoM}$. Activated memes $\mathbf{M}$s then undergo the \emph{variation} operator to arrive at innovative meme $\mathbf{M_t}$ that serves as the instructions \emph{imitated} to bias subsequent evolutionary optimizations by $OS$. In this manner, useful memes of experiences attained from previously solved problem instances are captured incrementally and archived in meme pool $\mathbf{SoM}$ to form the society of mind, which are appropriately activated to enhance future search performances.

Like the memes stored in the human mind to aid in the coping of our everyday life and problem solving, memes residing in the artificial mind of the evolutionary solver play the role of positively biasing the search on newly encountered problems. In this manner, the intellectual capability of the evolutionary solver evolves along with the number of problems solved, with transferrable memes accumulating with time. When a new problem is encountered, fit memes are activated and varied to instruct and guide the intelligent solver in the search process. The mind-universe of memes thus formed the evolving domain knowledge that may be activated to solve future evolutionary search effectively and efficiently.

\section{Case Studies on Routing Problems} \label{CSCC}
In this section, we illustrate the proposed memetic computational paradigm for ``intelligent'' evolutionary optimization on the class of combinatorial optimization problem. Particularly, our interest is on the general problem scenario where there exist a number of agents and a number of tasks. Any agent can be assigned to perform any task, incurring some cost and profit that may vary depending on the agent-task assignment. Moreover, each agent has a budget and the sum of the costs of tasks assigned to it cannot exceed this budget. It is required to find an assignment in which all agents do not exceed their budget and the total profit of the assignment is maximized. It is worth highlighting the practicality of described scenarios, where a plethora of today's real world complex problems can be established to exhibit many of the properties outlined. These include the well known combinatoric problems like vehicle routing in urban waste collection or post delivery, plant location, resource scheduling, flexible manufacturing systems, generalized assignment problem \cite{Chu199717}, job scheduling problem \cite{MT03}, and arc routing problem \cite{KY09, LYQA10}, etc. To demonstrate the memetic computational search paradigm for ``intelligent'' evolutionary optimization, the two widely studied challenging domains of the capacitated vehicle routing problem (CVRP) and capacitated arc routing problem (CARP) are showcased in the present section.

\subsection{Capacitated Vehicle Routing Problem}
The capacitated vehicle routing problem (CVRP) introduced by Dantzig and Ramser \cite{GJH59}, is a problem to design a set of vehicle routes in which a fixed fleet of delivery vehicles of uniform capacity must service known customer demands for single commodity from a common depot at minimum cost. The CVRP can be formally defined as follows. Given a connected undirected graph $G = (V, E)$, where vertex set $V = \{v_i\}, i=1 \ldots n$, $n$ is the number of vertices, edge set $E = \{e_{ij}\}, i, j=1 \ldots n$ denoting the arc between vertices $v_i$ and $v_j$. Vertices $v_d$ corresponds to the depot at which $k$ homogeneous vehicles are based, and the remaining vertices denote the customers. Each arc $e_{ij}$ is associated with a non-negative weight $c_{ij}$ , which represents the travel distance from $v_i$ to $v_j$. Consider a demand set $D = \{d(v_i) | v_i \in V\}$, where $d(v_i) > 0$ implies customer $v_i$ requires servicing (i.e., known as task), the CVRP consists of designing a set of least cost vehicle routes $\mathcal{R} = \{\mathcal{C}_i\},~i=1 \ldots k$ such that
\begin{enumerate}
\item Each route $\mathcal{C}_i,~i\in [1,k]$ must start and end at the depot node $v_d \in V$.
\item The total load of each route must be no more than the capacity $W$ of each vehicle, $\sum_{\forall v_i \in \mathcal{C}} d(v_i) \le W$.
\item $\forall v_i \in V~and~d(v_i) > 0$, there exists one and only one route $\mathcal{C}_i \in \mathcal{R}$ such that $v_i \in \mathcal{C}_i$.
\end{enumerate}

The objective of the CVRP is to minimize the overall distance $cost(R)$ traveled by all $k$ vehicles and is defined as:
\begin{eqnarray} \label{equation:objectivefunction}
cost(R) &=& \sum_{i=1}^{k} c(\mathcal{C}_i)
\end{eqnarray}
where $c(\mathcal{C}_i)$ is the summation of the travel distance $e_{ij}$ contained in route $\mathcal{C}_i$.

\subsection{Capacitated Arc Routing Problem}
The capacitated arc routing problem (CARP) was first proposed by Golden and Wong \cite{BR81} in 1981. It can be formally stated as follows: Given a connected undirected graph $G = (V, E)$, where vertex set $V = \{v_i\}, i=1 \ldots n$, $n$ is the number of vertices, edge set $E = \{e_{i}\}, i=1 \ldots m$ with $m$ denoting the number of edges. Consider a demand set $D = \{d(e_i) | e_i \in E\}$, where $d(e_i) > 0$ implies edge $e_i$ requires servicing (i.e., known as task), a travel cost vector $C_t = \{c_t(e_i) | e_i \in E\}$ with $c_t(e_i)$ representing the cost of traveling on edge $e_i$, a service cost vector $C_s = \{c_s(e_i) | e_i \in E\}$ with $c_s(e_i)$ representing the cost of servicing on edge $e_i$. A solution of CARP can be represented as a set of  travel circuits $\mathcal{R} = \{\mathcal{C}_i\},~i=1 \ldots k$ which satisfies the following constraints:

\begin{enumerate}
\item Each travel circuit $\mathcal{C}_i,~i\in [1,k]$ must start and end at the depot node $v_d \in V$.
\item The total load of each travel circuit must be no more than the capacity $W$ of each vehicle, $\sum_{\forall e_i \in \mathcal{C}} d(e_i) \le W$.
\item $\forall e_i \in E~and~d(e_i) > 0$, there exists one and only one circuit $\mathcal{C}_i \in \mathcal{R}$ such that $e_i \in \mathcal{C}_i$.
\end{enumerate}

The cost of a travel circuit is then defined by the total service cost for all edges that needed service together with the total travel cost of the remaining edges that formed the circuit:
\begin{equation} \label{cst}
cost(\mathcal{C}) = \sum_{e_i \in \mathcal{C}_s} c_{s}(e_i) + \sum_{e_i \in \mathcal{C}_t} c_{t}(e_i)
\end{equation}
where $\mathcal{C}_s$ and $\mathcal{C}_t$ are edge sets that required servicing and those that do not, respectively. And the objective of CARP is then to find a valid solution $\mathcal{R}$ that minimizes the total cost:
\begin{equation} \label{Tcst}
C_\mathcal{R} = \sum_{\forall \mathcal{C}_i \in \mathcal{R}} cost(\mathcal{C}_i)
\end{equation}

\subsection{Present State-of-the-art Evolutionary Optimization Methodologies}
Theoretically, CVRP and CARP have been proven to be NP-hard with only explicit enumeration approaches known to solve them optimally. However, large scale problems are generally computationally intractable due to the poor scalability of most enumeration methods \cite{NAP81}. From a survey of the literature, metaheuristics, heuristics and evolutionary computation have played important roles in algorithms capable of providing good solutions within tractable computational time. For CVRP, Cordeau \emph{et al.} \cite{JFGA01} considered a unified tabu search algorithm (UTSA) for solving VRP. Prins \cite{CP04} presented an effective evolutionary algorithm with local search for the CVRP, while Reimann \emph{et al.} \cite{MKR04} proposed a D-ants algorithm for CVRP which equipped ant colony algorithm with individual learning procedure. Recently, Lin \emph{et al.} \cite{SZKC09} takes the advantages of both simulated annealing and tabu search, and proposed a hybrid meta-heuristic algorithm for solving CVRP. Further, Chen \emph{et al.} \cite{XYMS11} proposed a domain-specific cooperative memetic algorithm for solving CVRP and achieved better or competitive results compared with a number of state-of-the-art memetic algorithms and meta-heuristics to date.

On the other hand, for CARP, Lacomme et al. in \cite{PC04} presented the basic components that have been embedded into memetic algorithms (MAs) for solving the extended version of CARP (ECARP). Lacomme's MA (LMA) was demonstrated to outperform all known heuristics on three sets of benchmarks. Recently, Mei et al. \cite{YK09} extended Lacomme's work by introducing two new local search methods, which successfully improved the solution qualities of LMA. In a separate study, a memetic algorithm with extended neighborhood search was also proposed for CARP in \cite{KY09}. Further, Liang \emph{et al.} proposed a formal probabilistic memetic algorithm for solving CARP, with new best-known solutions to date found on $9$ of the benchmark problems \cite{LYQA10}.

\begin{figure*}[ht]
    \centering
    \includegraphics[width=0.8\textwidth]{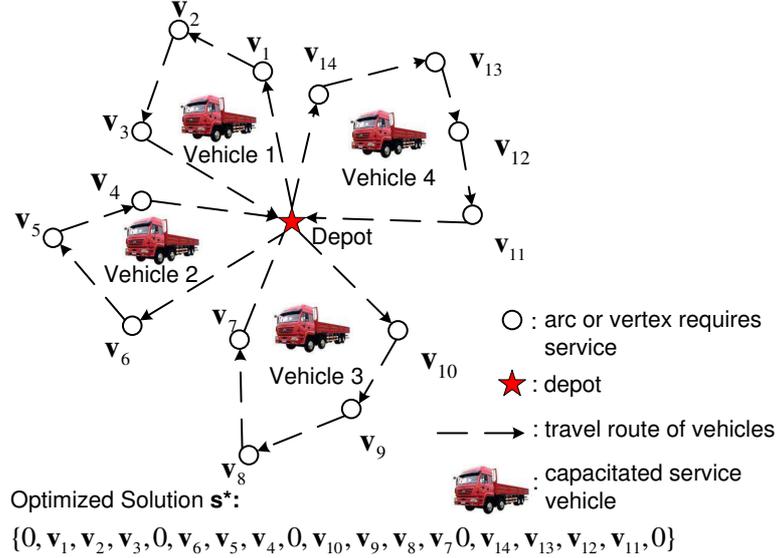}
    \caption{An example of the CARP and CVRP.}
    \label{carp}
\end{figure*}

The problem of solving CVRP and CARP can be generalized as searching for the suitable task assignments (i.e., vertices or arcs that require to be serviced) of each vehicle, and then finding the optimal service order of each vehicle for the assigned tasks. In the evolutionary search literature, the task assignment stage has been realized by means of simple task randomization \cite{KYX09} to more advance strategies like heuristic search \cite{KYX09, XYMS11}, clustering \cite{NY99}, etc., while the optimal service order of each vehicle is achieved via the mechanisms of evolutionary search operators. The example of an optimized CVRP or CARP solution can be illustrated in Fig. \ref{carp}, where four vehicle routes, namely, $R_1 = \{0, \mathbf{v}_1, \mathbf{v}_2, \mathbf{v}_3, 0\}$, $R_2 = \{0, \mathbf{v}_6, \mathbf{v}_5, \mathbf{v}_4, 0\}$, $R_3 = \{0, \mathbf{v}_{10}, \mathbf{v}_9, \mathbf{v}_8, \mathbf{v}_7, 0\}$ and $R_4 = \{0, \mathbf{v}_{14}, \mathbf{v}_{13}, \mathbf{v}_{12}, \mathbf{v}_{11}, 0\}$, can be observed. A `$0$' index value is assigned at the beginning and end of route to denote that each route starts and ends at the depot.

In the rest of the paper, we shall illustrate the concept of memetic evolution, which learns and extracts useful traits from past problem solving experiences as memes or structured knowledge, for ``intelligent'' evolutionary optimization of real world CVRPs and CARPs. In particular, we model the mapping of solved CVRP or CARP instances to their optimized solutions as memes, which serves as the instructions for carrying out the behavior that shall subsequently be used to appropriately bias the search on new unseen CVRP or CARP instances, respectively.

\section{Memetic Computational Paradigm for ``Intelligent'' Evolutionary Optimization of Routing Problems}\label{CPIEO}
In this section, we present a realization of the proposed memetic computational search paradigm, particularly focusing on the cultural-inspired operators, namely, \emph{meme learning}, \emph{meme selection}, \emph{meme variation} and \emph{meme imitation}, which are designed for solving routing problems (i.e., such as CARPs and CVRPs) and other related problem domains. The ``intelligent'' evolutionary optimization process of routing problem instances using the mechanisms of memetic evolution is next described.

\begin{figure*}[ht]
    \centering
    \includegraphics[width=0.8\textwidth]{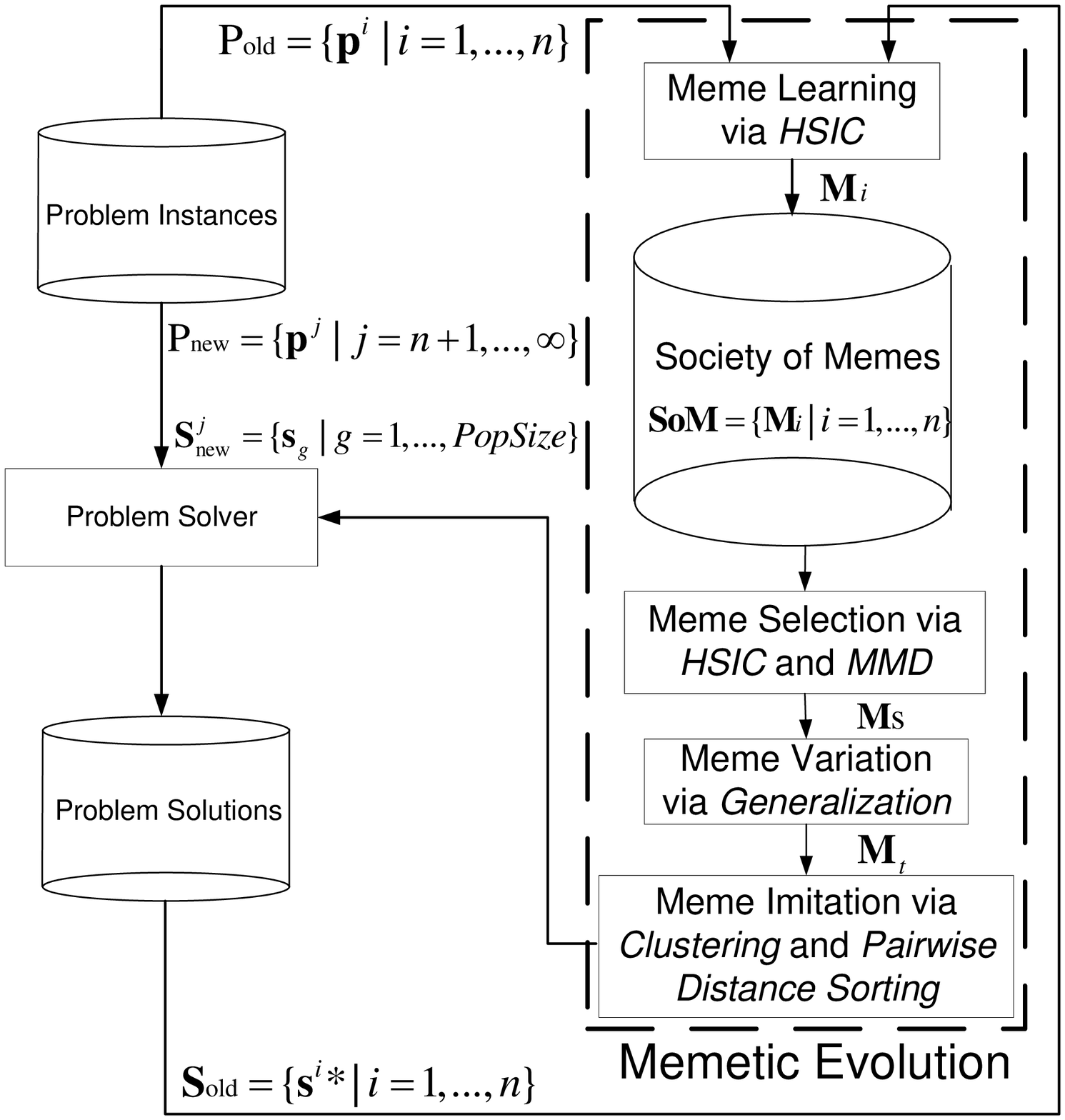}
    \caption{Memetic Computational Paradigm for ``Intelligent'' Evolutionary Optimization of Problems.}
    \label{AFlow11}
\end{figure*}

In particular, the pseudo-code and detailed workflow of the proposed memetic computational paradigm for evolutionary optimization of routing problem instances are outlined in Algo. \ref{alg1} and Fig. \ref{AFlow11}, respectively. For a given new routing problem instance $\mathbf{p}^j_{new}$ (with data representation $\mathbf{X}^j_{new}$) posed to evolutionary solver $OS$, the mechanisms of the \emph{meme selection} operator kicks in to select the fit memes $\mathbf{M}$s to activate, if the $\mathbf{SoM}$ meme pool is not empty. \emph{Meme variation}, which takes inspirations from the human's ability to generalize from past knowledge learned in previous problem solving experiences, then operates to generalize the activated $\mathbf{M}$s to arrive at $\mathbf{M_t}$. Subsequently, for given new problem instance $\mathbf{X}^j_{new}$, \emph{meme imitation} then proceeds to positively bias the search of evolutionary optimization solver $OS$, with meme (i.e., generalized $\mathbf{M_t}$) induced solutions that enhance the search performances on $\mathbf{p}^j_{new}$. In our work, the meme induced solutions of tasks assignment and service orders are obtained by means of simple clustering and pairwise distance sorting ($PDS$), respectively, which shall be detailed later in Section \ref{MIOIRP}. When the search on $\mathbf{p}^{j}_{new}$ completes, the problem instance $\mathbf{p}^{j}_{new}$ together with the obtained optimized solution $\mathbf{s}^{j*}_{new}$ of $OS$, i.e., ($\mathbf{X}^j_{new}$, $\mathbf{Y}^j_{new}$), shall then undergo the \emph{meme learning} operation so as to update the $\mathbf{SoM}$ meme pool of the intelligent $OS$ evolutionary solver.
\begin{algorithm}

    \SetKwInOut{Input}{Input} \SetKwInOut{Output}{Output}
    \SetKwInOut{Define}{Define}\SetKwInOut{Procedure}{Algorithm 1}

\textbf{Begin}:\\

\For{$j=1:\infty$ new problem instances $\mathbf{p}^j_{new}$ or $\mathbf{X}^j_{new}$}{
\If {$\mathbf{SoM} != \emptyset$}{ /*meme pool not empty*/\\
\textbf{Perform} \emph{meme selection} to identify fit memes $\mathbf{M}$s $\in\mathbf{SoM}$ /*see Eqn. \ref{sel2}* in later Section \ref{MSVORP}/\\
\vspace{2mm}
\textbf{Perform} \emph{meme variation} to generalize $\mathbf{M}$s to derive $\mathbf{M_t}$. \\
\vspace{2mm}
\textbf{Empty} the initial population $\Omega$.\\
\vspace{2mm}
\For{$g=1:Population~Size$}{
\textbf{Perform} \emph{meme imitation} with $\mathbf{X}^j_{new}$ and $\mathbf{M_t}$ to obtain solution $\mathbf{s}_g$.\\
\hspace{2mm} a. $\mathbf{X}^{j'}_{new} = Transform(\mathbf{X}^j_{new},\mathbf{M_t})$\\
\hspace{2mm}\hspace{2mm}/*Fig. \ref{imitat}(a)$\rightarrow$Fig. \ref{imitat}(b)*/ \\
\hspace{2mm} b. $Task~Assignment~of~\mathbf{s}_g =$\\
\hspace{2mm}\hspace{2mm}\hspace{2mm} $KMeans(\mathbf{X}^{j'}_{new}, Vehicle~No., RI)$\\
\hspace{2mm}\hspace{2mm}/Fig. \ref{imitat}(b)$\rightarrow$Fig. \ref{imitat}(c), $RI$ denotes random\\
\hspace{2mm}\hspace{2mm}initial points*/\\
\hspace{2mm} c. $Service~Order~of~\mathbf{s}_g = PDS(\mathbf{X}^{j'}_{new})$\\
\hspace{2mm}\hspace{2mm}/*Fig. \ref{imitat}(c)$\rightarrow$Fig. \ref{imitat}(d), $PDS(\cdot)$ denotes the\\
\hspace{2mm}\hspace{2mm}pairwise distance sorting*/\\
\hspace{2mm} d. Insert $\mathbf{s}_g$ into $\Omega$.\\
}
}
\Else{
\textbf{Proceed} with the original population initialization scheme of the evolutionary solver $OS$.\\
}
/*Start of Evolutionary Solver $OS$ Search*/\\
\textbf{Perform} reproduction and selection operations of $OS$ with generated population $\mathbf{s}_g$s until the predefined stopping criteria are satisfied.\\
/*End of Evolutionary Solver OS Search*/\\
\vspace{2mm}
\textbf{Perform} \emph{meme learning} on given $\mathbf{p}^j_{new}$ and corresponding optimized solution $\mathbf{s}^{j*}_{new}$ denoted by ($\mathbf{X}^j_{new},\mathbf{Y}^j_{new}$), attained by $OS$ evolutionary solver to derive meme $\mathbf{M}^j_{new}$.\\
\vspace{2mm}
\textbf{Archive} the learned meme of $\mathbf{p}^j_{new}$ into $\mathbf{SoM}$ meme pool to update the society of meme for subsequent reuse. \\
}
\textbf{End}\\
\caption{Pseudo code of the Memetic Computational Search Paradigm for ``Intelligent'' Evolutionary Optimization of Routing Problems}
\label{alg1}
\end{algorithm}
\subsection{Meme Learning Operator in Routing Problem} \label{MLOCC}
\begin{figure*}[ht]
    \centering
    \includegraphics[width=0.8\textwidth]{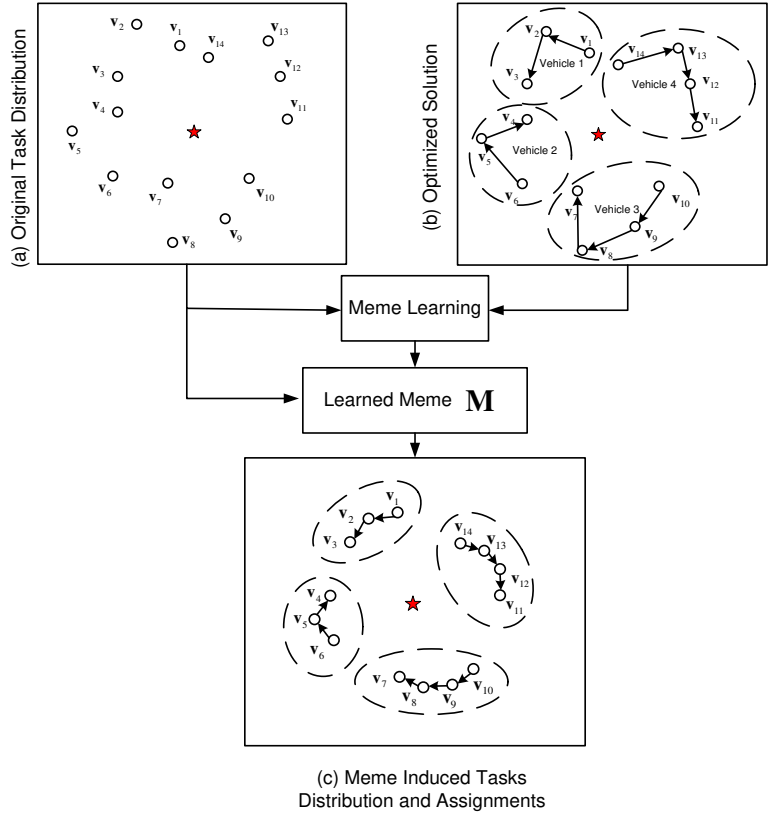}
    \caption{Meme as the instructions for ``intelligent'' tasks assignment and ordering of routing problems.}
    \label{imeme}
\end{figure*}

This subsection describes the learning and extraction of memes as building blocks of useful traits from given routing problem instances $\mathbf{p}$ and the corresponding optimized solutions $\mathbf{s^*}$ (i.e., Line $22$ in Alg. \ref{alg1}). To begin, we refer to Fig. \ref{carp} and Fig. \ref{imeme}, which shall serve as the example problem instance used for our illustrations. Fig. \ref{imeme}(a) depicts the distribution of the tasks in the example routing problem of Fig. \ref{carp} that need to be serviced. Fig. \ref{imeme}(b), on the other hand, denotes the optimized routing solution of the $OS$ evolutionary solver on problem Fig. \ref{carp} and Fig. \ref{imeme}(a). The dashed circles in Fig. \ref{imeme}(b) denote the optimized task assignments of the individual vehicles, and the arrows then indicate the optimized tasks service orders by $OS$.

Here a meme $\mathbf{M}$ thus hosts the instructions for transforming the original distribution and service orders of tasks, to one that realigns well or is closer to the desired tasks distribution and tasks service orders, as defined by the optimized routing solution $\mathbf{s}^*$ attained by solver $OS$. Using the example routing problem instance in Fig. \ref{carp}, our objective is thus to find the meme $\mathbf{M}$ that transforms the task distributions depicted in Fig. \ref{imeme}(a), to the desired tasks distribution of $\mathbf{s}^*$ and preserves the corresponding tasks service orders as depicted in Fig. \ref{imeme}(b). In this manner, whenever a new routing problem instance is encountered, suitable learned or captured memes from previously optimized problem instances can then be deployed to realign the tasks distribution and service orders accordingly. For instance, Fig. \ref{imeme}(c) showcases the desirable scaled or transformed tasks distribution of Fig. \ref{imeme}(a) when the appropriate meme $\mathbf{M}$ is put to work. In particular, it can be observed in Fig. \ref{imeme}(c) that we seek for meme(s) capable of re-locating tasks serviced by a common vehicle to become closer to one another (as desired by the optimized solution $\mathbf{s}^*$ shown in Fig. \ref{imeme}(b)), while tasks serviced by different vehicles are kept further apart. In addition, to match the service orders of each vehicle to that of the optimized solution $\mathbf{s}^*$, the task distribution is adjusted according to the sorted pairwise distances in ascending order (e.g., the distance between $v_1$ and $v_3$ is the largest among $v_1, v_2$ and $v_3$, while the distance between $v_{10}$ and $v_9$ is smaller than that of $v_{10}$ and $v_8$).

Next, the mathematical definitions of meme $\mathbf{M}$ to achieve the transformations of tasks distribution are detailed. In particular, given $\mathbf{V} = \{\mathbf{v}_i\ | i=1, \ldots, n\}$, $n$ is the number of tasks, denoting the tasks of a problem instance to be assigned. The distance between any two tasks $\mathbf{v}_i = (v_{i1},\ldots, v_{ip})^T$ and $\mathbf{v}_j = (v_{j1},\ldots, v_{jp})^T$ in the $p$-dimensional space $\mathbb{R}^p$ is then given by:

\begin{equation} \label{MM}
d_M(\mathbf{v}_{i}, \mathbf{v}_{j}) = ||\mathbf{v}_{i} - \mathbf{v}_{j}||_M = \sqrt{(\mathbf{v}_{i} - \mathbf{v}_{j})^T\mathbf{M}(\mathbf{v}_{i} - \mathbf{v}_{j})} \nonumber
\end{equation}

where $T$ denotes the transpose of a matrix or vector. Meme $\mathbf{M}$ is positive semidefinite, and can be decomposed as $\mathbf{M} = \mathbf{L}\mathbf{L}^T$ by means of singular value decomposition (SVD). By substituting this decomposition into $d_M(\mathbf{v}_{i}, \mathbf{v}_{j})$, we arrive at:
\begin{eqnarray} \label{simL}
d_M(\mathbf{v}_{i}, \mathbf{v}_{j}) = \sqrt{(\mathbf{L}^T\mathbf{v}_{i} - \mathbf{L}^T\mathbf{v}_{j})^T(\mathbf{L}^T\mathbf{v}_{i} - \mathbf{L}^T\mathbf{v}_{j})}
\end{eqnarray}

From equation \ref{simL}, it is worth noting that the distances among the tasks are scaled by meme $\mathbf{M}$. Such a meme $\mathbf{M}$ shall then perform the realignment of tasks distribution and service orders of a given new problem instance to be closer to the optimized solution $\mathbf{s^*}$ of a related problem instance that has been previously encountered.

\begin{figure*}[ht]
    \centering
    \includegraphics[width=0.8\textwidth]{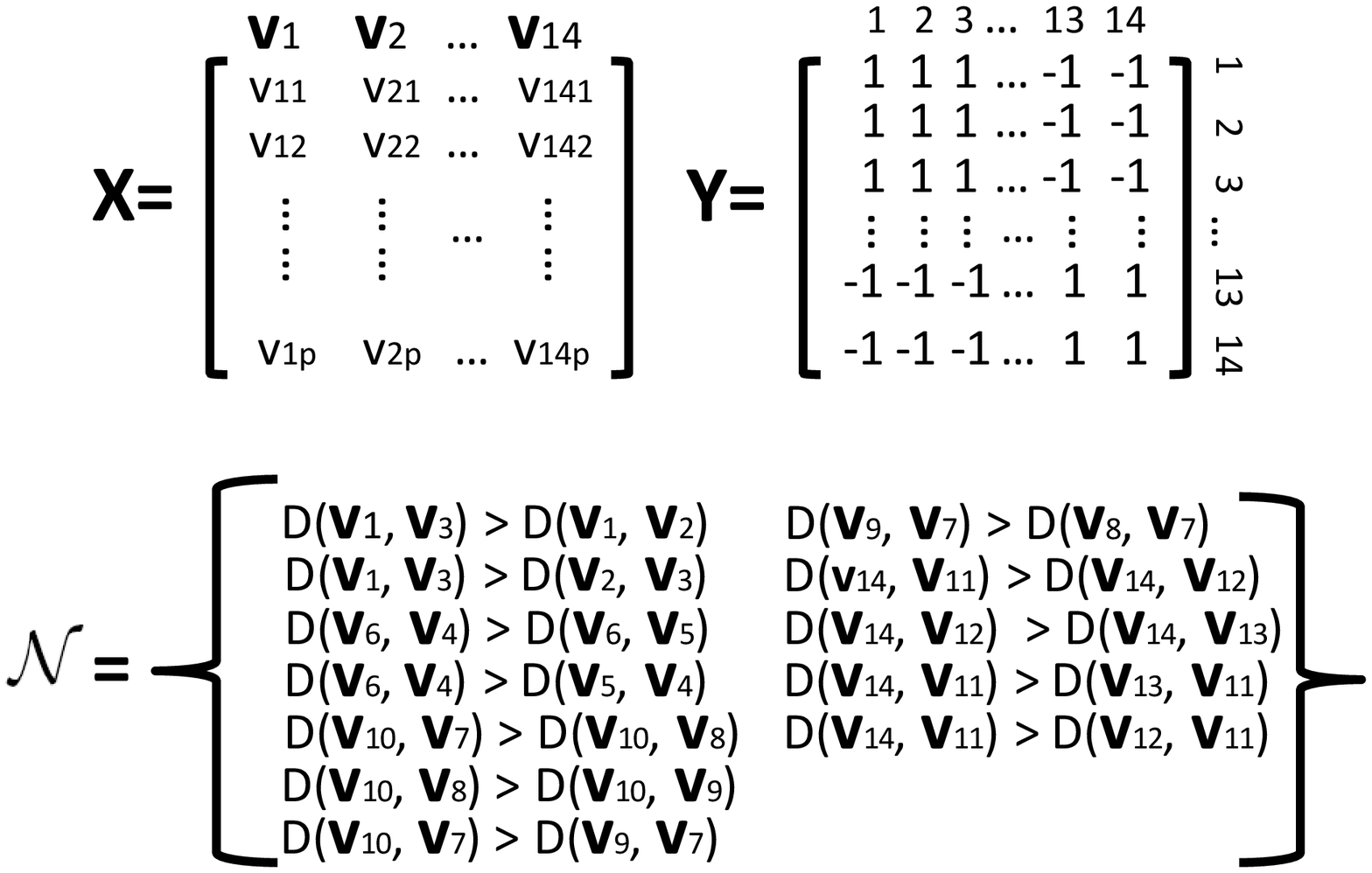}
    \caption{Data representations of a problem instance $\mathbf{p=X}$, the corresponding optimized solution $\mathbf{s^*=Y}$ and distance constraints set $\mathcal{N}$.}
    \label{illM}
\end{figure*}

In what follows, the mathematical formulations for the automated learning of meme $\mathbf{M}$ are detailed. The schemata representations of a problem instance ($\mathbf{p}$), optimized solution ($\mathbf{s^*}$) and distance constraints set $\mathcal{N}$ are first defined. In particular, the data representations of the example problem instance in Fig. \ref{carp} is depicted in Fig. \ref{illM}, where $v_{11}$, $v_{12}$, etc., denote the features representation of each task, and $D(\cdot)$ indicates the Euclidean distance metric. Further, if task $\mathbf{v}_i$ and task $\mathbf{v}_j$ are served by the same vehicle, $\mathbf{Y}(i,j) = 1$, otherwise, $\mathbf{Y}(i,j) = -1$.

To capture meme $\mathbf{M}$ from a given CVRP or CARP problem instance, denoted by ($\mathbf{p}$, $\mathbf{s^*}$), we formulate the learning task as a maximization of the dependency between $\mathbf{X}$ and $\mathbf{Y}$ according to the \emph{Hilbert-Schmidt Independence Criterion} (HSIC) \cite{Gretton05measuringstatistical} with distance constraints, which is mathematically defined as:
\begin{eqnarray}\label{f1}
\max_{\mathbf{K}} && tr( \mathbf{H} \mathbf{K} \mathbf{H} \mathbf{Y}) \\
\textrm{s.t.} && \mathbf{K} = \mathbf{X}^T*\mathbf{M}*\mathbf{X} \nonumber\\
&& D_{ij}>D_{iq}, \forall(i,j,q)\in \mathcal{N} \nonumber\\
&& \mathbf{K} \succeq 0 \nonumber
\end{eqnarray}
where $tr(\cdot)$ denotes the trace operation of a matrix. $\mathbf{X}$, $\mathbf{Y}$ are the matrix representations of a CARP or CVRP instance $\mathbf{p}$ and the corresponding problem solution $\mathbf{s^*}$, respectively.

Further, $\mathbf{H} = \mathbf{I} - \frac{1}{n}\mathbf{I}\mathbf{I}'$ centers the data and the labels in the feature space, $\mathbf{I}$ denotes the identity matrix, $n$ equals to the number of tasks. $D_{ij}>D_{iq}$ is the constraint to impose that upon serving task $i$, task $q$ is served before task $j$ by the same vehicle.


Let $\mathbf{T}_{ij}$ denotes a $n\times n$ matrix that takes non-zeros at $T_{ii} = T_{jj} = 1$, $T_{ij} = T_{ji} = -1$. The distance constraints $D_{ij}>D_{iq}$ in Equation \ref{f1} can be reformulated as $tr(\mathbf{KT}_{ij})>tr(\mathbf{KT}_{iq})$. Further, slack variables $\xi_{ijq}$ are introduced to measure the violations of distance constraints and penalize the corresponding square loss. Consequently, by substituting the constraints into Equation \ref{f1}, we arrive at:
\begin{eqnarray} \label{f2}
\min_{\mathbf{M}, \xi} && -tr( \mathbf{X} \mathbf{H} \mathbf{Y}\mathbf{H}\mathbf{X}^T \mathbf{M} ) + \frac{C}{2}\sum\xi^2_{ijq} \\
\textrm{s.t.} && \mathbf{M} \succeq 0 \nonumber \\ \label{const1}
&& tr(\mathbf{X}^T\mathbf{MXT}_{ij})>tr(\mathbf{X}^T\mathbf{MXT}_{iq}), \forall(i,j,q)\in \mathcal{N} \nonumber
\end{eqnarray}
where $C$ balances between the two parts of the criterion. The first constraint enforces the learnt meme denoted by matrix $\mathbf{M}$ to be positive semi-definite, while the second constraint imposes the scaled distances among the tasks to align well with the desired service orders of the optimized solution $\mathbf{s^*}$ (i.e., $\mathbf{Y}$).


To solve the learning problem in Equation \ref{f2}, we first derive the minimax optimization problem by introducing dual variables $\alpha$ for the inequality constraints based on Lagrangian theory.
\begin{eqnarray}\label{eqn1}
Lr &=&tr(-\mathbf{H}\mathbf{X}^T\mathbf{MXHY}) + \frac{C}{2}\sum\xi^2_{ij}  \nonumber\\
& &- \sum\alpha_{ij}(tr(\mathbf{X}^T\mathbf{MXT}_{ij}) - tr(\mathbf{X}^T\mathbf{MXT}_{iq}))
\end{eqnarray}
Set $\frac{\partial Lr}{\partial \xi_{ij}} = 0$, we have:
\begin{equation} \label{eqn2}
C\sum\xi_{ijq} - \sum\alpha_{ijq} = 0
\Longrightarrow \xi_{ijq} = \frac{1}{C}\sum\alpha_{ijq}
\end{equation}
By substituting Equation \ref{eqn2} into Equation \ref{eqn1}, we reformulate the learning problem in Equation \ref{f2} as a minimax optimization problem, which is given by:
\begin{eqnarray}\label{eqn3}
\max_{\alpha}\min_\mathbf{M}&&tr[(-\mathbf{XHYHX}^T - \sum\alpha_{ijq}\mathbf{XT}_{ij}\mathbf{X}^T \nonumber\\
& & + \sum\alpha_{ijq}\mathbf{XT}_{iq}\mathbf{X}^T )\mathbf{M}] - \frac{1}{2C}\sum\alpha^2_{ijq} \\
\textrm{s.t.} && \mathbf{M} \succeq 0 \nonumber
\end{eqnarray}

By setting
\begin{equation}
\mathbf{A} = \mathbf{XHYHX}^T + \sum\alpha_{ijq}\mathbf{XT}_{ij}\mathbf{X}^T - \sum\alpha_{ijq}\mathbf{XT}_{iq}\mathbf{X}^T \nonumber
\end{equation}
and
\begin{equation}
\Delta J^t_{ijq} = tr[(\mathbf{XT}_{iq}\mathbf{X}^T-\mathbf{XT}_{ij}\mathbf{X}^T)\mathbf{M}] - \frac{1}{C}\alpha_{ijq}\nonumber
\end{equation}

$C$ is configured with a default value of $10$ in the learning problem of Equation \ref{eqn3}, and then Equation \ref{eqn3} can be solved as described in \cite{JIS01}.

\subsection{Meme Selection Operator in Routing Problem} \label{MSVORP}
Since different meme introduces unique biases into the evolutionary search, inappropriately chosen memes and hence biases can lead to negative impairments of the evolutionary search. To facilitate a positive transfer of memes \cite{SQ10} that would lead to enhanced evolutionary search, the \emph{meme selection} operator (i.e., Line $5$ in Alg. \ref{alg1}) is designed to select fit memes that shares common characteristics with the given new problem of interest to replicate.

Suppose there is a set of $n$ unique $\mathbf{M}$s in $\mathbf{SoM}$, i.e., $\mathbf{SoM} = \{\mathbf{M}_1, \mathbf{M}_2, \ldots, \mathbf{M}_n\}$ that form the society of memes. The \emph{meme selection} operator is here realized as to identify the weight $\mu_i$ of each meme. A fitter meme should have a higher weight and the summation of the weights of all memes equates to $1$ (i.e., $\sum_{i=1}^n\mu_i = 1$).

In particular, the meme coefficient vector $\boldsymbol{\mu}$ is determined based on the HSIC and Maximum Mean Discrepancy criteria \cite{KMAM06}.
\begin{eqnarray}\label{sel1}
\max_{\mu} && tr( \mathbf{H} \mathbf{K} \mathbf{H} \mathbf{Y}) + \sum_{i=1}^m (\mu_i)^2 Sim_i\\
\textrm{s.t.} && \mathbf{M}_t = \sum_{i=1}^n \mu_i \mathbf{M}_i  \nonumber \\
&& \mathbf{K} = \mathbf{X}^T*\mathbf{M}_t*\mathbf{X}, \mathbf{K} \succeq 0 \nonumber\\
&& \mu_i \ge 0  \nonumber \\
&& \sum_{i=1}^n \mu_i = 1 \nonumber
\end{eqnarray}

By substituting the constraints of Equation \ref{sel1} into the objective function, we arrive at:
\begin{eqnarray}\label{sel2}
\max_{\mu} && tr( \mathbf{H} \mathbf{X}^T\mathbf{M}_t\mathbf{X} \mathbf{H} \mathbf{Y}) + \sum_{i=1}^m (\mu_i)^2 Sim_i\\
\textrm{s.t.} && \mathbf{M}_t = \sum_{i=1}^n \mu_i \mathbf{M}_i, \mathbf{M}_i \succeq 0   \nonumber \\
&& \mu_i \ge 0  \nonumber \\
&& \sum_{i=1}^n \mu_i = 1\nonumber
\end{eqnarray}
where $Sim_i$ is the similarity measure between two given problem instances. In the present context, $Sim_i = -(\beta*MMD_i + (1-\beta)* Dif_i)$, where $MMD_i$ denotes the Maximum Mean Discrepancy \cite{KMAM06}, which is used to compare the distribution similarity between two given instances by measuring the distance between their corresponding means. $MMD(D_s, D_t) = ||\frac{1}{n_s}\sum_{i=1}^s\phi(x_i^s) - \frac{1}{n_t}\sum_{i=1}^t\phi(x_i^t)||$, where $\phi(\cdot)$ maps the original input to a high dimensional space. Here, for the purpose of computational efficiency, we consider a linear mapping with $\phi(\mathbf{x}) = \mathbf{x}$. $Dif_i$ denotes the difference in vehicle capacity for two given CARP or CVRP instances. $\beta$ balances between the two parts (i.e., $MMD_i$ and $Dif_i$) in $Sim_i$. From experiences, due to the higher relevance of the task distribution features than vehicles capacity when defining the similarities between problem instances, here $\beta$ is configured as $0.8$. In Equation \ref{sel1}, the first term serves to maximize the statistical dependence between input $\mathbf{X}$ and output label $\mathbf{Y}$ for clustering \cite{Song2007}. The second term measures the similarity between the previous problem instances solved to the given new problem of interest.

In Equation \ref{sel2}, two unknown variables exist (i.e., $\mu$ and $\mathbf{Y}$). $\mathbf{Y}$ is obtained from the results of task assignment. With $\mathbf{Y}$ fixed, Equation \ref{sel2} becomes a quadric programming problem of $\mu$. To solve the optimization problem of Equation \ref{sel2}, we first perform clustering (e.g., K-Means) on input $\mathbf{X}$ directly to obtain the label matrix $\mathbf{Y}$. By keeping $\mathbf{Y}$ fixed, we obtained $\mu$ by maximizing Equation \ref{sel2} via quadric programming solver. Next, by maintaining the chosen $\mathbf{M}$ fixed, clustering is made on the new $\mathbf{X}^{'}$ (i.e., transformed by selected $\mathbf{M}$. $\mathbf{X}' = \mathbf{L}^T\mathbf{X}$, where $\mathbf{L}$ is obtained by SVD on $\mathbf{M}$) to obtain label matrix $\mathbf{Y}$.

\subsection{Meme Variation Operator in Routing Problem}
Further, to introduce innovations into the selected fit memes during subsequent reuse, the \emph{meme variation} operator (i.e., Line $6$ in Alg. \ref{alg1}) then kicks in to operate on the selected memes. In the present context, we take inspirations of human's ability to generalize from past problem solving experiences. Hence \emph{meme variation} is realized here in the form of \emph{memes generalization}. However, it is worth noting that other alternative forms of probabilistic scheme in meme variations may also be considered since uncertainties can generate growth and variations of knowledge that we have of the world \cite{MS99}, hence leading to higher adaptivity capabilities for solving complex and non-trivial problems.

In particular, the generalized meme $\mathbf{M_t}$ in \emph{meme variation} is realized as a linear combination of the selected memes:
\begin{equation}
\mathbf{M}_t = \sum_{i=1}^n \mu_i \mathbf{M}_i, (\sum_{i=1}^n \mu_i = 1, \mu_i \in [0, 1]) \nonumber
\end{equation}

\begin{figure*}[ht]
    \centering
    \includegraphics[width=0.8\textwidth]{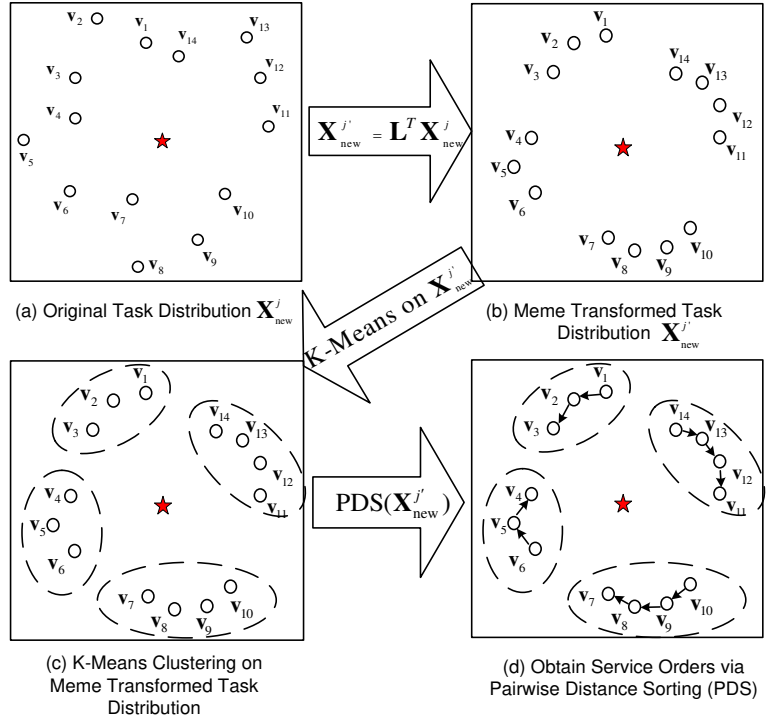}
    \caption{Illustration of meme imitation for generating intelligently biased CVRP or CARP solutions.}
    \label{imitat}
\end{figure*}
\subsection{Meme Imitation Operator in Routing Problem}\label{MIOIRP}
In many routing problems, such as CVRP and CARP, the search for optimal solution is typically solved as two separate phases. The first phase involves the assignment of the tasks that require services to the appropriate vehicles. The second phase then serves to find the optimal service order of each vehicle for the assigned tasks obtained in phase $1$.
\newsavebox{\tablebox}
\begin{lrbox}{\tablebox}
\begin{tabular}{|c|l|}
\hline
$\bf{Criterion}$ & $~~~~~~~~~~~~~~~~~~~~~~~~~~~~~~~~~~~~~~~~~~\bf{Definition}$ \\
 \hline
 $Number~of~Fitness~Evaluation$& Average number of fitness evaluations across all independent runs conducted\\
  \hline
 $Cpu~Time$& Average computational cost in wall-clock time across all independent runs conducted\\
 \hline
 $Ave.Cost$ & Average fitness of the solutions obtained across all independent runs conducted\\
 \hline
 $B.Cost$& Best fitness of the solutions obtained across all independent runs conducted\\
 \hline
 $Std.Dev$& Standard deviation of the solutions' fitness across all independent runs conducted\\
 \hline
 $Success~No.$& Number of independent runs that fare above the $Ave.Cost$ value\\
 \hline
\end{tabular}
\end{lrbox}
\begin{table*}
\centering \small \caption{Criteria for measuring performance.}\label{mea} \scalebox{0.65}{\usebox{\tablebox}}
\end{table*}
\begin{lrbox}{\tablebox}
\begin{tabular}{c|c c c c c c c c c c c c}
\hline\hline
Data Set &A-n32-k5& A-n54-k7 & A-n60-k9 & A-n69-k9  & A-n80-k10  & B-n41-k6 &B-n57-k7 & B-n63-k10  & B-n68-k9& B-n78-k10 & P-n50-k7 & P-n76-k5 \\
\hline
 $V$ & $31$ & $53$ & $59$ & $68$ & $79$& $40$ & $56$ & $62$ & $67$ & $77$ & $49$ & $75$ \\
 \hline
 $C_v$ & $100$ & $100$ & $100$ & $100$ & $100$ & $100$ & $100$ & $100$ & $100$ & $100$ & $150$ & $280$\\
 \hline
 $LB$ &$784$ & $1167$ & $1354$ & $1159$ & $1763$ & $829$ & $1140$ & $1496$ & $1272$ & $1221$ & $554$ & $627$ \\
 \hline\hline
\end{tabular}
\end{lrbox}
\begin{table*}
\centering \small \caption{Properties of the ``Augerat'' CVRP data set.}\label{AugC} \scalebox{0.5}{\usebox{\tablebox}}
\end{table*}

In what follows, the meme induced solutions that serve as initial population to intelligently bias the $OS$ search shall be described. For each solution $\mathbf{s}_g$, the imitation of tasks assignments and tasks service orders using generalized meme $\mathbf{M}_t$ then lead to a transformation (i.e., Line $10$ in Alg. \ref{alg1}) of the original tasks distribution $\mathbf{X}^{j}_{new}$ to a meme induced tasks distribution $\mathbf{X}^{j'}_{new}$, which is given by:

\begin{eqnarray}\label{imitationT}
\mathbf{X}^{j'}_{new} = \mathbf{L}^T\mathbf{X}^{j}_{new}
\end{eqnarray}
where $\mathbf{L}$ is derived by means of singular value decomposition on $\mathbf{M}_t$. The illustrative example is depicted in Fig. \ref{imitat}, where Fig. \ref{imitat}(a) denote the original task distribution $\mathbf{X}^{j}_{new}$ and Fig. \ref{imitat}(b) is the resultant meme induced tasks distribution $\mathbf{X}^{j'}_{new}$ using $\mathbf{M}_t$.

In phase $1$, simple clustering scheme such as K-Means clustering with random initializations is then conducted on the meme induced tasks distribution $\mathbf{X}^{j'}_{new}$ to derive the tasks assignments of the vehicles as depicted in Fig. \ref{imitat}(c), where the dashed circles denote the task assignments of the individual vehicles, i.e., denoting the tasks that shall be serviced by a common vehicle.

In phase $2$, the service orders of each vehicle are subsequently achieved by sorting the pairwise distances among tasks in an ascending order. The two tasks with largest distance shall denote the first and last tasks to be serviced. Taking the first task as reference, the service order of the remaining tasks are defined according to the sorted orders. Taking Fig. \ref{imitat}(d) as example, where the arrows indicate the service orders of the tasks, the distance between $\mathbf{v}_1$ and $\mathbf{v}_3$ are largest among $\mathbf{v}_1$, $\mathbf{v}_2$ and $\mathbf{v}_3$. By assigning $\mathbf{v}_1$ as the reference task to be served, $\mathbf{v}_2$ shall be the next task to be serviced, since the distance between $\mathbf{v}_1$ and $\mathbf{v}_2$ is smaller than that of $\mathbf{v}_1$ and $\mathbf{v}_3$.

\section{Experimental Study}\label{ES}
To evaluate the efficiency and effectiveness of the proposed memetic evolutionary search paradigm, empirical studies conducted on the challenging domains of capacitated vehicle routing problems (CVRPs) and capacitated arc routing problems (CARPs) are presented in this section. These consist of problems of diverse properties in terms of vertices size, graph topologies, etc., which cannot be handled using existing case based reasoning approaches \cite{SJJM04, PB97} as previously discussed in Section \ref{RW}. Two state-of-the-art evolutionary algorithms for solving CVRPs and CARPs, labeled in their respective published works as \emph{CAMA} \cite{XY12} and \emph{ILMA} \cite{YK09}, are considered here as the baselines conventional evolutionary solver, of the respective domains, in the proposed memetic search paradigm. In this study, several criteria have been defined to measure the search performances, which are listed in Table \ref{mea}. Among these criteria, $Number~of~Fitness~Evaluation$ and $Cpu~Time$ are used to measure the efficiency of the algorithms, while $Ave.Cost$, $B.Cost$ and $Success~No.$ serve as the criteria for measuring the solution qualities of the algorithms.

\subsection{Capacitated Vehicle Routing Problem}
\subsubsection{Empirical Configuration}
All three commonly used CVRP benchmark data sets are investigated in the present empirical study, namely ``AUGERAT'' \cite{PJEA95}, ``CE'' \cite{NS69} and ``CHRISTOFIDES'' \cite{NAP79}. The ``Augerat'' dataset includes $12$ CVRP instances, ``CE'' consists of $6$ CVRP instances, and ``Christofides'' has $7$ CVRP instances. The detailed properties (e.g., number of vertices, lower bound, etc.) of the CVRP instances considered are summarized in Table \ref{AugC} and Table \ref{CEC}, where $V$ denotes the number of vertices that need to be serviced, $C_v$ gives the capacity of the vehicles in each instance, and $LB$ describes the lower bound of each problem instance. Note the diversity in the properties of the problems considered.
\begin{lrbox}{\tablebox}
\begin{tabular}{c|c c c c c c c c c c c c c}
\hline\hline
Data Set  & E-n33-k4  & E-n76-k7  & E-n76-k8 & E-n76-k10 & E-n76-k14&E-n101-k8 & c50 & c75 &c100 & c100b & c120 & c150 & c199\\
\hline
 $V$ & $32$ & $75$ & $75$ & $75$& $75$ & $100$ & $50$ & $75$ & $100$ & $100$ & $120$& $150$ & $199$\\
 \hline
 $C_v$ & $8000$  & $220$ & $180$ & $140$ & $100$ & $200$ & $160$ & $140$ & $200$ & $200$ & $200$ & $200$ & $200$\\
 \hline
 $LB$  & $835$ & $682$ & $735$ & $830$ & $1021$ &$815$ &$524.61$ & $835.26$ & $826.14$ & $819.56$ & $1042.11$ & $1028.42$ & $1291.45$\\
 \hline\hline
\end{tabular}
\end{lrbox}
\begin{table*}
\centering \small \caption{Properties of the ``CE'' and ``Christofides'' CVRP data sets.}\label{CEC} \scalebox{0.5}{\usebox{\tablebox}}
\end{table*}
\begin{lrbox}{\tablebox}
\begin{tabular}{l|c c c c| c c c c|c c c c}
\hline\hline
Data & & $CAMA$& &  & & $CAMA-R$ &  & & & $CAMA-M$&(Proposed Method)& \\
 Set & $B.Cost$& $Ave.Cost$& $Std.Dev$ &$Success~No.$ & $B.Cost$& $Ave.Cost$& $Std.Dev$ &$Success~No.$& $B.Cost$& $Ave.Cost$& $Std.Dev$ &$Success~No.$ \\
 \hline
 1.A-n32-k5 &$784$ &$748$ & $0$ &$30$& $784$& $784$ & $0$  & $30$ & $784$ & $784$ & $0$& $30$\\
 2.A-n54-k7 &$1167$ &$1169.50$ & $3.36$  &$19$& $1167$& $\mathbf{1167}$ & $0$  & $30$ & $1167$ & $\mathbf{1167}$& $0$ & $\mathbf{30}$ \\
 3.A-n60-k9 &$1354$ &$1356.73$ & $3.59$  &$17$& $1354$& $1355.20$ & $1.86$  & $21$ & $1354$ & $\mathbf{1354.4}$ & $1.22$ & $\mathbf{27}$ \\
 4.A-n69-k9 &$1159$ &$1164.17$ & $3.07$  &$16$&$1159$ &$1162.20$ & $2.41$  &$24$&$1159$ &$\mathbf{1161.43}$ & $2.37$  &$\mathbf{28}$ \\
 5.A-n80-k10 &$1763$ &$1778.73$ & $9.30$  &$9$& $1763$ & $1777.07$ & $7.94$  & $10$ & $1763$ & $\mathbf{1775.7}$ &$8.80$ & $\mathbf{16}$ \\
 6.B-n41-k6 & $829$&$\mathbf{829.30}$ & $0.47$ & $\mathbf{21}$  &$829$&  $829.93$ & $0.94$  & $10$ & $829$ & $829.53$& $0.73$ & $17$ \\
 7.B-n57-k7 &$1140$ &$1140$ & $0$  &$30$&$1140$ &$1140$ & $0$  &$30$&$1140$ &$1140$ & $0$  &$30$\\
 8.B-n63-k10 &$1537$ &$1537.27$ & $1.46$  &$29$& $\mathbf{1496}$&  $1528.77$ & $15.77$  & $\mathbf{30}$ & $\mathbf{1496}$ & $\mathbf{1525.86}$ & $17.45$ & $\mathbf{30}$\\
 9.B-n68-k9 &$1274$ &$1281.47$ & $5.56$ &$\mathbf{15}$& $1274$& $1284.80$ & $4.51$  & $7$ & $\mathbf{1273}$ & $\mathbf{1281.43}$ & $5.74$ &$\mathbf{15}$ \\
 10.B-n78-k10 &$1221$&$1226.07$ & $5.48$ & $22$& $1221$& $1226.80$  & $6.39$ & $19$ & $1221$ & $\mathbf{1224.37}$ & $3.23$ & $\mathbf{26}$\\
 11.P-n50-k7 &$554$&$556.33$ & $2.34$ & $14$& $554$& $554.93$  & $1.72$ & $23$ & $554$ & $\mathbf{554.26}$ & $1.01$ & $\mathbf{28}$\\
 12.P-n76-k5 &$627$&$630.70$ & $5.34$ & $22$& $627$& $628.87$  & $1.61$ & $25$ & $627$ & $\mathbf{628.63}$ & $1.51$ & $\mathbf{26}$\\
 \hline\hline
\end{tabular}
\end{lrbox}
\begin{table*}
\centering \small \caption{Statistic results of \emph{CAMA}, \emph{CAMA-R}, and \emph{CAMA-M} on ``AUGERAT'' CVRP benchmarks.}\label{raug} \scalebox{0.45}{\usebox{\tablebox}}
\end{table*}
\begin{lrbox}{\tablebox}
\begin{tabular}{l|c c c c| c c c c|c c c c}
\hline\hline
Data & & $CAMA$& &  & & $CAMA-R$ &  & & & $CAMA-M$ &(Proposed Method)&  \\
 Set & $B.Cost$& $Ave.Cost$& $Std.Dev$ &$Success~No.$ & $B.Cost$& $Ave.Cost$& $Std.Dev$ &$Success~No.$& $B.Cost$& $Ave.Cost$& $Std.Dev$ &$Success~No.$ \\
 \hline
 1.E-n33-k4 &$835$ &$835$ & $0$ &$30$&$835$ &$835$ & $0$ &$30$&$835$ &$835$ & $0$ &$30$\\
 2.E-n76-k7 &$682$ &$685.67$ & $2.17$  &$23$& $682$& $684.73$ & $1.31$  & $27$ & $682$ & $\mathbf{684.66}$& $1.12$ & $\mathbf{28}$ \\
 3.E-n76-k8 &$735$ &$737.57$ & $2.36$  &$18$& $735$& $737.17$ & $1.60$  & $21$ & $735$ & $\mathbf{737.06}$ & $1.91$ & $\mathbf{23}$ \\
 4.E-n76-k10 &$830$ &$837.03$ & $3.56$  &$18$&$831$ &$835.80$ & $3.19$  &$21$&$830$ &$\mathbf{834.73}$ & $2.92$  &$\mathbf{27}$ \\
 5.E-n76-k14 &$1021$ &$\mathbf{1025.67}$ & $3.48$  &$16$& $1021$ & $1026.27$ & $3.33$  & $15$ & $1021$ & $1025.80$ &$3.64$ & $\mathbf{17}$ \\
 6.E-n101-k8 & $816$&$820.63$ & $3.20$ & $17$  &$815$&  $818.97$ & $1.94$  & $26$ & $815$ & $\mathbf{818.53}$& $1.55$ & $\mathbf{27}$ \\
 \hline\hline
\end{tabular}
\end{lrbox}
\begin{table*}
\centering \small \caption{Statistic results of \emph{CAMA}, \emph{CAMA-R}, and \emph{CAMA-M} on ``CE'' CVRP benchmarks.}\label{rce} \scalebox{0.45}{\usebox{\tablebox}}
\end{table*}
\begin{lrbox}{\tablebox}
\begin{tabular}{l|c c c c| c c c c|c c c c}
\hline\hline
Data & & $CAMA$& &  & & $CAMA-R$ &  & & & $CAMA-M$ &(Proposed Method)& \\
 Set & $B.Cost$& $Ave.Cost$& $Std.Dev$ &$Success~No.$ & $B.Cost$& $Ave.Cost$& $Std.Dev$ &$Success~No.$& $B.Cost$& $Ave.Cost$& $Std.Dev$ &$Success~No.$ \\
 \hline
 1.c50 &$524.61$ &$525.45$ & $2.56$ &$27$& $524.61$& $\mathbf{524.61}$ & $0$  & $\mathbf{30}$ & $524.61$ & $525.73$ & $2.90$& $26$\\
 2.c75 &$835.26$ &$842.32$ & $4.04$  &$15$& $835.26$& $840.28$ & $3.52$  & $22$ & $835.26$ & $\mathbf{839.53}$& $3.45$ & $\mathbf{26}$ \\
 3.c100 &$826.14$ &$829.43$ & $2.39$  &$11$& $826.14$& $829.73$ & $2.08$  & $9$ & $826.14$ & $\mathbf{829.13}$ & $1.94$ & $\mathbf{12}$ \\
 4.c100b &$819.56$ &$819.56$ & $0$  &$30$&$819.56$ &$819.56$ & $0$  &$30$&$819.56$ &$819.56$ & $0$  &$\mathbf{30}$\\
 5.c120 &$1042.11$ &$1044.18$ & $2.21$  &$18$& $1042.11$ & $1043.08$ & $1.13$  & $22$ & $1042.11$ & $\mathbf{1042.83}$ &$0.94$ & $\mathbf{27}$ \\
 6.c150 & $1032.50$&$1043.27$ & $5.67$ & $13$  &$1034.19$&  $1044.73$ & $5.87$  & $12$ & $\mathbf{1030.67}$ & $\mathbf{1041.97}$& $6.27$ & $\mathbf{16}$ \\
 7.c199 &$\mathbf{1304.87}$ &$\mathbf{1321.17}$ & $5.98$  &$\mathbf{14}$& $1313.46$& $1325.48$ & $5.99$& $9$  & $1308.92$ & $1322.18$ & $7.51$ & $\mathbf{14}$\\
 \hline\hline
\end{tabular}
\end{lrbox}
\begin{table*}
\centering \small \caption{Statistic results of \emph{CAMA}, \emph{CAMA-R}, and \emph{CAMA-M} on ``CHRISTOFIDES'' CVRP benchmarks.}\label{rchr} \scalebox{0.45}{\usebox{\tablebox}}
\end{table*}

In CVRP, each task or vertex has a corresponding $coordinates$ and $demand$. Using the $coordinates$ of the vertex, the tasks assignment of each vehicle are generated using K-Means clustering. A CVRP instance is thus represented as input matrix $\mathbf{X}$, see Fig. \ref{illM}, which is composed of the coordinate features for all tasks in the problem. The desired vehicle assigned for each task (i.e., task assignment) is then given by the $OS$ optimized solution, $\mathbf{Y}$,
of the respective CVRP instances.

Besides the proposed meme induced population initialization procedure, two other commonly used initialization procedures for generating the population of solution individuals in the state-of-the-art baseline \emph{CAMA} are also investigated here to verify the efficiency and effectiveness of the proposed memetic evolutionary search paradigm. The first is the simple random approach for generating the initial population, which is labeled here as \emph{CAMA-R}. The second is the informed heuristic population initialization procedure proposed in baseline \emph{CAMA} \cite{XY12}. In particular, the initial population is a fusion of solution generated by \emph{Backward Sweep}, \emph{Saving}, and \emph{Forward Sweep} and random initialization approaches. The CAMA that employs the meme induced population initialization procedure of the proposed memetic evolution is then notated as \emph{CAMA-M}, where the population of individuals intelligently generated based on the fit memes that have been accumulated from past CVRP solving experiences via cultural evolutionary mechanisms of the \emph{meme learning}, \emph{meme selection}, \emph{meme variation} and \emph{meme imitation}. If no memes have been accumulated so far, \emph{CAMA-M} shall behave exactly like the baseline \emph{CAMA}.

Last but not the least, the operator and parameter settings of \emph{CAMA-R}, \emph{CAMA}, \emph{CAMA-M} are kept the same as that of \cite{XY12} for the purpose of fair comparison. For \emph{CAMA-M}, the MMD of Equation \ref{sel1} is augmented with the demand of each task as one of the problem feature.
\begin{figure*}
\begin{tabular}{ccc}
\includegraphics[width=0.3\textwidth]{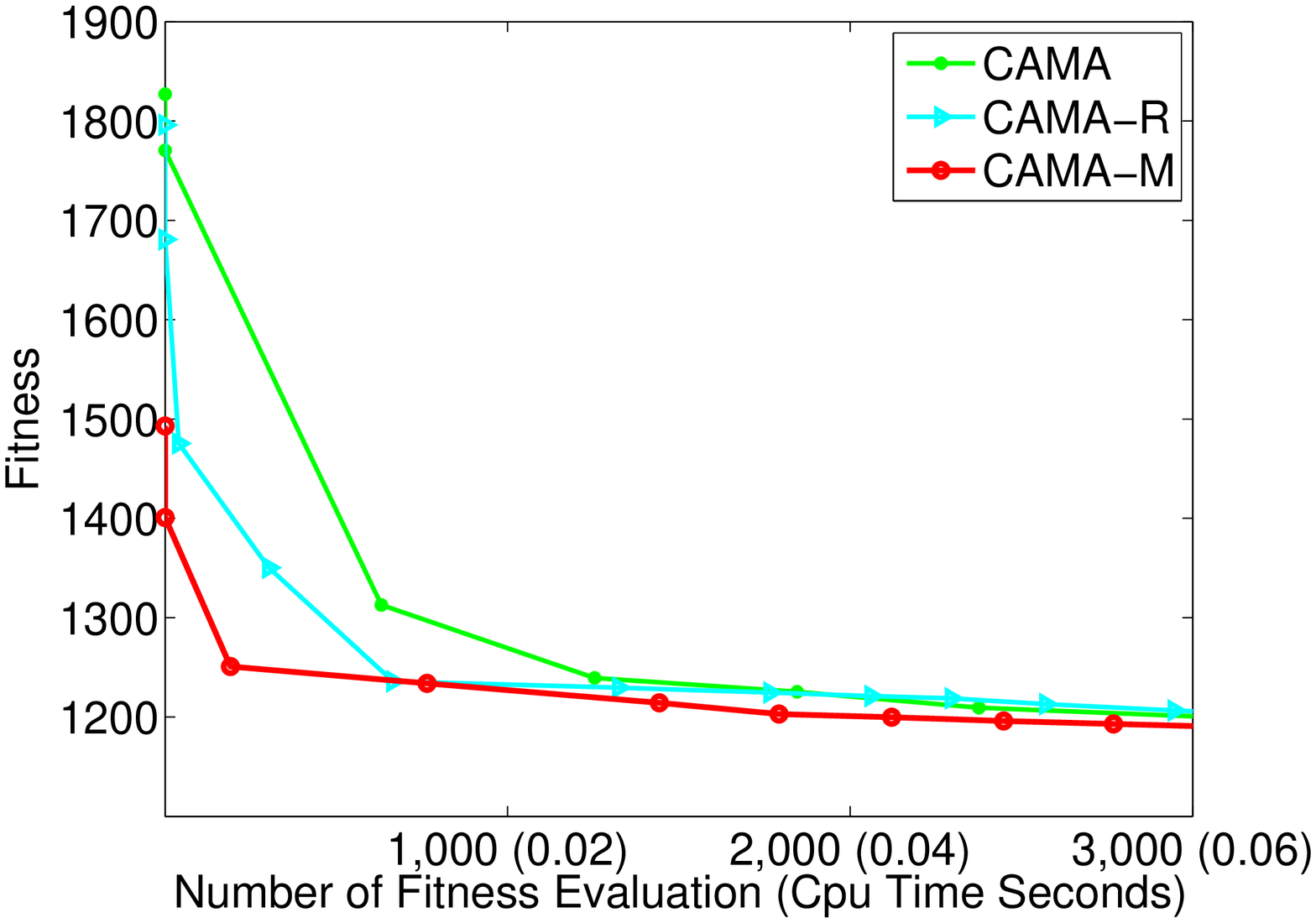} & \includegraphics[width=0.3\textwidth]{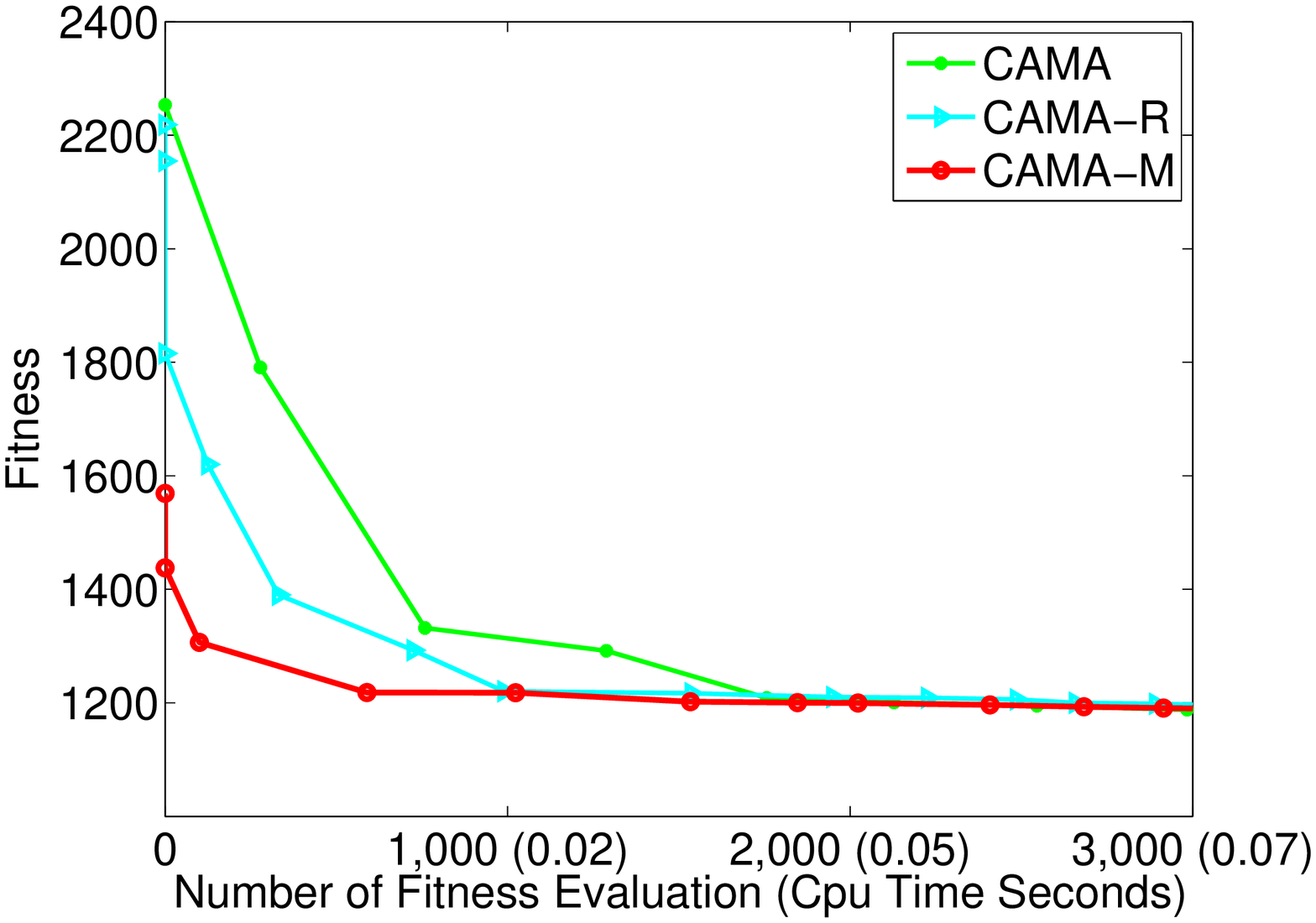} & \includegraphics[width=0.3\textwidth]{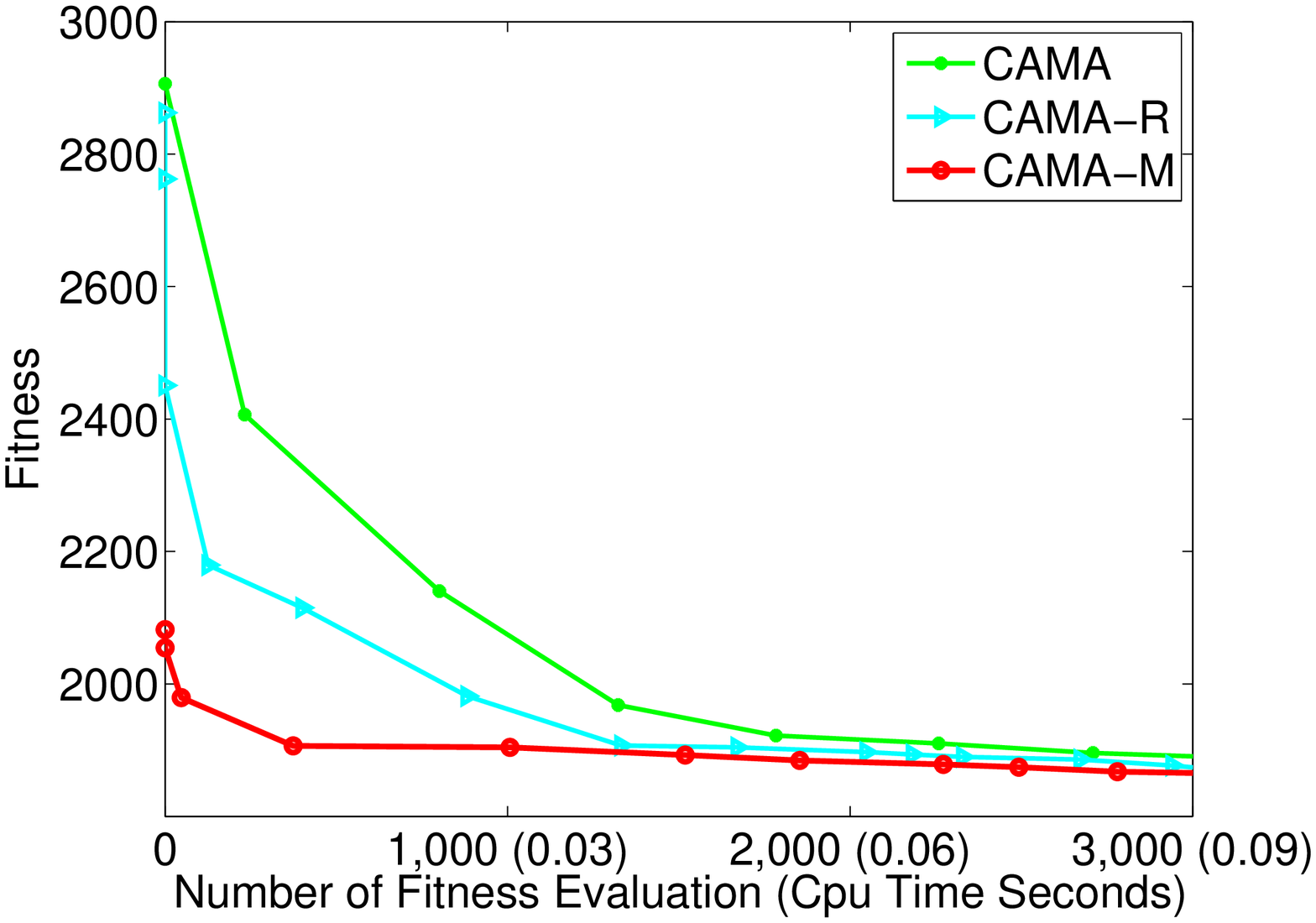} \\
(a) A-n54-k7 & (b) A-n69-k9 &(c) A-n80-k10\\
\includegraphics[width=0.3\textwidth]{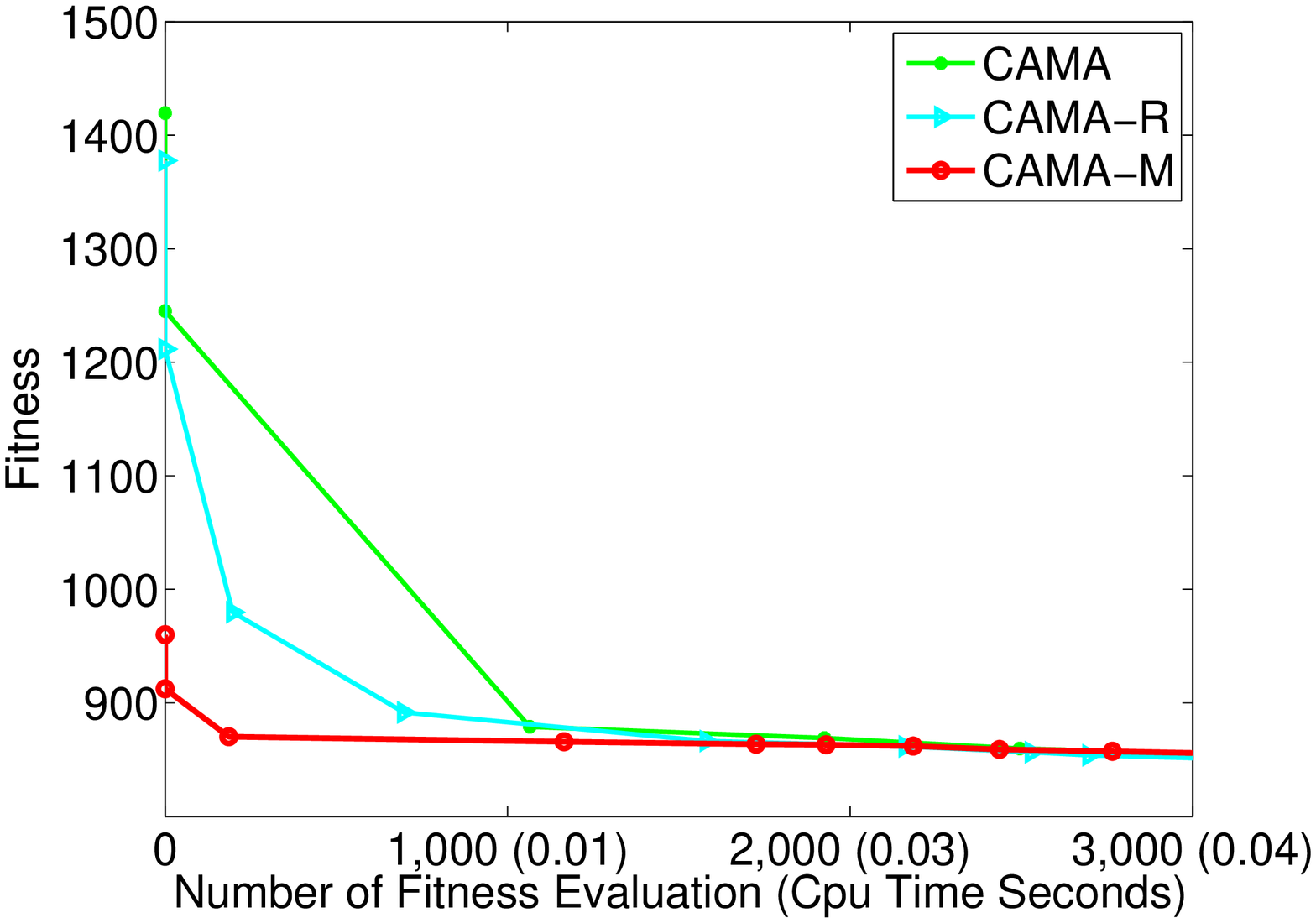} & \includegraphics[width=0.3\textwidth]{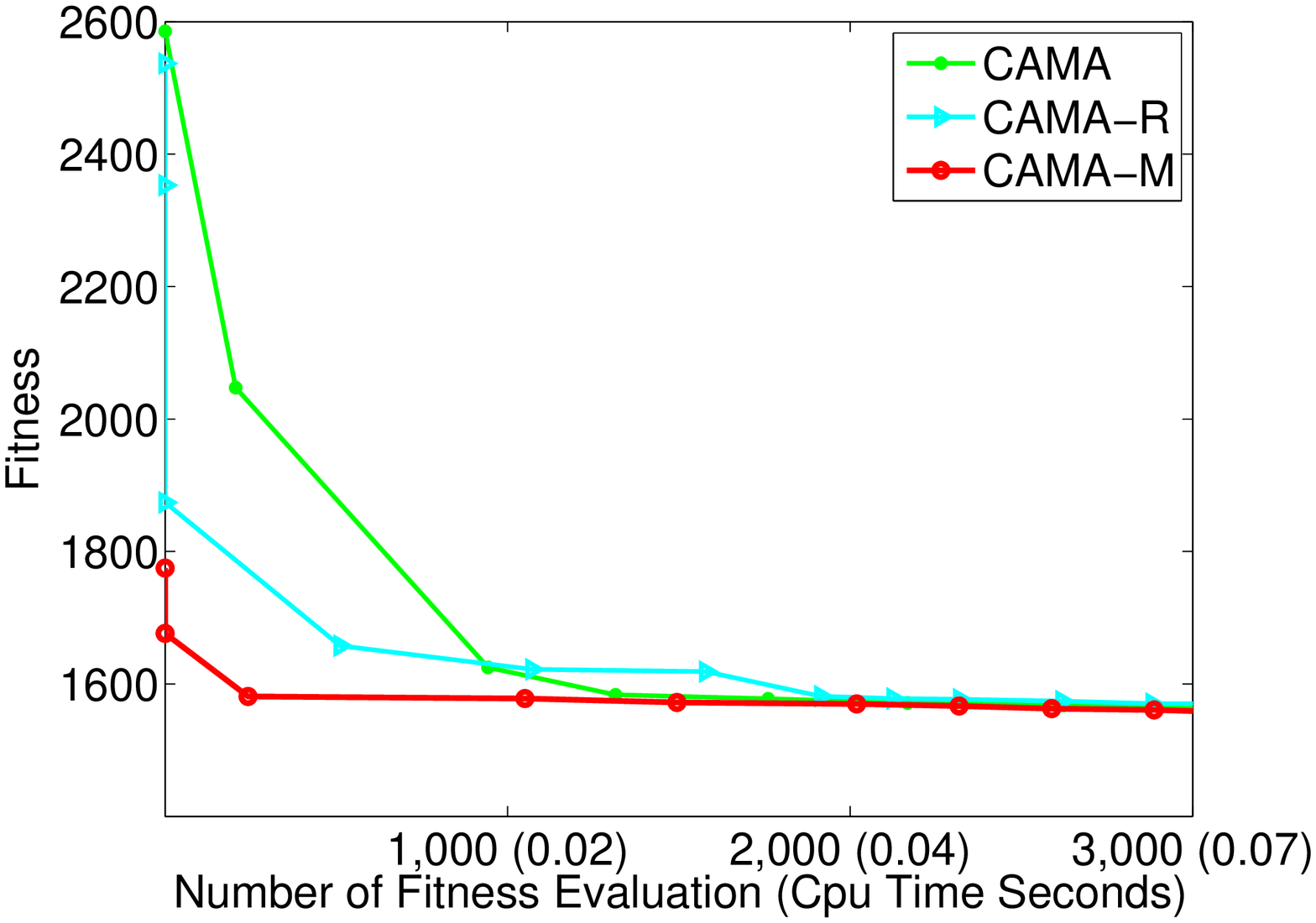} & \includegraphics[width=0.3\textwidth]{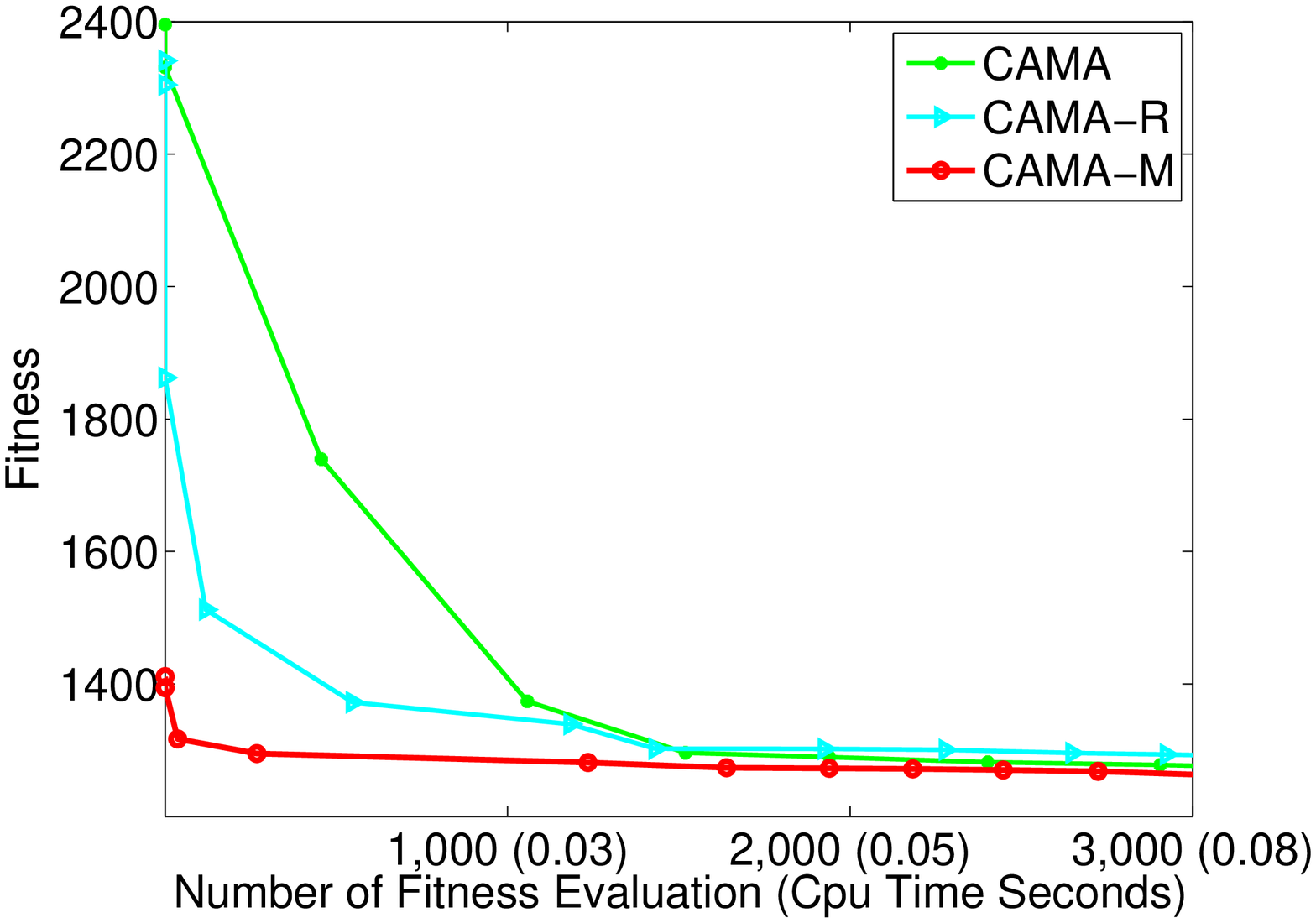} \\
(d) B-n41-k6 & (e) B-n63-k10 &(f) B-n78-k10
\end{tabular}
\caption{Search convergence graphs of $CAMA$, $CAMA-R$, and $CAMA-M$ on representative CVRP ``AUGERAT'' benchmarks. $Y$-axis: Fitness value, $X$-axis: Number of Fitness Evaluation or CPU Time in Seconds.}\label{fig:cvrponAUG}
\end{figure*}

\subsubsection{Result and Discussion}
All the results obtained by the respective algorithms considered, across $30$ independent runs on the CVRP instances, are summarized in Table \ref{raug}, Table \ref{rce} and Table \ref{rchr}. The $Ave.Cost$ value of \emph{CAMA} on each CVRP instance is used as the baseline for the $Success.No$ criterion. The values in ``$B.Cost$'', ``$Ave.Cost$'' and ``$Success~No.$'' denoting superior performance are then highlighted in bold font.

It can be observed from the results in Table \ref{raug} and Table \ref{rce} that in overall, \emph{CAMA-R} achieved improved solution quality over \emph{CAMA} in terms of $Ave.Cost$ and $Success~No.$ on most of the ``AUGERAT'' and ``CE'' CVRP instances. Particularly, on instance ``A-n54-k7'', \emph{CAMA-R} consistently converges to the best solution fitness across all the $30$ independent runs, as denoted by ``$B.Cost$''. \emph{CAMA-R} also attained an improved $B.Cost$ solution over \emph{CAMA} on instance ``B-n63-k10''. However, as observed in Table \ref{rchr}, on the ``CHRISTOFIDES'' benchmarks, where the size of the problem instances scale up, relative to ``AUGERAT'' and ``CE'' (i.e., the graphs are bigger in size, with larger number of customers or vertices that require servicing), the current baseline state-of-the-art \emph{CAMA} is noted to exhibit superior performances over \emph{CAMA-R} in terms of $Ave.Cost$. In terms of $B.Cost$, \emph{CAMA} is also observed to have attained superior solution quality over \emph{CAMA-R} on instance ``c199''. Since the only difference between \emph{CAMA-R} and \emph{CAMA} lies in the heuristic bias introduced in the population initialization phase of the latter, it is possible to infer from the performances of \emph{CAMA} and \emph{CAMA-R} that the appropriate inductive biases made available in \emph{CAMA} have been effective in narrowing down the search regions of the large scale problems.
\begin{figure*}
\begin{tabular}{ccc}
\includegraphics[width=0.3\textwidth]{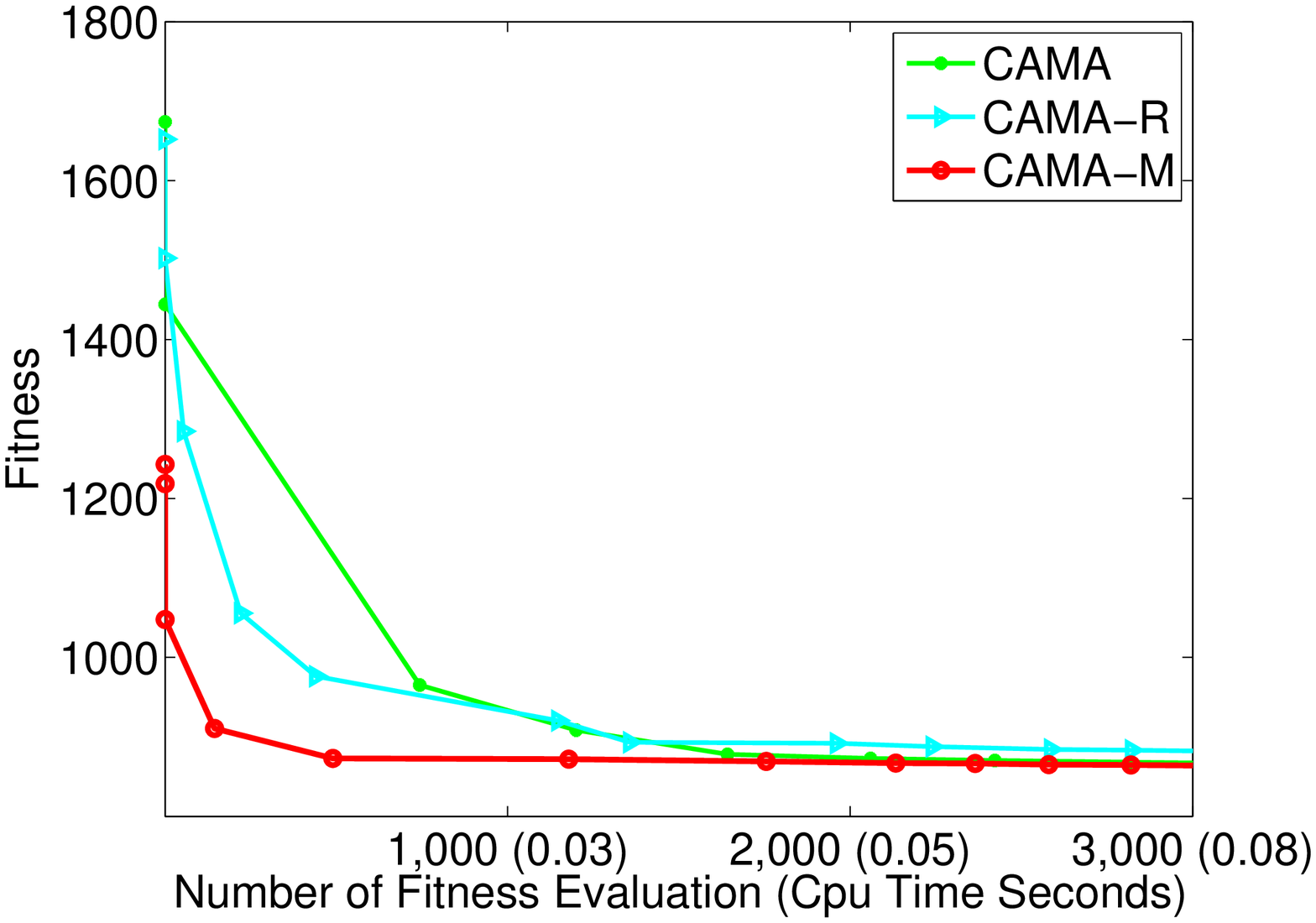} & \includegraphics[width=0.3\textwidth]{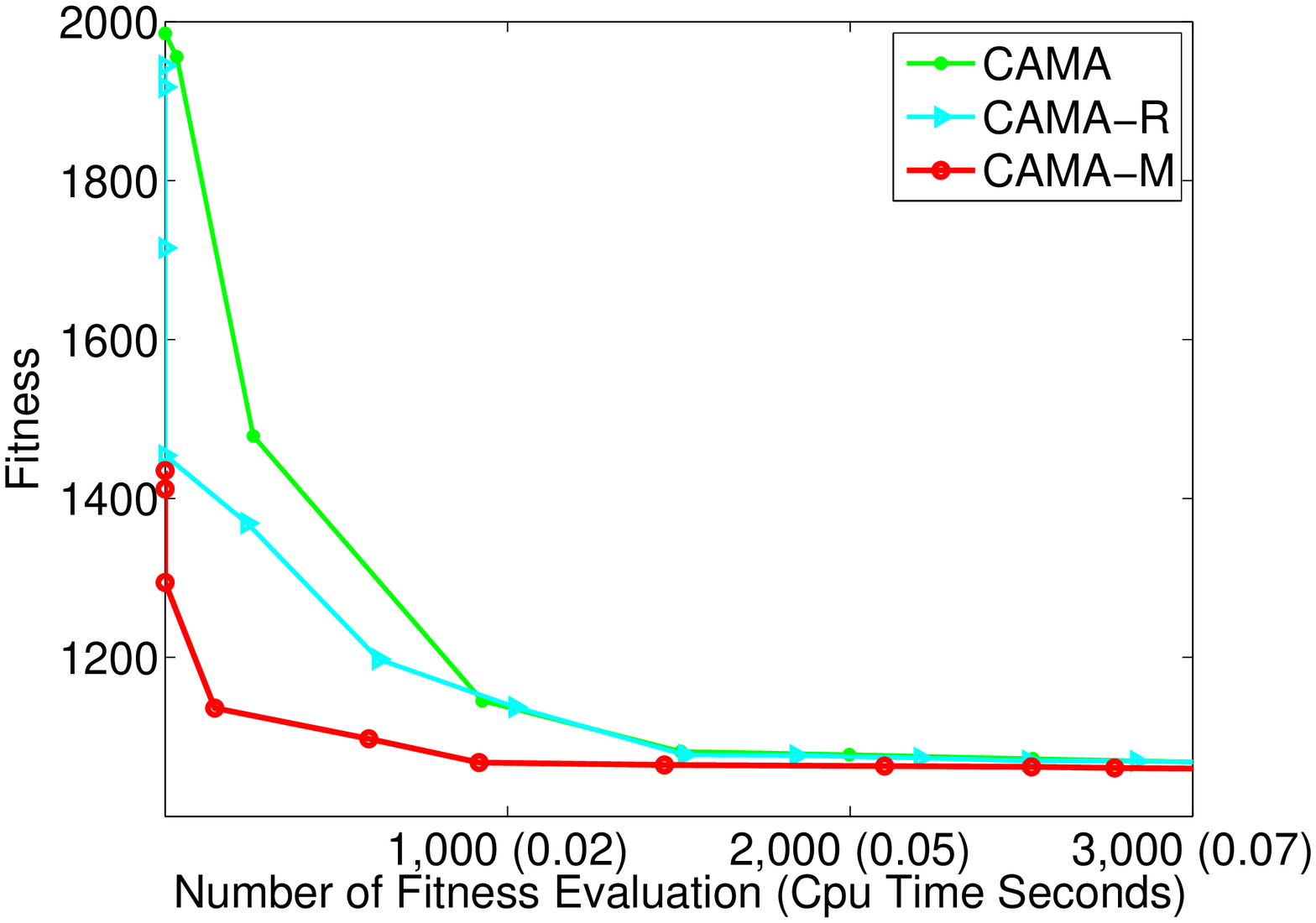} & \includegraphics[width=0.3\textwidth]{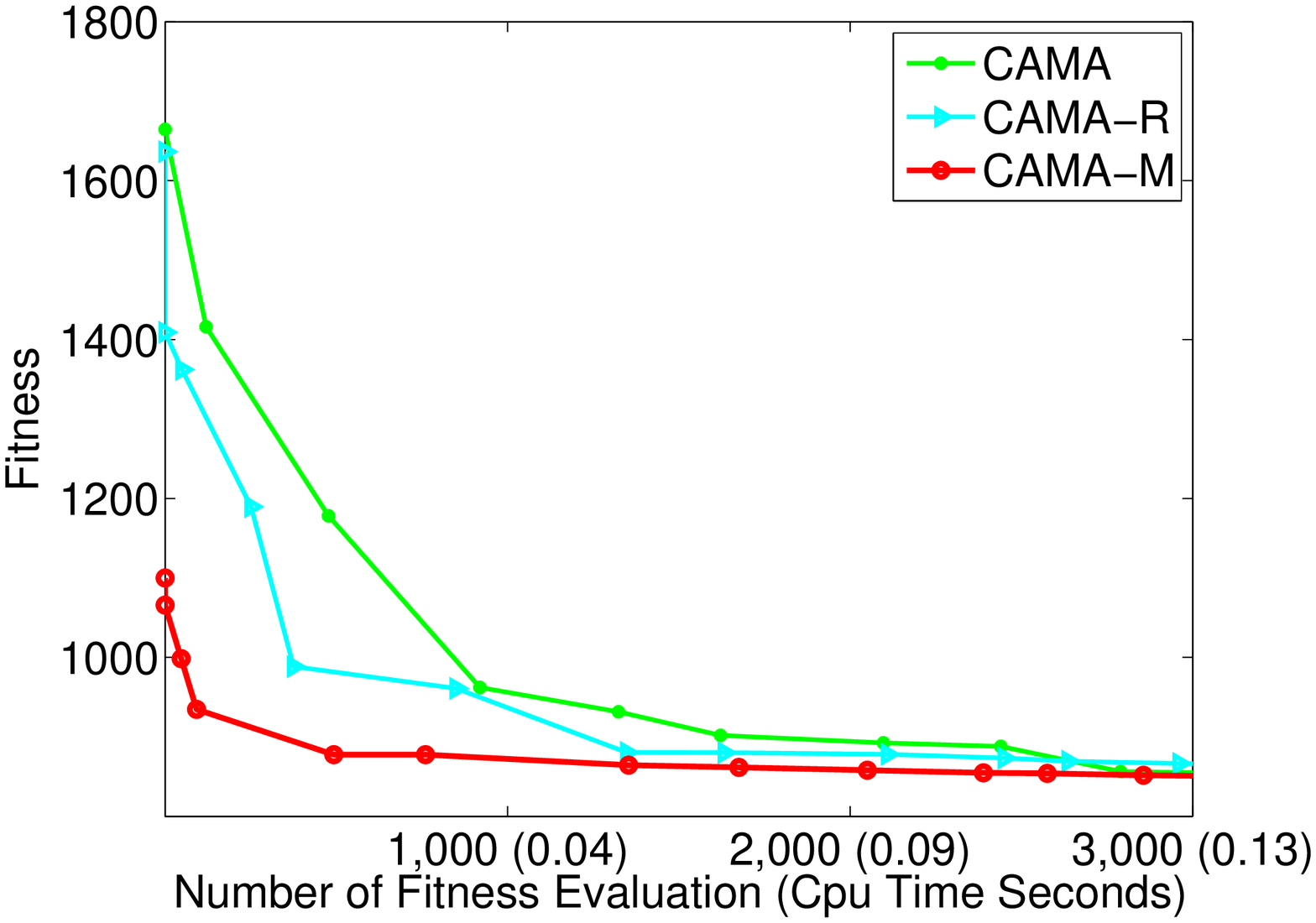} \\
(a) E-n76-k10 & (b) E-n76-k14 &(c) E-n101-k8
\end{tabular}
\caption{Search convergence graphs of $CAMA$, $CAMA-R$, and $CAMA-M$ on representative CVRP ``CE'' benchmarks. $Y$-axis: Fitness value, $X$-axis: Number of Fitness Evaluation or CPU Time in Seconds.}\label{fig:cvrponCE}
\end{figure*}

Moving on next to the proposed memetic computational search paradigm, the generation of the initial population of solutions is now guided by the memes learned from past problem-solving experiences. These memes thus serve as the instructions learned from the experiences along with the CARP instances solved, which are then imitated to enhance the search on new problem instances. As discussed in Section \ref{ME}, the ``meme pool'' of \emph{CAMA-M} is empty when the first CVRP instance is encountered (e.g., ``A-n32-k5'' of ``AUGERAT'' benchmark problem), and thus \emph{CAMA-M} behaves like the baseline \emph{CAMA}. As more CVRP instances are encountered, the \emph{meme learning}, \emph{selection}, \emph{variation} and \emph{imitation} mechanisms shall kick in to learn and generalize memes that would induced positive bias to the evolutionary search of new CVRP problems. It can be observed from Table \ref{raug}, Table \ref{rce} and Table \ref{rchr} that the \emph{CAMA-M} converges to similar solutions attained by both \emph{CAMA-R} and \emph{CAMA} on the first problem instance of each CVRP benchmarks (since no memes are learned yet), but demonstrates superior performances over \emph{CAMA-R} and \emph{CAMA} on subsequent CVRP instances. In particular, on ``AUGERAT'' and ``CE'' benchmarks, \emph{CAMA-M} exhibits superior performances in terms of $Ave.Cost$ and $Success~No.$ on $13$ out of total $19$ CVRP instances. In addition, on ``CHRISTOFIDES'' benchmarks, \emph{CAMA-M} also attained improved solution quality in terms of $Ave.Cost$ and $Success~No.$ on $4$ out of total $7$ CVRP instances. Since \emph{CAMA}, \emph{CAMA-R} and \emph{CAMA-M} shares a common baseline evolutionary solver, i.e., \emph{CAMA}, and differing only in terms of the population initialization phase, the superior performance of \emph{CAMA-M} can clearly be attributed to the effectiveness of the proposed memetic evolutionary search paradigm where meme induced solutions are intelligently generated from previous problem solving experiences for enhancing future evolutionary searches.

\begin{figure*}
\begin{tabular}{ccc}
\includegraphics[width=0.3\textwidth]{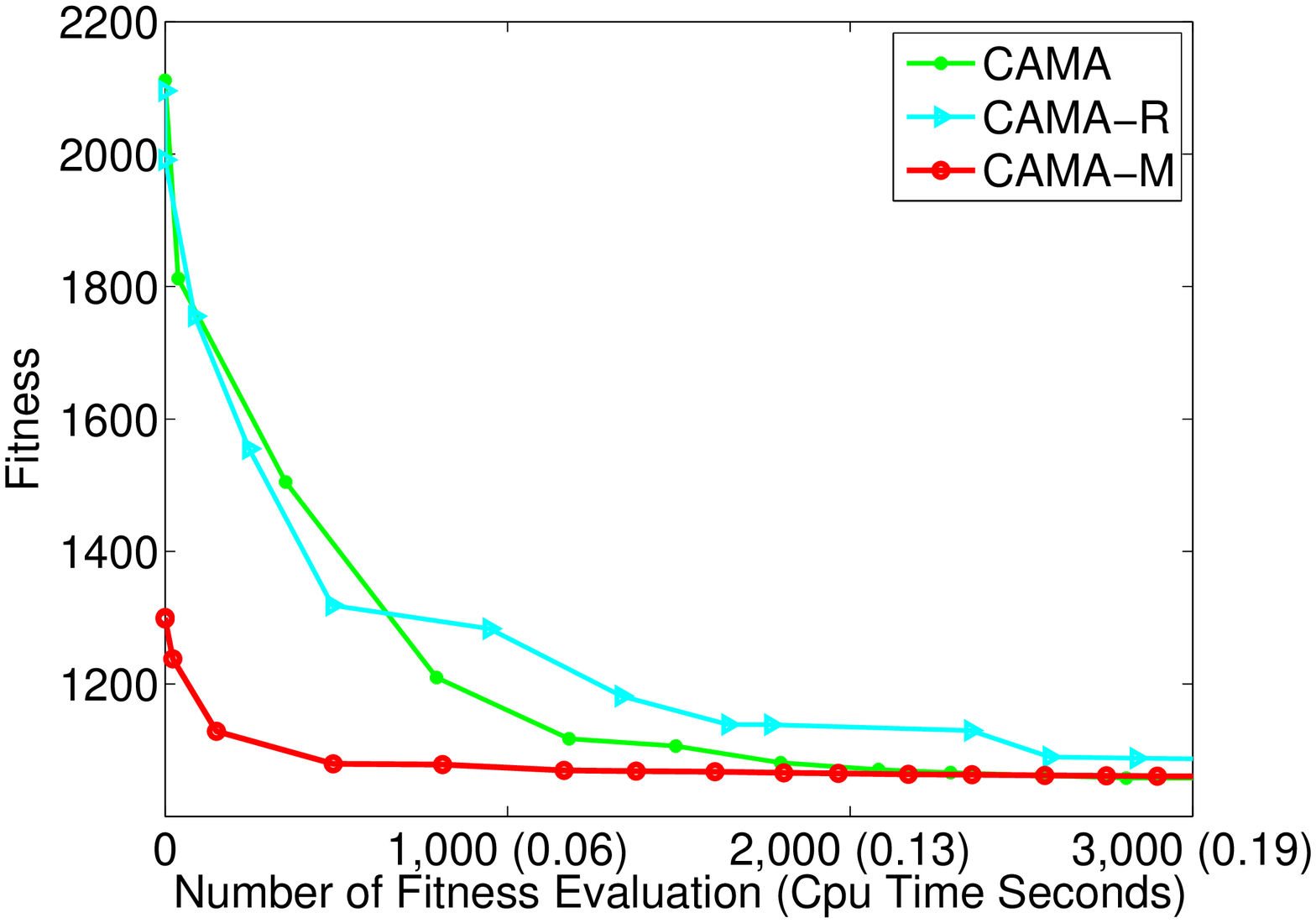} & \includegraphics[width=0.3\textwidth]{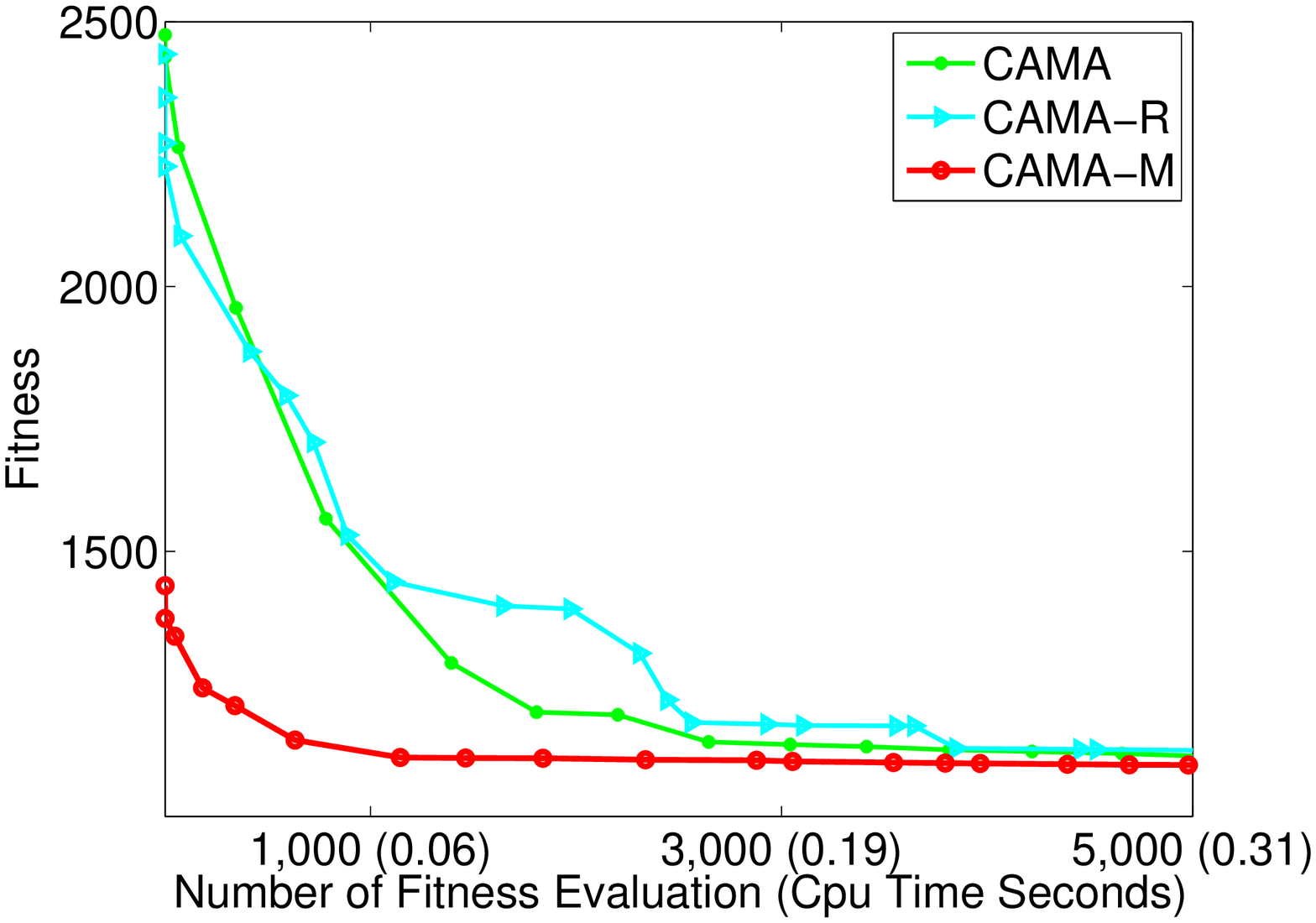} & \includegraphics[width=0.3\textwidth]{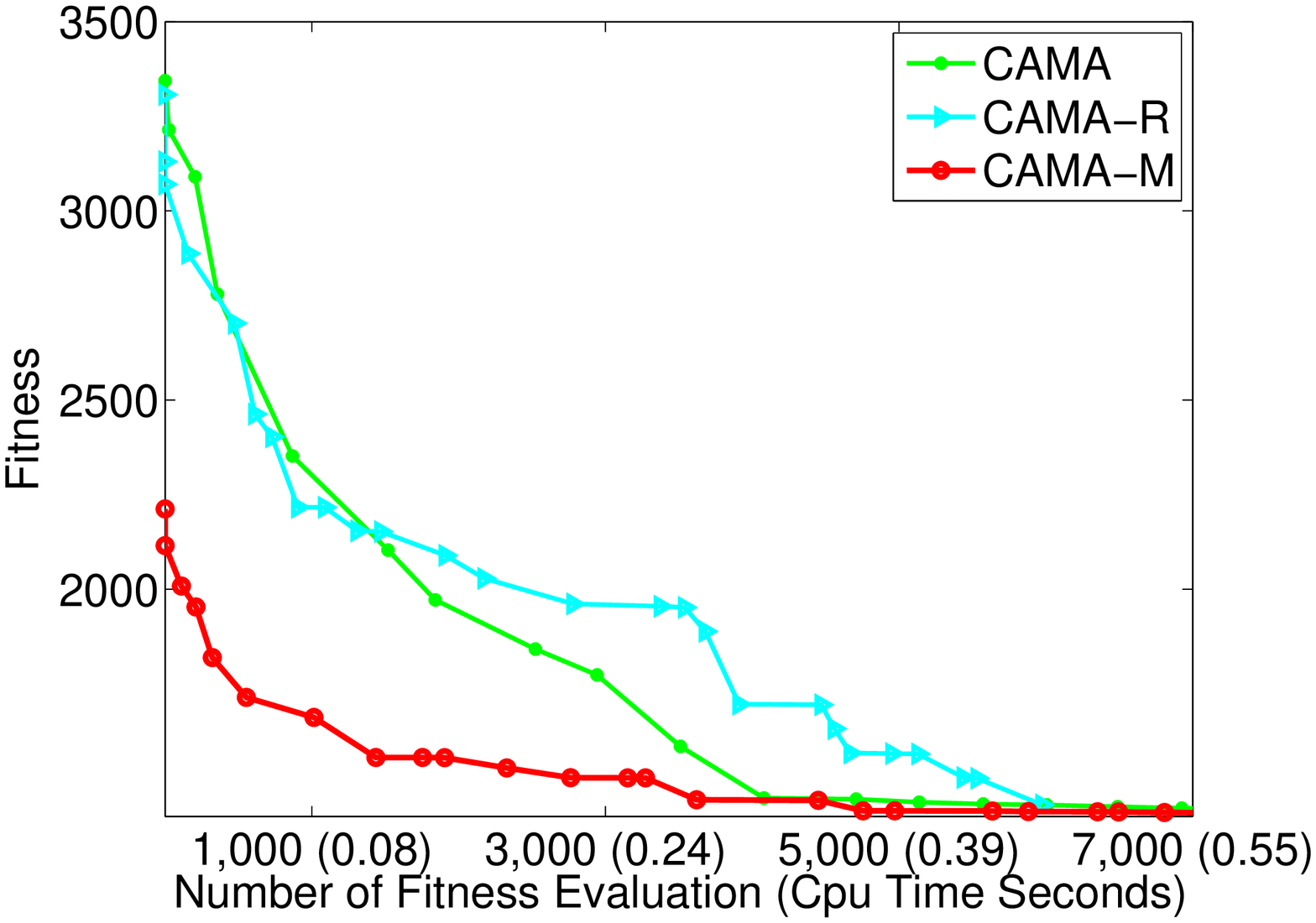} \\
(a) c120 & (b) c150 &(c) c199
\end{tabular}
\caption{Search convergence graphs of $CAMA$, $CAMA-R$, and $CAMA-M$ on representative CVRP ``CHRISTOFIDES'' benchmarks. $Y$-axis: Fitness value, $X$-axis: Number of Fitness Evaluation or CPU Time in Seconds.}\label{fig:cvrponCHR}
\end{figure*}

To assess the efficiency of the proposed memetic search paradigm, representative convergence graphs of \emph{CAMA}, \emph{CAMA-R} and \emph{CAMA-M} on the CVRP benchmarks are also presented in Fig. \ref{fig:cvrponAUG}, Fig. \ref{fig:cvrponCE} and Fig. \ref{fig:cvrponCHR}, respectively. Note that the $Y$-axis of the figures denote the actual fitness value obtained, while the $X$-axis gives the respective computational effort incurred in terms of both the Number of Fitness Evaluation made so far and CPU Time in seconds. As observed, \emph{CAMA-R} is noted to converge faster than \emph{CAMA} on most of the instances in ``AUGERAT'' and ``CE'' benchmarks (e.g., Fig. \ref{fig:cvrponAUG}(a), Fig. \ref{fig:cvrponAUG}(f), Fig. \ref{fig:cvrponCE}(a), Fig. \ref{fig:cvrponCE}(b), etc.). However, on the large scale ``CHRISTOFIDES'' benchmarks, \emph{CAMA} poses to search more efficiently, especially from beyond $1000$ number of fitness evaluations (e.g., Fig. \ref{fig:cvrponCHR}(a), Fig. \ref{fig:cvrponCHR}(b), Fig. \ref{fig:cvrponCHR}(c), etc.). On the other hand, it can be observed that \emph{CAMA-M} converges faster than both \emph{CAMA-R} and \emph{CAMA} on all the CVRP benchmarks. Particularly, on instances ``B-n41-k6'' (Fig. \ref{fig:cvrponAUG}(d)), ``B-n63-k10'' (Fig. \ref{fig:cvrponAUG}(e)) and ``B-n78-k10'' (Fig. \ref{fig:cvrponAUG}(f)), etc., \emph{CAMA-M} takes only approximately $250$ number of fitness evaluations to arrive at the solution quality of \emph{CAMA-R} and \emph{CAMA}, which incurred more than $1500$ number of fitness evaluations. On the large scale instances, such as ``c120'' (Fig. \ref{fig:cvrponCHR}(a)), ``c150'' (Fig. \ref{fig:cvrponCHR}(b)), etc., \emph{CAMA-M} brings about at least $2000$ number of fitness evaluations savings to arrive at the similar solution qualities of \emph{CAMA} and \emph{CAMA-R}.

\begin{figure*}[ht]
    \centering
    \includegraphics[width=0.8\textwidth]{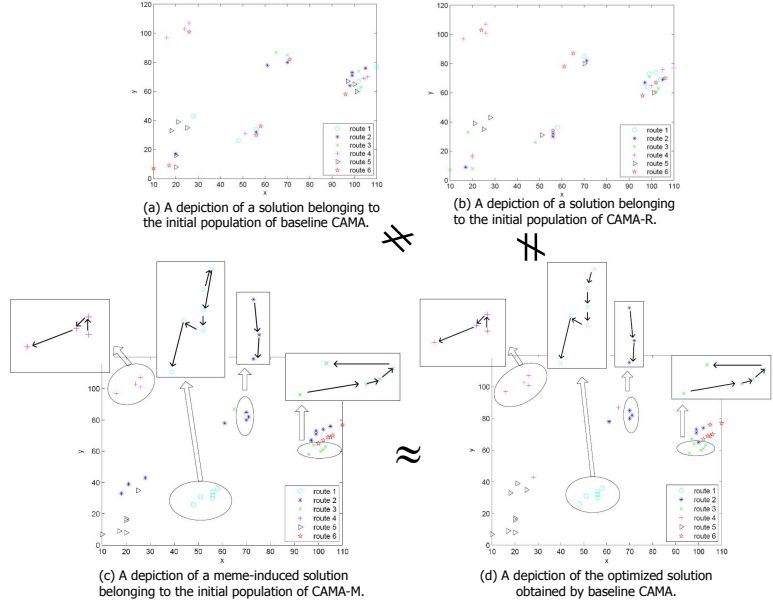}
    \caption{Illustration of the optimized solution and initial solutions of the algorithms considered on ``B-n41-k6'' CVRP benchmark.}
    \label{figTAS}
\end{figure*}

Next, we showcase samples of the initial solutions by \emph{CAMA}, \emph{CAMA-R} and \emph{CAMA-M}, as well as the converged optimized solution of \emph{CAMA} on solving problem instance ``B-n41-k6'' in Fig. \ref{figTAS}. In the present context, an initial solution of \emph{CAMA-M}, which has been intelligently generated using the positive biases introduced by memes learned and generalized from past experiences of problem solving on instances ``A-n32-k5'', ``A-n54-k7'', ``A-n60-k9'' and ``A-n69-k9'', is illustrated. In Fig. \ref{figTAS}, each node denotes a customer that needs to be serviced, and the nodes with the same color and shape shall be serviced by a common route or vehicle. As observed, the task distributions for the initial solution of \emph{CAMA-M} is noted to be most similar to that of the converged optimized solution of \emph{CAMA}, as compared to that of \emph{CAMA} and \emph{CAMA-R}. Besides the task distributions, we also magnify portions of the figures to showcase some service orders of the initial solution in \emph{CAMA-M}, relative to that of the converged optimized solution of baseline \emph{CAMA}, in Fig. \ref{figTAS}(c) and Fig. \ref{figTAS}(d), respectively. The magnified subfigures illustrate high similarities between their respective service orders. This suggests that the service order structures of the converged optimized solution for instances ``A-n32-k5'', ``An54-k7'', ``A-n60-k9'' and ``A-n69-k9'' were successful learned and preserved by the \emph{meme learning operator}, which are subsequently imitated as positively biased solutions (of the initial population in \emph{CAMA-M} search) induced by the generalized meme, through the cultural evolutionary or memetic mechanisms of \emph{meme selection}, \emph{variation} and \emph{imitation}.

\subsection{Capacitated Arc Routing Problem}
Next, we proceed to evaluate the efficiency and effectiveness of the proposed memetic evolutionary search paradigm on the Capacitated Arc Routing Problem domain.

\subsubsection{Empirical Configuration}
The well-established \emph{egl} benchmark is used in the present experimental study on CARP. The data set was generated by Eglese based on data obtained from the winter gritting application in Lancashire \cite{RW94,RL96,LR96}. This data set has been commonly used as benchmark dataset in the literature for CARP solving \cite{LYQA10, YK09}. It consists of two series of datasets (i.e., ``E'' and ``S'' series) with a total of $24$ instances. In particular, CARP instances in ``E'' series have smaller number of vertices, task or edges than that in ``S'' series, thus the problem structures of ``E'' series are deem to be simpler than ``S'' series. The detailed properties of each \emph{egl} instance are presented in Table \ref{egle} and Table \ref{egls}. ``$|V|$'', ``$|E_R|$'', ``$E$'' and ``$LB$'' denote the number of vertices, number of tasks, total number of edges and lower bound, of each problem instance, respectively.
\begin{lrbox}{\tablebox}
\begin{tabular}{c|c c c c c c c c c c c c}
\hline\hline
& & & & & & & ``E'' Series && & & & \\
\hline
Data Set &E1A & E1B& E1C& E2A & E2B & E2C &E3A & E3B & E3C& E4A& E4B & E4C\\
\hline
 $V$ & $77$ & $77$ & $77$ & $77$ & $77$& $77$ & $77$ & $77$ & $77$ & $77$ & $77$ & $77$ \\
 \hline
 $E_r$ & $51$ & $51$ & $51$ & $72$ & $72$ & $72$ & $87$ & $87$ & $87$ & $98$ & $98$ & $98$\\
 \hline
 $E$ &$98$ & $98$ & $98$& $98$& $98$&$98$ & $98$ & $98$ & $98$  & $98$  & $98$  & $98$  \\
 \hline
 $LB$ &$3548$ & $4498$ & $5566$ & $5018$ & $6305$ & $8243$ & $5898$ & $7704$ & $10163$ & $6048$ & $8884$ & $11427$ \\
 \hline\hline
\end{tabular}
\end{lrbox}
\begin{table*}
\centering \small \caption{Properties of the \emph{egl} ``E'' Series CARP benchmarks.}\label{egle} \scalebox{0.7}{\usebox{\tablebox}}
\end{table*}

\begin{lrbox}{\tablebox}
\begin{tabular}{c|c c c c c c c c c c c c}
\hline\hline
& & & & & & &``S'' Series  & && & & \\
\hline
Data Set &S1A & S1B& S1C& S2A & S2B & S2C &S3A & S3B & S3C& S4A& S4B & S4C\\
\hline
 $V$ & $140$ & $140$ & $140$ & $140$ & $140$ & $140$ & $140$ & $140$ & $140$ & $140$ & $140$ & $140$  \\
 \hline
 $E_r$ & $75$& $75$& $75$ & $147$ & $147$ & $147$ & $159$ & $159$ & $159$ & $190$ & $190$ & $190$ \\
 \hline
 $E$ &$190$ &$190$ &$190$ &$190$ &$190$ &$190$ &$190$ &$190$ &$190$ &$190$ &$190$ &$190$ \\
 \hline
 $LB$ & $5018$ & $6384$ & $8493$& $9824$ & $12968$& $16353$ & $10143$ & $13616$& $17100$& $12143$ & $16093$ & $20375$   \\
 \hline\hline
\end{tabular}
\end{lrbox}
\begin{table*}
\centering \small \caption{Properties of the \emph{egl} ``S'' Series CARP benchmarks. $Y$-axis: Fitness value, $X$-axis: Number of Fitness Evaluation or CPU Time in Seconds.}\label{egls} \scalebox{0.7}{\usebox{\tablebox}}
\end{table*}

\begin{lrbox}{\tablebox}
\begin{tabular}{l|c c c c| c c c c|c c c c}
\hline\hline
Data & & $ILMA$& &  & & $ILMA-R$ &  & & & $ILMA-M$ &(Proposed Method)& \\
 Set & $B.Cost$& $Ave.Cost$& $Std.Dev$ &$Success~No.$ & $B.Cost$& $Ave.Cost$& $Std.Dev$ &$Success~No.$& $B.Cost$& $Ave.Cost$& $Std.Dev$ &$Success~No.$ \\
 \hline
 1.E1A &$3548$ &$3548$ & $0$ &$30$& $3548$& $3548$ & $0$  & $30$ & $3548$ & $3548$ & $0$& $30$\\
 2.E1B &$4498$ &$4517.63$ & $12.45$  &$9$& $4498$& $4517.80$ & $13.19$  & $11$ & $4498$ & $\bf{4513.27}$& $14.93$ & $\mathbf{14}$ \\
 3.E1C &$5595$ &$5599.33$ & $7.56$  &$22$& $5595$& $5601.73$ & $8.84$  & $18$ & $5595$ & $\bf{5598.07}$ & $8.58$ & $\mathbf{26}$ \\
 4.E2A &$5018$ &$5018$ & $0$  &$30$&$5018$ &$5018$ & $0$  &$30$&$5018$ &$5018$ & $0$  &$30$ \\
 5.E2B &$6317$ &$6341.53$ & $20.15$  &$14$& $6317$ & $6344.03$ & $22.38$  & $10$ & $6317$ & $\bf{6337.90}$ &$11.90$ & $\mathbf{16}$ \\
 6.E2C & $8335$&$8359.87$ & $36.61$ & $21$  &$8335$&  $8355.07$ & $39.26$  & $24$ & $8335$ & $\bf{8349.97}$& $26.16$ & $\mathbf{25}$ \\
 7.E3A &$5898$ &$5921.23$ & $30.07$  &$20$& $5898$& $5916.93$ & $30.50$& $21$  & $5898$ & $\bf{5910.97}$ & $30.57$ & $\mathbf{25}$\\
 8.E3B &$7777$ &$7794.77$ & $23.08$  &$22$& $7777$&  $7792.17$ & $29.95$  & $23$ & $\bf{7775}$ & $\bf{7788.70}$ & $15.74$ & $\mathbf{25}$\\
 9.E3C &$10292$ &$\bf{10318.73}$ & $40.89$ &$\mathbf{20}$& $10292$& $10327.07$ & $33.46$  & $15$ & $10292$ & $10319.16$ & $36.15$ &$\mathbf{20}$ \\
 10.E4A &$6461$&$6471.37$ & $15.16$ & $\mathbf{21}$& $\bf{6458}$& $6481.77$  & $22.77$ & $15$ & $6461$ & $\bf{6469.80}$ & $10.27$ & $\mathbf{21}$\\
 11.E4B &$8995$&$9060.67$ & $45.29$ & $10$& $8993$& $9067.93$  & $50.54$ & $12$ & $\bf{8988}$ & $\bf{9053.97}$ & $41.49$ & $\mathbf{15}$\\
 12.E4C &$\bf{11555}$&$\bf{11678.47}$ & $73.57$ & $\mathbf{16}$& $11594$& $11728.30$  & $82.39$ & $10$ & $11576$ & $11697.27$ & $76.98$ & $14$\\
 \hline\hline
\end{tabular}
\end{lrbox}
\begin{table*}
\centering \small \caption{Statistic results of \emph{ILMA}, \emph{ILMA-R}, and \emph{ILMA-M} on $egl$ ``E''-Series CARP benchmarks.}\label{regle} \scalebox{0.45}{\usebox{\tablebox}}
\end{table*}

\begin{lrbox}{\tablebox}
\begin{tabular}{l|c c c c| c c c c|c c c c}
\hline\hline
Data & & $ILMA$& &  & & $ILMA-R$ &  & & & $ILMA-M$ &(Proposed Method)& \\
 Set & $B.Cost$& $Ave.Cost$& $Std.Dev$ &$Success~No.$ & $B.Cost$& $Ave.Cost$& $Std.Dev$ &$Success~No.$& $B.Cost$& $Ave.Cost$& $Std.Dev$ &$Success~No.$ \\
 \hline
 1.S1A &$5018$ &$5023.93$ & $18.14$ &$27$& $5018$& $5025.97$ & $26.97$  & $27$ & $5018$ & $5023.67$ & $25.39$& $27$\\
 2.S1B &$6388$ &$6404.07$ & $22.96$  &$20$& $6388$& $6403.30$ & $20.89$  & $20$ & $6388$ & $\bf{6392.80}$& $14.65$ & $\mathbf{27}$ \\
 3.S1C &$8518$ &$8577.63$ & $44.18$  &$\mathbf{11}$& $8518$& $8581.67$ & $33.98$  & $8$ & $8518$ & $\bf{8576.53}$ & $33.12$ & $\mathbf{11}$ \\
 4.S2A &$9920$ &$10037.43$ & $61.51$  &$14$&$9925$ &$10050.30$ & $54.24$  &$11$&$\bf{9896}$ &$\bf{10010.20}$ & $67.13$  &$\mathbf{20}$ \\
 5.S2B &$13191$ &$13260.03$ & $45.37$  &$16$& $13173$ & $13257.90$ & $48.94$  & $15$ & $\bf{13147}$ & $\bf{13245.56}$ &$53.02$ & $\mathbf{22}$ \\
 6.S2C & $16507$&$\bf{16605.10}$ & $65.26$ & $\mathbf{16}$  &$16480$&  $16626.43$ & $62.90$  & $10$ & $\bf{16468}$ & $16615.40$& $76.79$ & $10$ \\
 7.S3A &$10248$ &$10342.77$ & $47.56$  &$11$& $10278$& $10369.40$ & $52.42$& $13$  & $\bf{10239}$ & $\bf{10339.40}$ & $53.29$ & $\mathbf{18}$\\
 8.S3B &$13764$ &$13912.97$ & $79.85$ &$12$& $13779$& $13899.70$ & $76.96$  & $17$ & $\bf{13749}$ & $\bf{13881.33}$ & $85.78$ &$\mathbf{21}$ \\
 9.S3C &$17274$ &$17371.10$ & $79.12$  &$20$& $17277$&  $17402.43$ & $74.37$  & $12$ & $\bf{17261}$ & $\bf{17355.03}$ & $48.23$ & $\mathbf{21}$\\
 10.S4A &$12335$&$12498.47$ & $67.72$ & $15$& $12407$& $12534.47$  & $63.23$ & $11$ & $\bf{12320}$ & $\bf{12489.43}$ & $83.91$ & $\mathbf{19}$\\
 11.S4B &$\bf{16378}$&$16542.93$ & $89.65$ & $17$& $16443$& $16540.43$  & $87.52$ & $19$ & $16415$ & $\bf{16512.43}$ & $57.54$ & $\mathbf{20}$\\
 12.S4C &$20613$&$20794.80$ & $77.51$ & $14$& $20589$& $20841.13$  & $85.53$ & $5$ & $\bf{20564}$ & $\bf{20774.20}$ & $86.78$ & $\mathbf{17}$\\
 \hline\hline
\end{tabular}
\end{lrbox}
\begin{table*}
\centering \small \caption{Statistic results of \emph{ILMA}, \emph{ILMA-R}, and \emph{ILMA-M} on $egl$ ``S''-Series CARP benchmarks.}\label{regls} \scalebox{0.45}{\usebox{\tablebox}}
\end{table*}
In traditional CARP, each task is represented by a corresponding \emph{head vertex}, \emph{tail vertex}, \emph{travel cost} and \emph{demand (service cost)}. The shortest distance matrix of the vertices is first derived by means of the Dijkstra's algorithm \cite{EW59}, i.e., using the distances available between the vertices of a CARP. The coordinate features (i.e., locations) of each task are then approximated by means of multidimensional scaling \cite{BorgGroenen2005}. In this manner, each task is represented as a node in the form of coordinates. A CARP instance in the current setting is thus represented by input vector $\mathbf{X}$ composing of the coordinate features of all tasks in the problem. Such a representation would allow standard clustering approaches, such as the K-Means algorithm to be conducted on the CARP in task assignments, and allow the pairwise distances sorting among tasks available for preserving service orders. The label information of each task in $\mathbf{Y}$ belonging to the CARP instance, is as defined by the optimized solution of CARP.

For empirical comparisons, three solution population initialization procedures based on \emph{ILMA} are also considered here. The first is a simple random approach, which is labeled here as \emph{ILMA-R}. The second is the informed heuristic based population generation procedure used in the baseline state-of-the-art \emph{ILMA} \cite{YK09} for CARP. There, the initial population is formed by a fusion of chromosomes generated from Augment\_Merge \cite{BR81}, Path\_Scanning \cite{BJE83}, Ulusoy's Heuristic \cite{GU85} and the simple random initialization procedures. The last is our proposed solution generation using past problem solving experiences with memetic evolution, which is labeled here as \emph{ILMA-M}. In \emph{ILMA-M}, the meme pool is empty at start, while new memes shall be accumulated with increasing CARP instances optimized. Thus, once again, the \emph{ILMA-M} performs exactly like \emph{ILMA} on the first problem instance encountered, since it is meant to serve as an intelligent \emph{ILMA} whose intellectual shall increase with increasing problems solved.
\begin{figure*}
\begin{tabular}{ccc}
\includegraphics[width=0.3\textwidth]{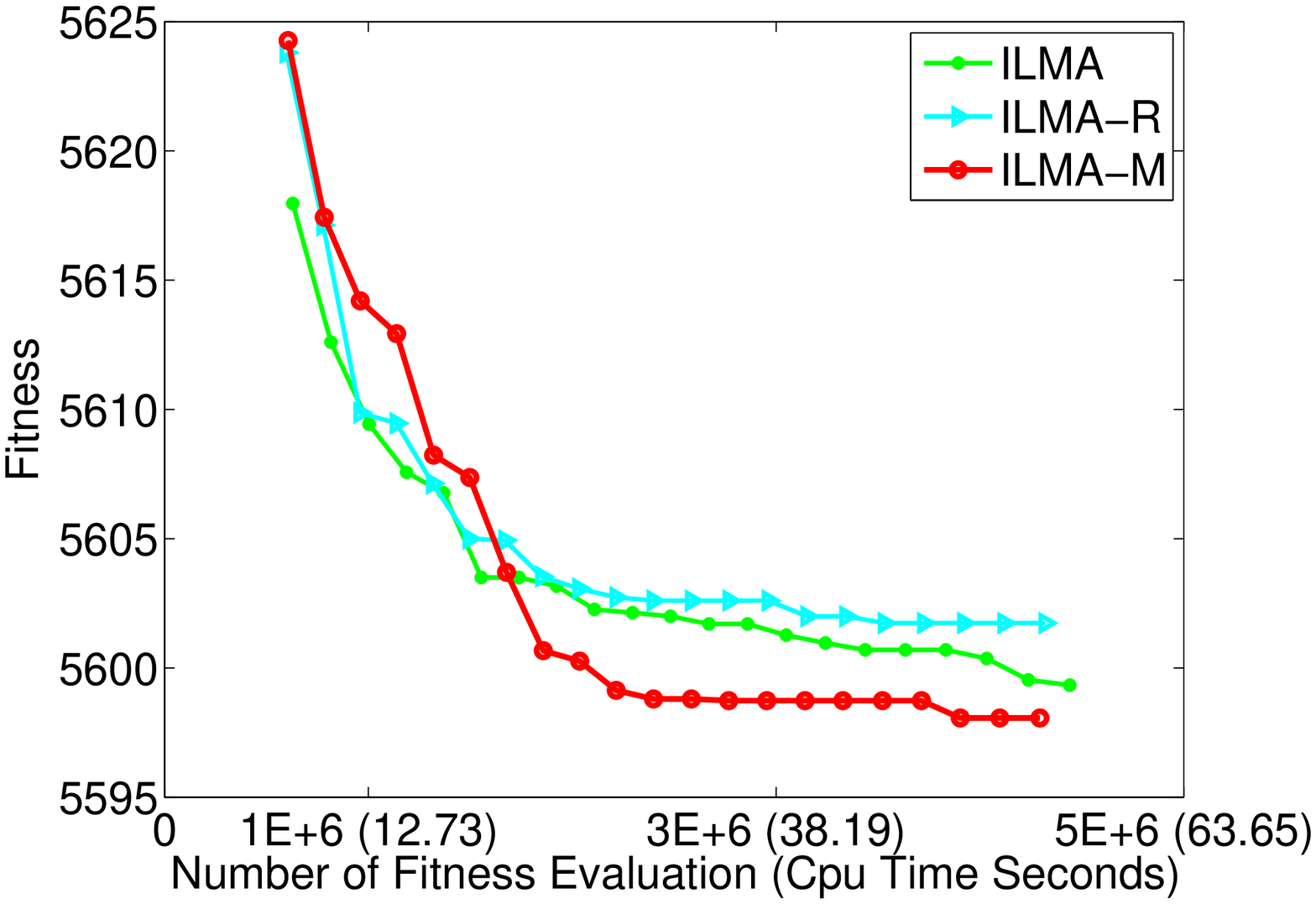} & \includegraphics[width=0.3\textwidth]{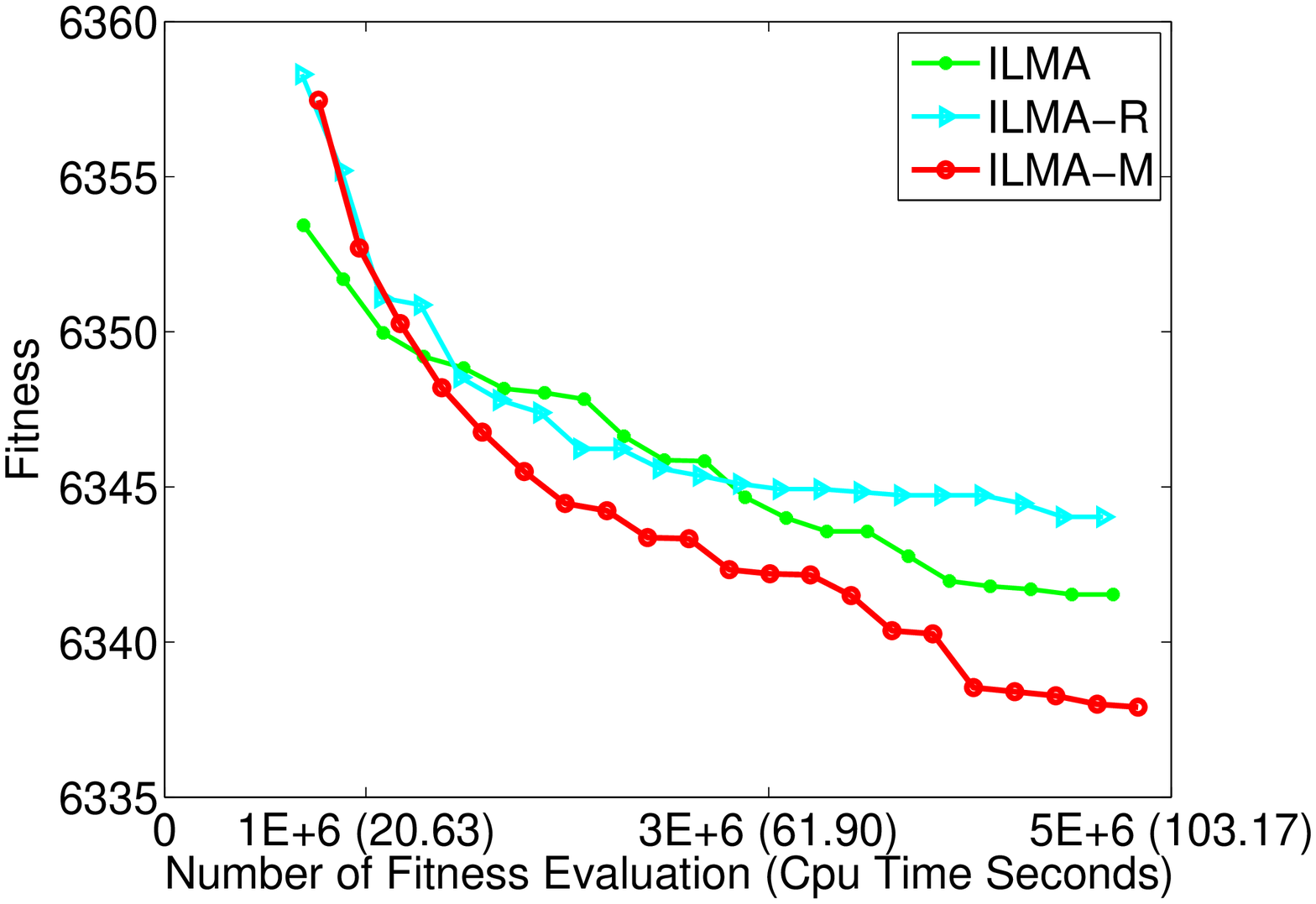} & \includegraphics[width=0.3\textwidth]{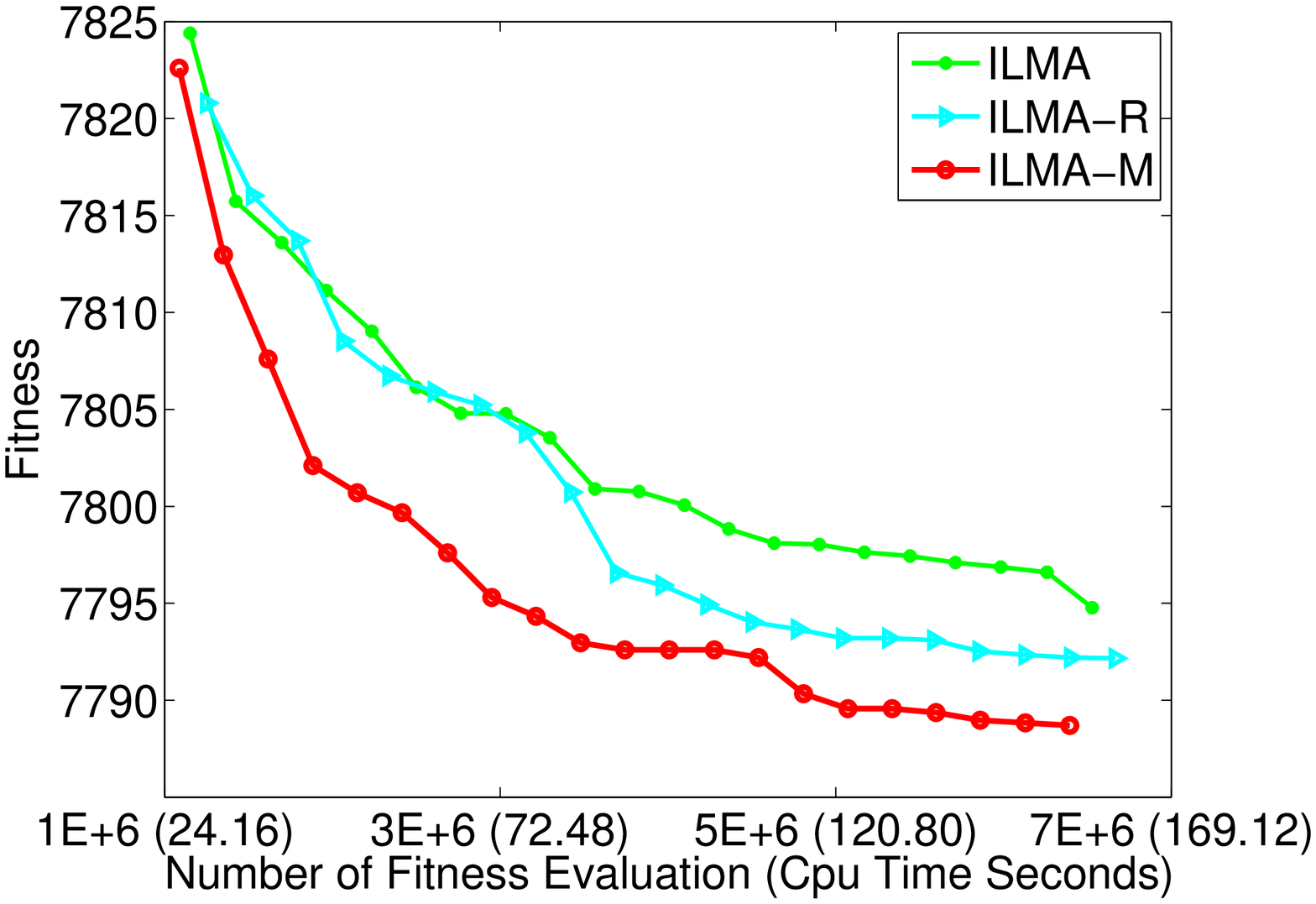} \\
(a) E1C & (b) E2B &(c) E3B \\
\includegraphics[width=0.3\textwidth]{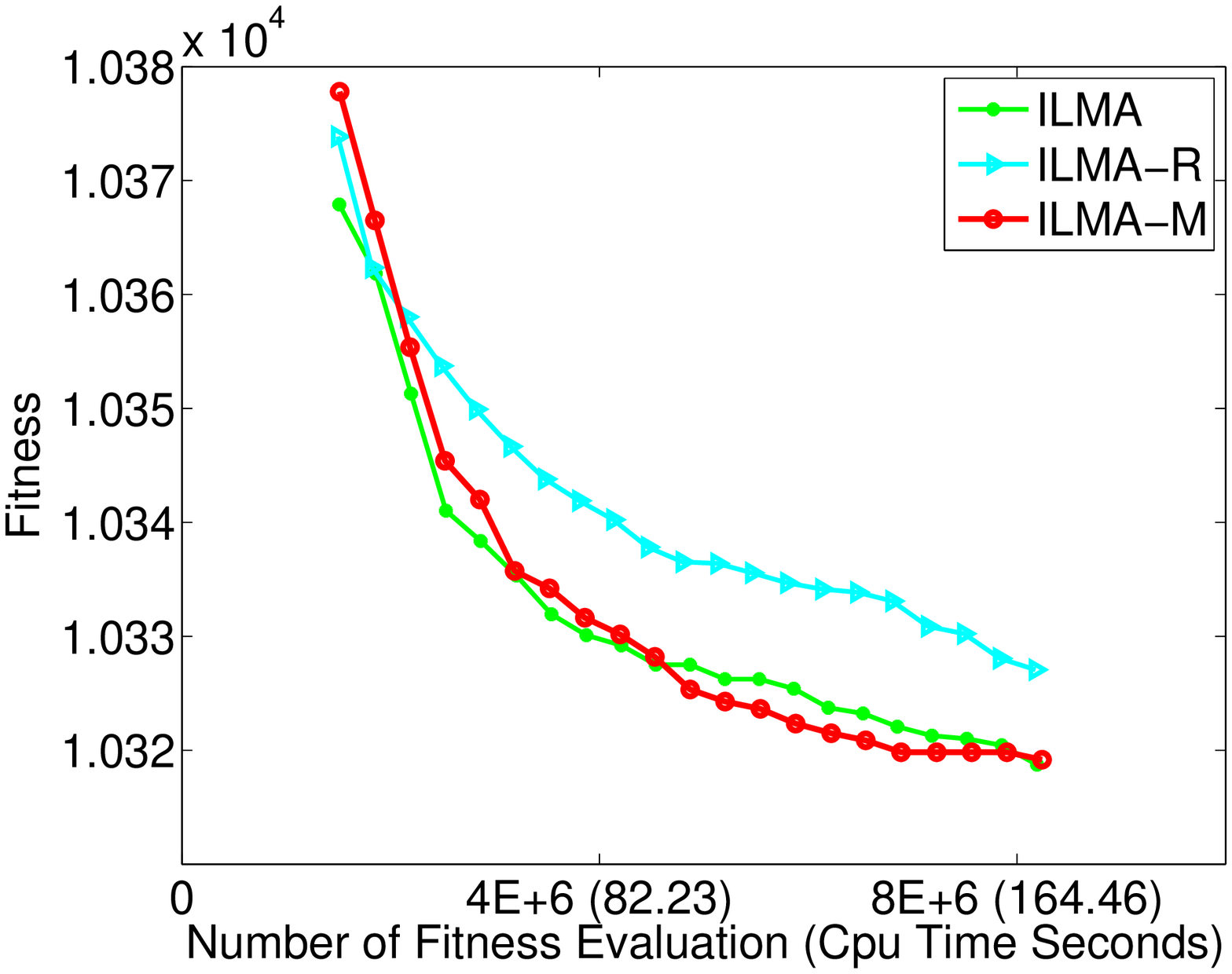} & \includegraphics[width=0.3\textwidth]{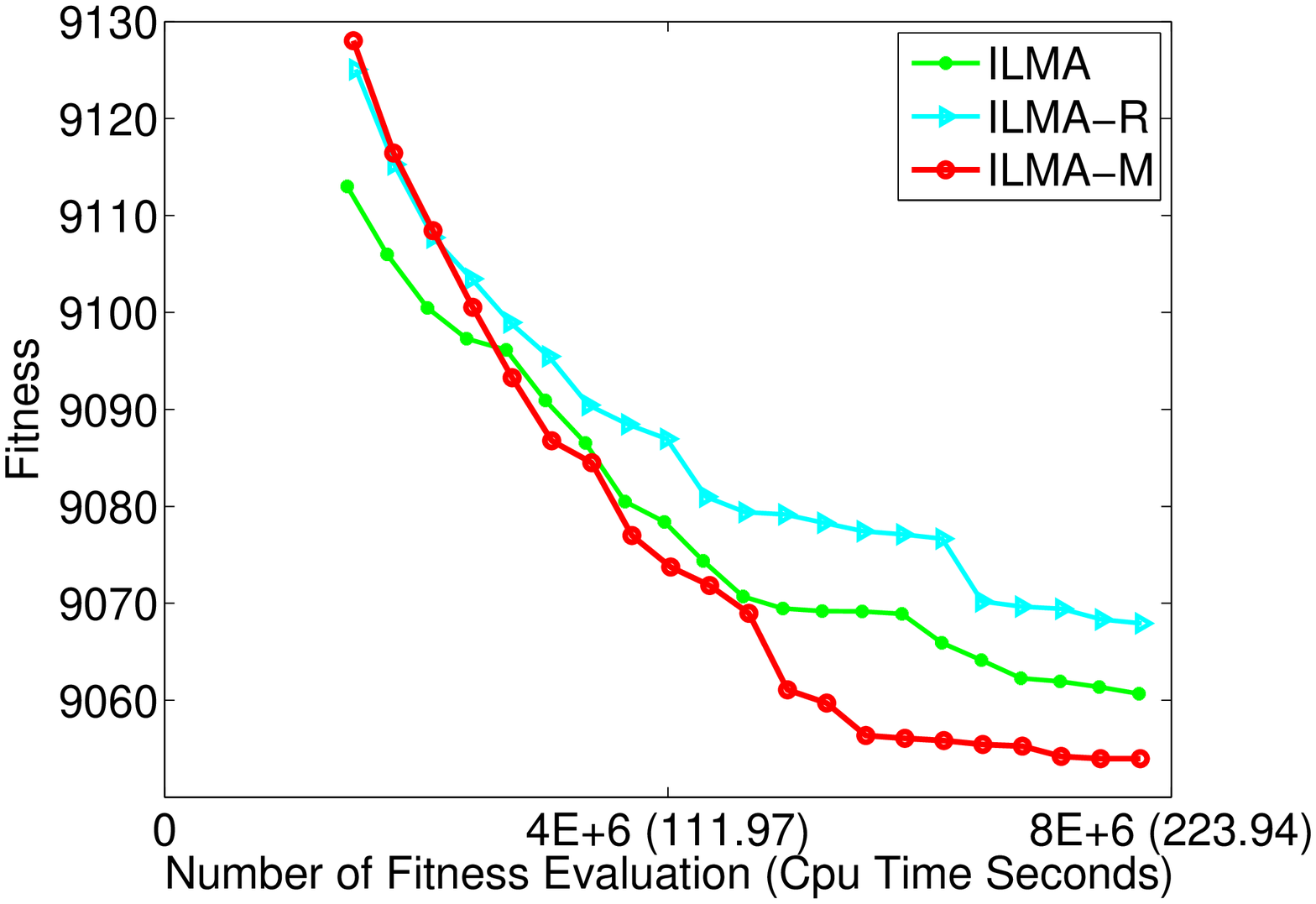} & \includegraphics[width=0.3\textwidth]{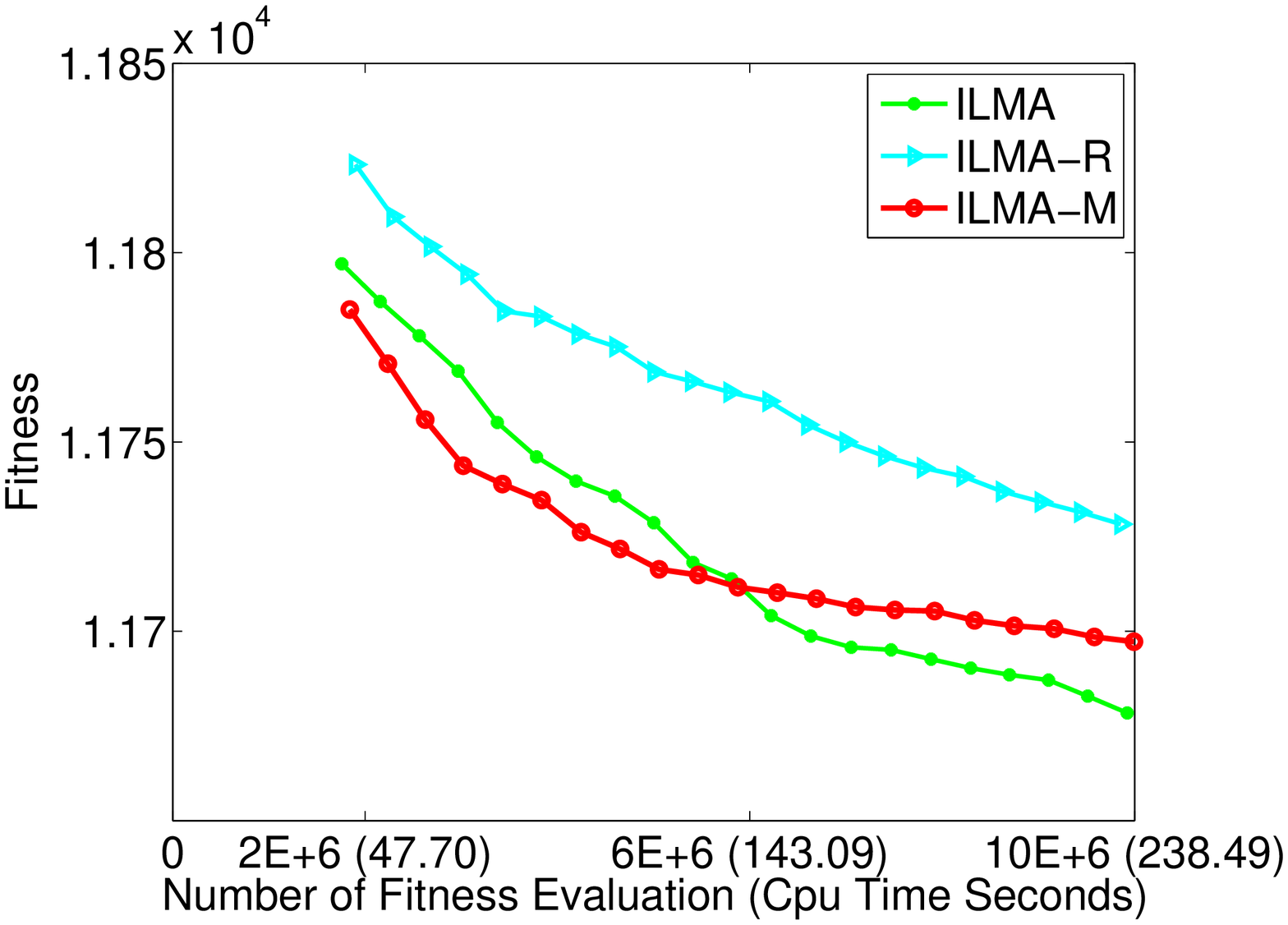} \\
(d) E3C & (e) E4B &(f) E4C
\end{tabular}
\caption{Search convergence graphs of $ILMA$, $ILMA-R$, and $ILMA-M$ on representative CARP ``E''-Series benchmarks. $Y$-axis: Fitness value, $X$-axis: Number of Fitness Evaluation or CPU Time in Seconds.}\label{fig:carponverge}
\end{figure*}

To facilitate a fair comparison and verify the benefits of the learning from past experiences, the evolutionary operators of \emph{ILMA} and its variants are configured as consistent to that reported in \cite{YK09}. Further, in \emph{ILMA-M}, the MMD of Equation \ref{sel1} is augmented with the demand of each task.
\subsubsection{Result and Discussion}
Table \ref{regle} and Table \ref{regls} tabulate the results that measure the solution quality on the ``E''-Series and ``S''-Series \emph{egl} CARP datasets as obtained by the respective algorithms, across $30$ independent runs. The $Ave.Cost$ value of \emph{ILMA} on each problem instance is used as threshold level in the $Success~No.$ criterion. In the tables, the method with superior performance with respect to ``$B.Cost$'', ``$Ave.Cost$'' and ``$Success~No.$'' are highlighted in bold font.

As can be observed from Table \ref{regle} and Table \ref{regls}, the \emph{ILMA-R} has attained competitive performance on solution quality over \emph{ILMA-R}, on instances such as ``E1-A'', ``E1-B'', ``E2-A'', ``S1-A'', ``S1-B'' and ``S2-A'', etc. However, with the incorporation of heuristic information as inductive search bias in the baseline state-of-the-art \emph{ILMA}, it is noted to attain improved performance in terms of solution quality over \emph{ILMA-R} on the large scale instances (i.e., instances with greater service arcs, travel edges, or number of vertices, etc.), which include ``E3-C'', ``E4-A'', ``E4-B'', ``E4-C'', ``S3-C'', ``S4-A'', and ``S4-C''.

Turning then to the proposed memetic evolutionary search paradigm, it can be seen from Table \ref{regle} and Table \ref{regls} that, \emph{ILMA-M} performed competitively to \emph{ILMA} on instances ``E1-A'' and ``S1-A'' as expected since these are the first encountered problem instances of \emph{ILMA-M} on the ``E'' and ``S'' \emph{egl} CARP benchmarks, respectively, where no memes are available in the meme pool. As more problem instances are optimized, the \emph{ILMA-M} is observed to demonstrate superior performance over \emph{ILMA} in terms of $Ave.Cost$ on $18$ out of total $24$ \emph{egl} benchmarks. In terms of $B.Cost$, \emph{ILMA-M} achieved $2$ and $8$ improved solution qualities over \emph{ILMA} and \emph{ILMA-R} on the ``E'' and ``S'' series \emph{egl} datasets, respectively.

\begin{figure*}
\begin{tabular}{ccc}
\includegraphics[width=0.3\textwidth]{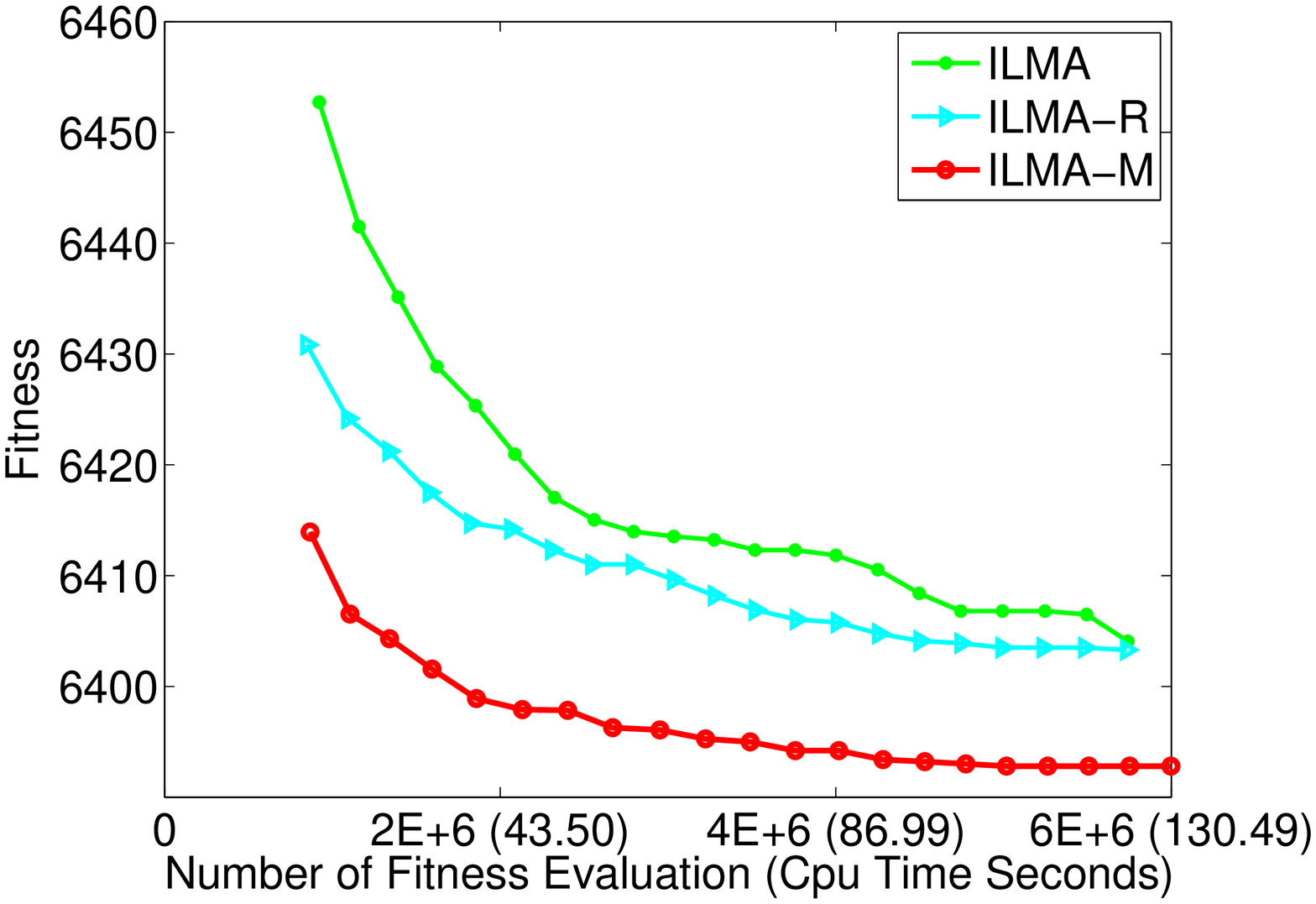} & \includegraphics[width=0.3\textwidth]{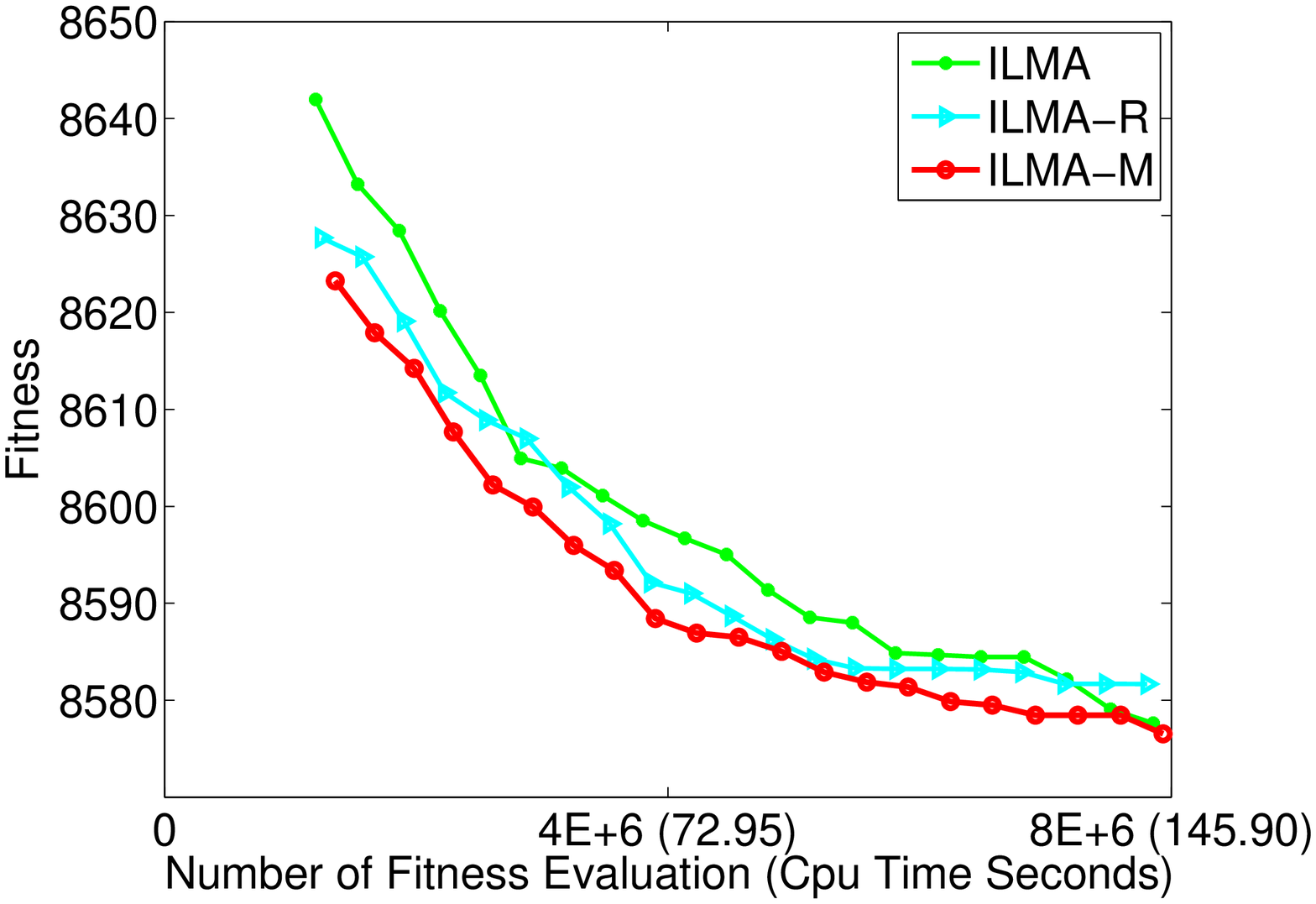} & \includegraphics[width=0.3\textwidth]{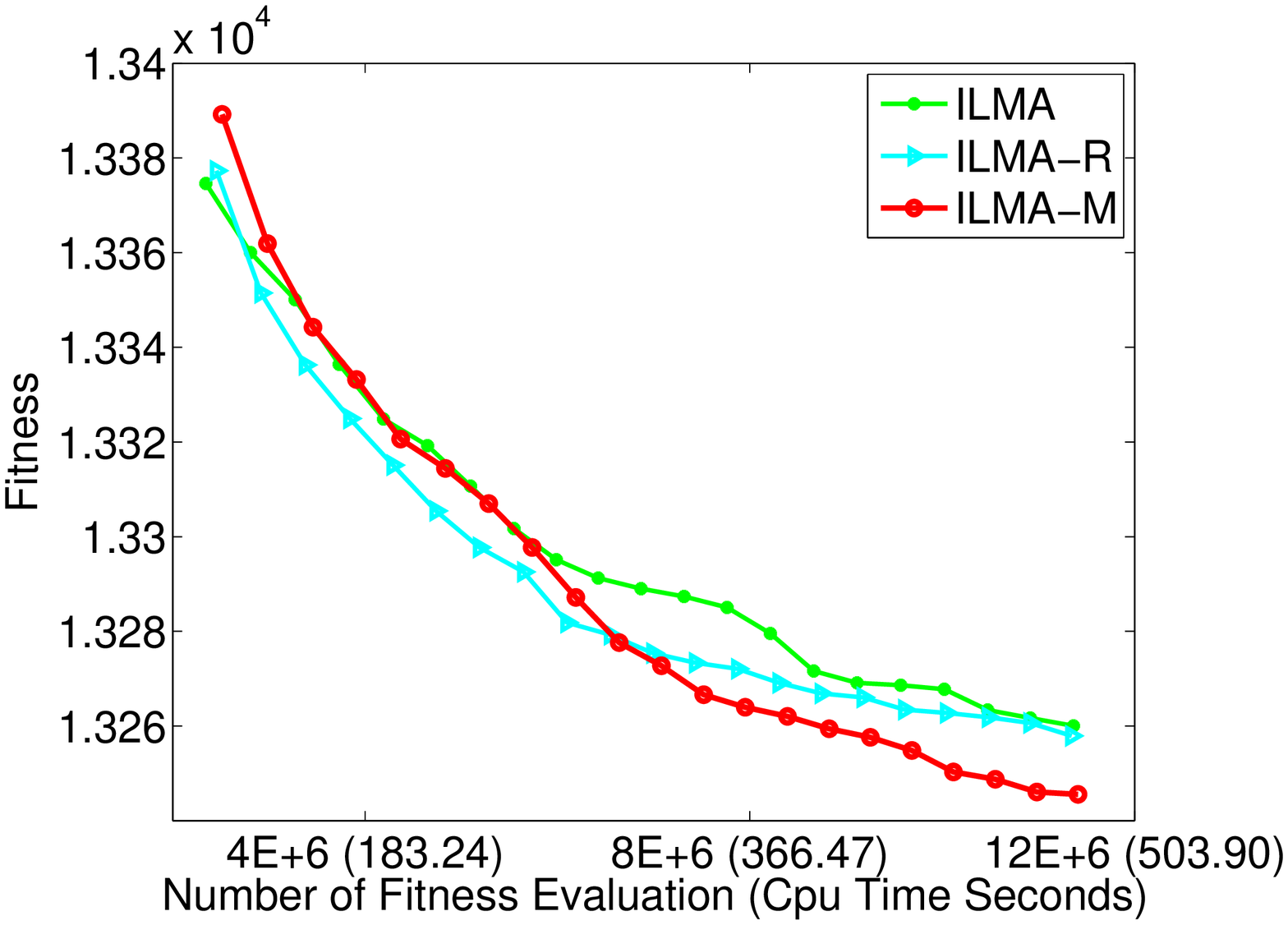} \\
(a) S1B & (b) S1C &(c) S2B\\
\includegraphics[width=0.3\textwidth]{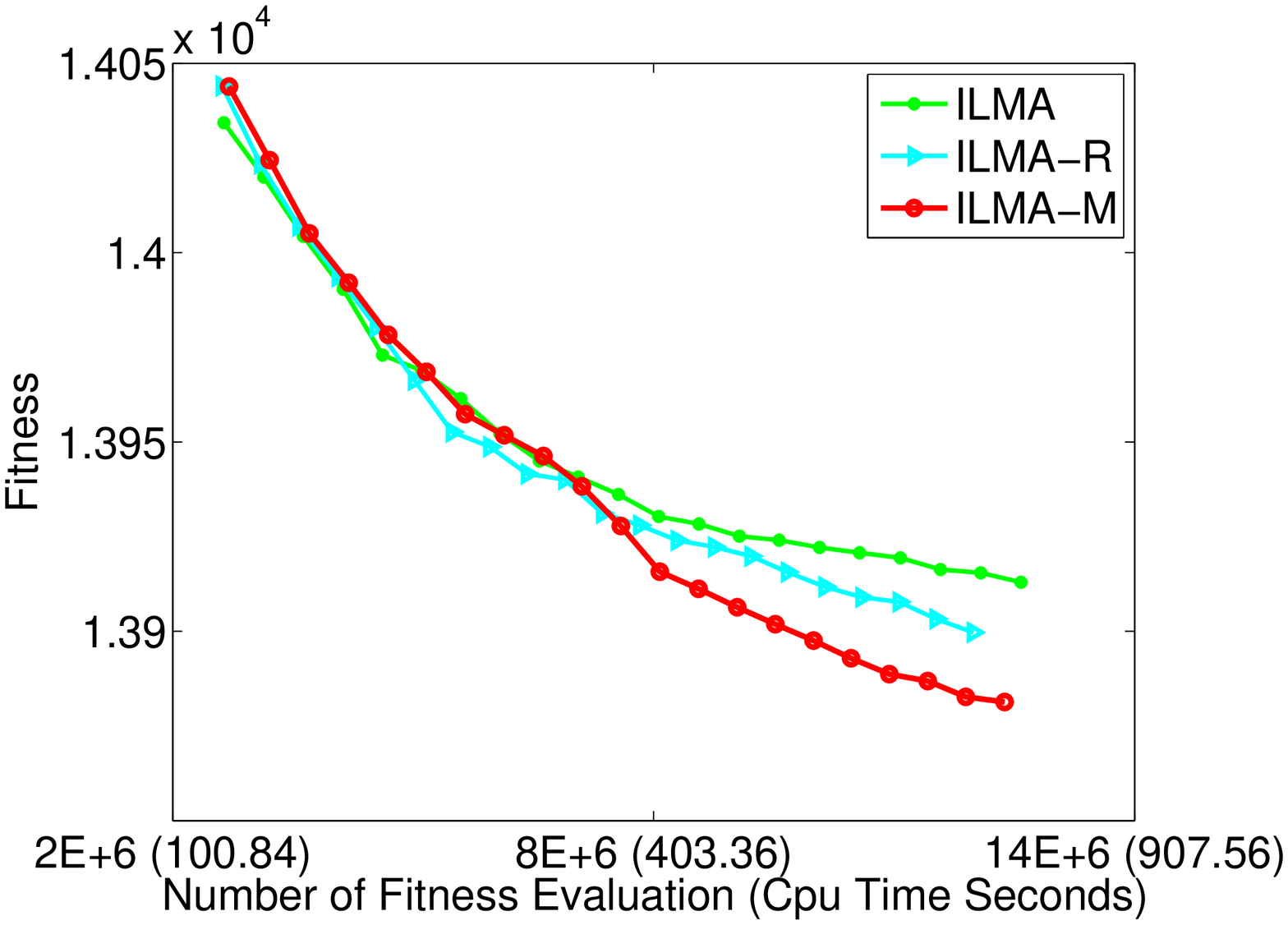} & \includegraphics[width=0.3\textwidth]{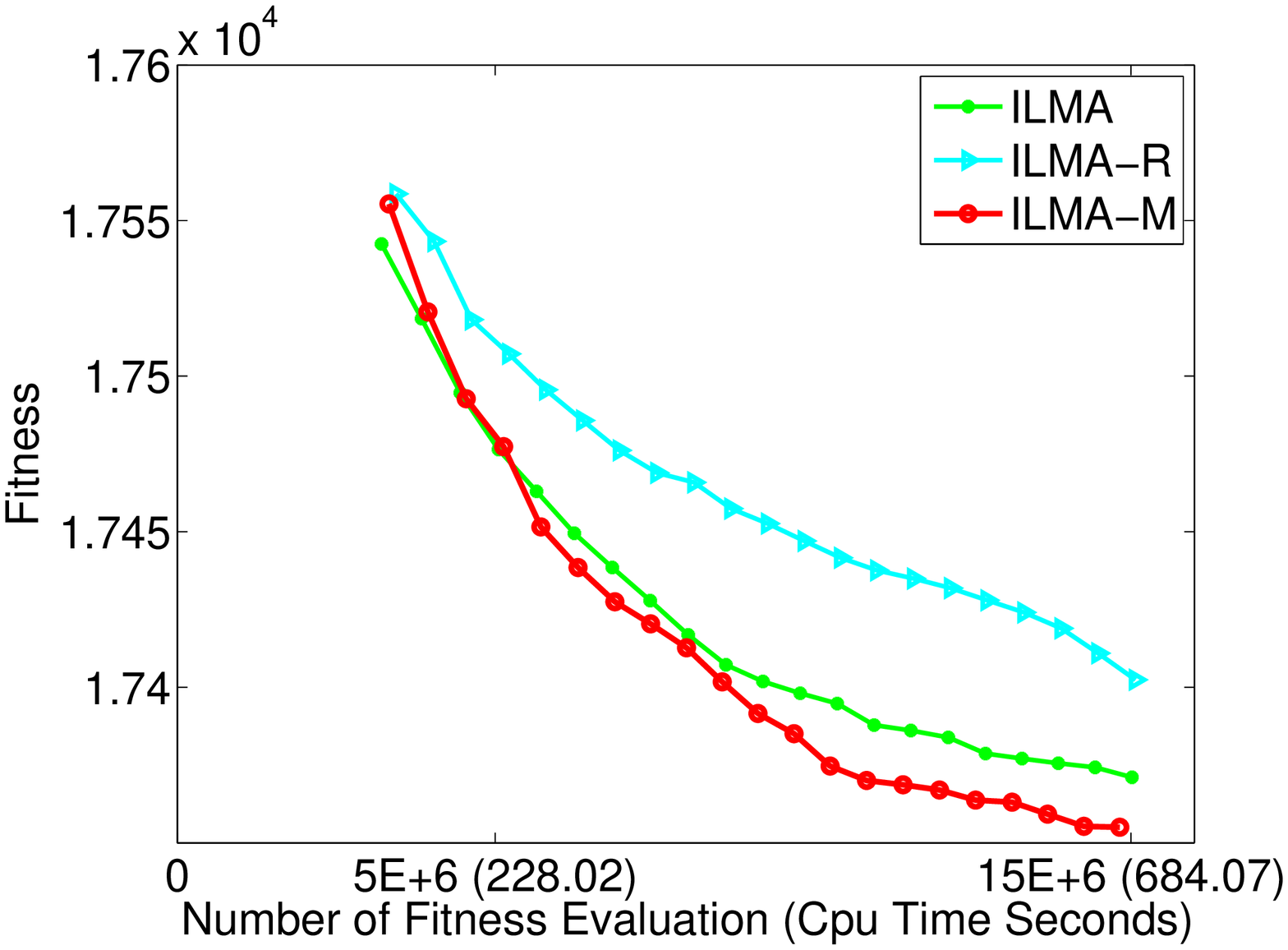} & \includegraphics[width=0.3\textwidth]{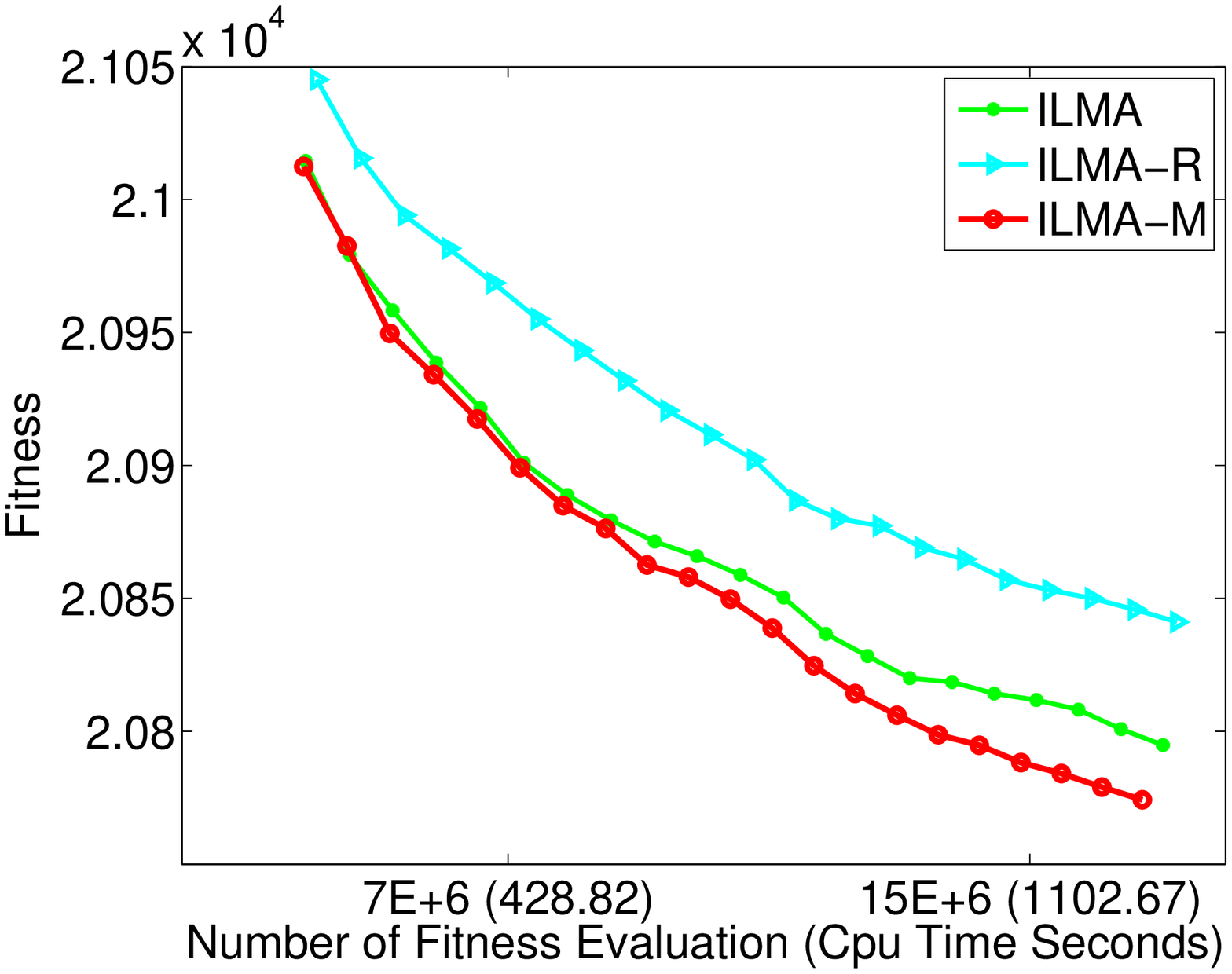} \\
(d) S3B & (e) S3C &(f) S4C
\end{tabular}
\caption{Search convergence graphs of $ILMA$, $ILMA-R$, and $ILMA-M$ on representative CARP ``S''-Series benchmarks. $Y$-axis: Fitness value, $X$-axis: Number of Fitness Evaluation or CPU Time in Seconds.}\label{fig:carponvergs}
\end{figure*}

Similarly, to demonstrate the efficiency of our proposed memetic evolution, the search convergence traces of \emph{ILMA}, \emph{ILMA-R}, and \emph{ILMA-M} on several representative instances of ``E'' and ``S'' series \emph{egl} benchmarks are depicted in Fig. \ref{fig:carponverge} and Fig. \ref{fig:carponvergs}, respectively. As can be observed, for \emph{ILMA} and \emph{ILMA-R}, on the simpler problems such as ``E1-C'' (Fig. \ref{fig:carponverge}(a)), ``E2-B'' (Fig. \ref{fig:carponverge}(b)), ``S1-C'' (Fig. \ref{fig:carponvergs}(b)) and ``S2-B'' (Fig. \ref{fig:carponvergs}(c)), etc., \emph{ILMA} and \emph{ILMA-R} demonstrated competitive convergence speed. Nevertheless, as the complexity of the problems increase, i.e., such as ``E3-C'' (Fig. \ref{fig:carponverge}(d)), ``E4B'' (Fig. \ref{fig:carponverge}(e)), ``E4C'' (Fig. \ref{fig:carponverge}(f)), ``S3-C'' (Fig. \ref{fig:carponvergs}(e)) and ``S4-C'' (Fig. \ref{fig:carponvergs}(f)), etc., \emph{ILMA} consistently converges faster than \emph{ILMA-R} with improved solution attained. Further, for our proposed memetic evolutionary search paradigm, on both the ``E'' and ``S'' series \emph{egl} benchmarks, \emph{ILMA-M} consistently converges more efficiently than \emph{ILMA} and \emph{ILMA-R} on almost all the instances presented. Overall, \emph{ILMA-M} is noted to bring about at least $2\times 10^6$ savings in the number of fitness function evaluations to arrive at the solutions attained by both \emph{ILMA} and \emph{ILMA-R} on most of the CARP instances (e.g., Fig. \ref{fig:carponverge}(a), Fig. \ref{fig:carponverge}(c), Fig. \ref{fig:carponverge}(e), Fig. \ref{fig:carponvergs}(c), Fig. \ref{fig:carponvergs}(d), etc.). For instance, it is worth noting that on instance ``S1-B'' (Fig. \ref{fig:carponverge}(a)), \emph{ILMA-M} used up a total of $1.5\times 10^6$ number of fitness function evaluations to converge at the solution incurred by \emph{ILMA} and \emph{ILMA-R}, which otherwise used up a significant large fitness evaluations of approximately $6\times 10^6$. This equates to a total savings of $4$ times by \emph{ILMA-M} over \emph{ILMA} and \emph{ILMA-R}.

\section{Conclusion}\label{Con}
In this paper, we have proposed a \emph{Memetic Computational Paradigm for search} to model how human solves problems and presented a novel study towards intelligent evolutionary optimization of problems through the transfers of structured knowledge in the form of memes learned from previous problem-solving experiences, to enhance future evolutionary searches.
In particular, the four culture-inspired operators, namely, \emph{Meme Learning}, \emph{Meme Selection}, \emph{Meme Variation} and \emph{Meme Imitation} have been realized based on the HSIC, MMD criterion and K-means Clustering in the context of combinatorial routing problems. In contrast to existing works, the proposed novel memetic search paradigm enables knowledge transfer and reuse in evolutionary optimization across problems of different size, structure, or representation, etc. Last but not least, comprehensive studies on two widely studied NP-hard routing problems, namely, CVRP and CARP, confirmed the efficiency and effectiveness of the proposed paradigm for intelligent evolutionary optimization across problems.
\bibliographystyle{plain}
\bibliography{MTL}

\begin{thebibliography}{10}

\bibitem{PJEA95}
P.~Augerat and J.~M.~Belenguer et. al.
\newblock Computational results with a branch and cut code for the capacitated
  vehicle routing problem.
\newblock {\em Research Report}, pages 949--M, 1995.

\bibitem{SB99}
S.~Blackmore.
\newblock The meme machine.
\newblock {\em Oxford University Press}, 1999.

\bibitem{BorgGroenen2005}
I.~Borg and P.~J.~F. Groenen.
\newblock {\em Modern Multidimensional Scaling: Theory and Applications}.
\newblock Springer, 2005.

\bibitem{KMAM06}
K.~M. Borgwardt, A.~Gretton, M.~J. Rasch, H.~P. Kriegel, B.~Sch$\ddot{o}$lkopf,
  and A.~J. Smola.
\newblock Integrating structured biological data by kernel maximum mean
  discrepancy.
\newblock {\em International Conference on Intelligent Systems for Molecular
  Biology}, pages 49--57, 2006.

\bibitem{PE08}
P.~Bosman and E.~de~Jong.
\newblock Adaptation of a success story in gas: Estimation-of-distribution
  algorithms for tree-based optimization problems.
\newblock In {\em Success in Evolutionary Computation}, volume~92 of {\em
  Studies in Computational Intelligence}, pages 3--18. Springer, 2008.

\bibitem{JAR00}
J.~D. Bransford, A.~L. Brown, and R.~R. Cocking.
\newblock {\em How People Learn: Brain, Mind, Experience, and School}.
\newblock National Academies Press, 2000.

\bibitem{RB96}
R.~Brodie.
\newblock Virus of the mind: The new science of the meme.
\newblock {\em Seattle: Integral Press}, 1996.

\bibitem{JP96}
J.~P. Byrnes.
\newblock {\em Cognitive development and learning in instructional contexts}.
\newblock Allyn and Bacon (Boston), 1996.

\bibitem{XY12}
X.~Chen and Y.~S. Ong.
\newblock A conceptual modeling of meme complexes in stochastic search.
\newblock {\em IEEE Transactions on Systems, Man, and Cybernetics, Part C:
  Applications and Reviews,}, (99):1--8, 2012.

\bibitem{XYMK11}
X.~S. Chen, Y.~S. Ong, M.~H. Lim, and K.~C. Tan.
\newblock A multi-facet survey on memetic computation.
\newblock {\em IEEE Transactions on Evolutionary Computation, In Press},
  (5):591--607, 2011.

\bibitem{XYMS11}
X.~S. Chen, Y.~S. Ong, M.~H. Lim, and S.~P. Yeo.
\newblock Cooperating memes for vehicle routing problems.
\newblock {\em International Journal of Innovative Computing}, 7(11), 2011.

\bibitem{NS69}
N.~Christofides and S.~Eilon.
\newblock An algorithm for the vehicle dispatching problem.
\newblock {\em Operational Research Quarterly}, 20:309 -- 318, 1969.

\bibitem{NAP79}
N.~Christofides, A.~Mingozzi, and P.~Toth.
\newblock The vehicle routing problem.
\newblock {\em Combinatorial Optimization}, pages 315--338, 1979.

\bibitem{NAP81}
N.~Christofides, A.~Mingozzi, and P.~Toth.
\newblock Exact algorithm for the vehicle routing problem, based on spanning
  tree and shortst path relaxations.
\newblock {\em Mathematical Programming}, 20:255–--282, 1981.

\bibitem{Chu199717}
P.~C. Chu and J.~E. Beasley.
\newblock A genetic algorithm for the generalised assignment problem.
\newblock {\em Computers \& Operations Research}, 24(1):17--23, 1997.

\bibitem{JFGA01}
J.~F. Cordeau, G.~Laporte, and A.~Mercier.
\newblock A unified tabu search heuristic for vehicle routing problems with
  time windows.
\newblock {\em Journal of The Operational Research Society}, 52:928–--936,
  2001.

\bibitem{PB97}
P.~Cunningham and B.~Smyth.
\newblock Case-based reasoning in scheduling: Reusing solution components.
\newblock {\em The International Journal of Production Research},
  35(4):2947--2961, 1997.

\bibitem{GJH59}
G.~Dantzig and J.~H. Ramser.
\newblock The truck dispatching problem.
\newblock {\em Management Science}, 6:80--91, 1959.

\bibitem{RD76}
R.~Dawkins.
\newblock The selfish gene.
\newblock {\em Oxford: Oxford University Press}, 1976.

\bibitem{EW59}
E.~W. Dijkstra.
\newblock A note on two problems in connection with graphs.
\newblock {\em Numerische Mathematik}, 1:269–--271, 1959.

\bibitem{RW94}
R.~W. Eglese.
\newblock Routing winter gritting vehicles.
\newblock {\em Discrete Applied Mathematics}, 48(3):231–--244, 1994.

\bibitem{RL96}
R.~W. Eglese and L.~Y.~O. Li.
\newblock A tabu search based heuristic for arc routing with a capacity
  constraint and time deadline.
\newblock {\em in Metaheuristics: theory and applications, I. H. Osman and J.
  P. Kelly, Eds. Boston: Kluwer Academic Publishers}, pages 633–--650, 1996.

\bibitem{LYQA10}
L.~Feng, Y.~S. Ong, Q.~H. Nguyen, and A.~H. Tan.
\newblock Towards probabilistic memetic algorithm: An initial study on
  capacitated arc routing problem.
\newblock {\em IEEE Congress on Evolutionary Computation 2010}, pages 18--23,
  2010.

\bibitem{BR81}
B.~Golden and R.~Wong.
\newblock Capacitated arc routing problems.
\newblock {\em Networks}, 11(3):305--315, 1981.

\bibitem{BJE83}
B.~L. Golden, J.~S. DeArmon, and E.~K. Baker.
\newblock Computational experiments with algorithms for a class of routing
  problems.
\newblock {\em Computer \& Operation Research}, 10(1):47--59, 1983.

\bibitem{GG90}
G.~Grant.
\newblock Memetic lexicon.
\newblock {\em in Principia Cybernetica Web}, 1990.

\bibitem{Gretton05measuringstatistical}
A.~Gretton, O.~Bousquet, A.~Smola, and B.~Sch$\ddot{o}$lkopf.
\newblock Measuring statistical dependence with hilbert-schmidt norms.
\newblock {\em Proceedings Algorithmic Learning Theory}, pages 63--77, 2005.

\bibitem{MT03}
M.~T. Jensen.
\newblock Generating robust and flexible job shop schedules using genetic
  algorithms.
\newblock {\em IEEE Transactions on Evolutionary Computation}, 7(3):275--288,
  2003.

\bibitem{JDDW00}
J.~D. Knowles and D.~W. Corne.
\newblock M-paes: a memetic algorithm for multiobjective optimization.
\newblock {\em IEEE Congress on Evolutionary Computation}, 1:325 -- 332, 2000.

\bibitem{PC04}
P.~Lacomme, C.~Prins, and W.~Ramdane-Ch{\'e}rif.
\newblock Competitive memetic algorithms for arc routing problem.
\newblock {\em Annals of Operational Research}, 141(1--4):159–--185, 2004.

\bibitem{LR96}
L.~Y.~O. Li and R.~W. Eglese.
\newblock An interactive algorithm for vehicle routing for winter-gritting.
\newblock {\em Journal of the Operational Research Society}, 47(2):217–--228,
  1996.

\bibitem{SZKC09}
S.~W. Lin, Z.~J. Lee, K.~C. Ying, and C.~Y. Lee.
\newblock Applying hybrid meta-heuristics for capacitated vehicle routing
  problem.
\newblock {\em Expert Systems with Applications}, 36(2, Part 1), 2009.

\bibitem{SJJM04}
S.~J. Louis and J.~McDonnell.
\newblock Learning with case-injected genetic algorithms.
\newblock {\em IEEE Transactions on Evolutionary Computation}, 8(4):316--328,
  2004.

\bibitem{AL91}
A.~Lynch.
\newblock Thought contagion as abstract evolution.
\newblock {\em Journal of Ideas}, 2:3–--10, 1991.

\bibitem{YK09}
Y.~Mei, K.~Tang, and X.~Yao.
\newblock Improved memetic algorithm for capacitated arc routing problem.
\newblock {\em IEEE Congress on Evolutionary Computation}, pages 1699--1706,
  2009.

\bibitem{Minsky1986}
M.~Minsky.
\newblock {\em The society of mind}.
\newblock Simon \& Schuster, Inc., 1986.

\bibitem{QYM09}
Q.~C. Nguyen, Y.~S. Ong, and M.~H. Lim.
\newblock A probabilistic memetic framework.
\newblock {\em IEEE Transactions on Evolutionary Computation}, 13(3):604--623,
  2009.

\bibitem{YA04}
Y.~S. Ong and A.~J. Keane.
\newblock Meta-lamarckian learning in memetic algorithms.
\newblock {\em IEEE Transactions on Evolutionary Computation}, 8(2):99--110,
  2004.

\bibitem{YMX10}
Y.~S. Ong, M.~H. Lim, and X.~S. Chen.
\newblock Research frontier: Memetic computation - past, present \& future.
\newblock {\em IEEE Computational Intelligence Magazine}, 5(2):24--36, 2010.

\bibitem{SQ10}
S.~J. Pan and Q.~Yang.
\newblock A survey on transfer learning.
\newblock {\em IEEE Transactions on Knowledge and Data Engineering},
  22(10):1345--1359, 2010.

\bibitem{CP04}
C.~Prins.
\newblock A simple and effective evolutionary algorithm for the vehicle routing
  problem.
\newblock {\em Computer \& Operations Research}, 31:1985–--2002, 2004.

\bibitem{MKR04}
M.~Reimann, K.~Doerner, and R.~F. Hartl.
\newblock D-ants: savings based ants divide and conquer the vehicle routing
  problem.
\newblock {\em Computer \& Operations Research}, 31:563–--591, 2004.

\bibitem{MS99}
M.~A. Runco and S.~Pritzker.
\newblock {\em Encyclopedia of Creativity}.
\newblock Academic Press, 1999.

\bibitem{Song2007}
L.~Song, A.~Smola, A.~Gretton, and K.~M. Borgwardt.
\newblock A dependence maximization view of clustering.
\newblock {\em Proceedings of the 24th international conference on Machine
  learning}, pages 815--822, 2007.

\bibitem{KY09}
K.~Tang, Y.~Mei, and X.~Yao.
\newblock Memetic algorithm with extended neighborhood search for capacitated
  arc routing problems.
\newblock {\em IEEE Transactions on Evolutionary Computation},
  13(5):1159–--1166, 2009.

\bibitem{KYX09}
K.~Tang, Y.~Mei, and X.~Yao.
\newblock Memetic algorithm with extended neighborhood search for capacitated
  arc routing problems.
\newblock {\em IEEE Transactions on Evolutionary Computation},
  13(5):1151--1166, 2009.

\bibitem{ARDC09}
A.~Barkat Ullah, R.~Sarker, D.~Cornforth, and C.~Lokan.
\newblock Ama: a new approach for solving constrained real-valued optimization
  problems.
\newblock {\em Soft Computing}, 13(8):741--762, 2009.

\bibitem{GU85}
G.~Ulusoy.
\newblock The fleet size and mix problem for capacitated arc routing.
\newblock {\em Eur. J. Oper. Res.}, 22(3):329--337, 1985.

\bibitem{WWY09}
H.~F. Wang, D.~W. Wang, and S.~X. Yang.
\newblock A memetic algorithm with adaptive hill climbing strategy for dynamic
  optimization problems.
\newblock {\em Soft Computing}, 13(8-9):763--780, 2009.

\bibitem{NY99}
N.~Yoshiike and Y.~Takefuji.
\newblock Vehicle routing problem using clustering algorithm by maximum neural
  networks.
\newblock {\em Intelligent Processing and Manufacturing of Materials}, 2:1109
  -- 1113, 1999.

\bibitem{JIS01}
J.~Zhuang, I.~Tsang, and S.~C.~H. Hoi.
\newblock A family of simple non-parametric kernel learning algorithms.
\newblock {\em Journal of Machine Learning Research (JMLR)}, 12:1313--1347,
  2011.

\end{thebibliography}

\end{document}